# One-Step Abductive Multi-Target Learning with Diverse Noisy Samples and Its Application to Tumour Segmentation for Breast Cancer

Yongquan Yang[1,2], Fengling Li[1,3], Yani Wei[1,3], Jie Chen[1], Ning Chen[4], Mohammad H. Alobaidi[5], Hong Bu[1,3]

1. Institute of Clinical Pathology, West China Hospital, Sichuan University, Chengdu, China
2. ZHONG JIU Flash Medical Technology Co., Ltd., Mianyang, China
3. Department of Pathology, West China Hospital, Sichuan University, Chengdu, China
4. School of Electronics and Information, Xi'an Polytechnic University, Xi'an, China
5. Department of Civil Engineering, McGill University, Montreal, Canada

Email addresses:
Yongquan Yang (remy_yang@foxmail.com),
Fengling Li (lifengling@stu.scu.edu.cn),
Yani Wei (yani_wei@stu.scu.edu.cn),
Jie Chen (jzcjedu@foxmail.com),
Ning Chen (chenn_victor@xpu.edu.cn),
Mohammad H. Alobaidi (mohammad.alobaidi@mcgill.ca),
Hong Bu (hongbu@scu.edu.cn).

Yongquan Yang, Fengling Li and Yani Wei are the co-first authors.
Correspondence to Yongquan Yang (remy_yang@foxmail.com) and Hong Bu (hongbu@scu.edu.cn).

**Abstract**

Recent studies have demonstrated the effectiveness of the combination of machine learning and logical reasoning, including data-driven logical reasoning, knowledge driven machine learning and abductive learning, in inventing advanced technologies for different artificial intelligence applications. One-step abductive multi-target learning (OSAMTL), an approach inspired by abductive learning, via simply combining machine learning and logical reasoning in a one-step balanced multi-target learning way, has as well shown its effectiveness in handling complex noisy labels of a single noisy sample in medical histopathology whole slide image analysis (MHWSIA). However, OSAMTL is not suitable for the situation where diverse noisy samples (DiNS) are provided for a learning task. In this paper, giving definition of DiNS, we propose one-step abductive multi-target learning with DiNS (OSAMTL-DiNS) to expand the original OSAMTL to handle complex noisy labels of DiNS. Applying OSAMTL-DiNS to tumour segmentation for breast cancer in MHWSIA, we show that OSAMTL-DiNS is able to enable various state-of-the-art approaches for learning from noisy labels to achieve more rational predictions. We released a model pre-trained with OSAMTL-DiNS for tumour segmentation in HE-stained pre-treatment biopsy images in breast cancer, which has been successfully applied as a pre-processing tool to extract tumour-associated stroma compartment for predicting the pathological complete response to neoadjuvant chemotherapy in breast cancer.

## 1 Introduction

Perception and cognition, two instincts of human beings for information processing in problem-solving, can be respectively realized by machine learning (Wichmann et al., 2005) and logical reasoning (Rips, 1983). Usually, the paradigm for information processing of machine learning is data-driven, i.e., machine learning constructs the flow of information processing based on collected data. Whereas, the paradigm for information processing of logical reasoning is knowledge-driven, i.e., logical reasoning constructs the flow of information processing based on accumulated knowledge. In most history of artificial intelligence research, machine learning and logical reasoning have been separately developed (Zhou, 2019), due to their significant differences in the representation for the flow of information processing. In recent years, however, with the argument that human beings perform information processing in problem-solving based on the leverage of both perception and cognition, researchers have shown that more effective flows of information processing can be constructed via the combination of machine learning and logical reasoning (D'Amato et al., 2012; D. Li et al., 2020; Zhou, 2019). Existing efforts of combining machine learning and logical reasoning for advanced artificial intelligence technologies can be summarized as in data-driven logical reasoning (DDLR) (D’Amato et al., 2012), knowledge-driven machine learning (KDML) (D. Li et al., 2020) and abductive learning (ABL) (Zhou, 2019).

DDLR maintains logical reasoning as the dominant paradigm for information processing, in which some data-driven flows constructed by machine learning serves as intermediate components. KDML maintains machine learning as the dominant paradigm for information processing, in which some knowledge-driven flows constructed by logical reasoning serves as intermediate components. DDLR prioritizes logical reasoning over machine learning while KDML prioritizes machine learning over logical reasoning. As a result, machine learning and logical reasoning are not fully exploited in DDLR and KDML. To overcome this problem, ABL unifies machine learning and logic reasoning in a balanced way and targets at achieving mutual benefit between machine learning and logical reasoning in an iterative strategy.

Apart from the fact that the original paradigm of ABL has been demonstrated to be effective in some real-world applications (Dai et al., 2019; Y. X. Huang et al., 2020), recent studies (Yang et al., 2020, 2024) show approach that simply exploits the concept of unifying machine learning and logical reasoning in a balanced way from ABL is as well fairly effective in specific application. This approach is so called one-step abductive multi-target learning (OSAMTL) (Yang et al., 2024), since it only combines machine learning and logical reasoning in a one-step balanced way without targeting at a mutual benefit in an iterative strategy. OSAMTL targets at alleviating the situation where it is often difficult for experts to manually achieve the accurate ground-truth labels, which leads to labels with complex noisy for a learning task (aka, learning from noisy labels (Frénay & Verleysen, 2014; Karimi et al., 2020; Song et al., 2022)). Based on the concept of unifying machine learning and logical reasoning in a balanced way from ABL, OSAMTL exploits multiple inaccurate targets abduced from a single noisy sample via logical reasoning to achieve more reasonable predictions via multi-target learning. On a H. pylori segmentation task in medical histopathology whole slide images (Hanna et al., 2020) analysis (MHWSIA) (Yang et al., 2020, 2024), for the first time, OSAMTL has been reported to possess potentials in handling complex noisy labels in MHWSIA. More details about existing approaches in combination of machine learning and logical reasoning can be found in section 2.1.

Since it is based on a single noisy sample, OSAMTL is naturally not suitable for the situation where diverse noisy samples (DiNS) are provided for a learning task. DiNS contain at least two types of noisy samples, where diversity exists between any two non-repeating noisy samples. Due to this property, DiNS can constitute a noisy data that have very complex noise. In this paper, formally giving definition of DiNS, we propose one-step abductive multi-target learning with DiNS (OSAMTL-DiNS) and provide analyses of OSAMTL-DiNS compared with the original OSAMTL. Being able to handle DiNS, OSAMTL-DiNS only require very inaccurately labelled (complex noisy) samples to produce a predictive model. This property forms the major advantage of OSAMTL-DiNS, since the data preparation can be much less labour-tensive though at least two types of noisy samples are needed. Thus, OSAMTL-DiNS is suitable to address some tasks in the field of medical analysis where the problem of low consistency always exists. Low consistency, here in the context of DiNS, can refer to that large is the difference between the noisy distributions of two different noisy samples prepared by experts for a same medical analysis task, which usually results in more complex noisy in data annotations. More details about DiNS and the proposed OSAMTL-DiNS can be found in section 2.2.

Quantitative evaluation of tumour in breast cancer can provide clues important to subsequent therapy of breast cancer (Pu et al., 2020; Yau et al., 2022). The key point is to achieve tumour segmentation for breast cancer (TSfBC), which is a fundamental key technique that can be leveraged to calculate the tumour-stroma ratio which has been proven to be a prognostic factor in breast cancer (de Kruijf et al., 2011). Existing deep learning (LeCun et al., 2015) enhanced approaches that can be leveraged to achieve TSfBC can be classified into two schemes: learning with noisy-free/accurate labels (Bhattacharjee et al., 2022; Priego-Torres et al., 2020, 2022) and learning with noisy/inaccurate labels (Diao et al., 2022; G. Xu et al., 2019; Y. Xu et al., 2014). The first type of scheme adopts the supervised learning paradigm. However, due to the difficulty in accurately labelling the tumour in breast cancer on whole slide images, very limited noisy-free data is often available, which will inevitably limit the generalization of the prediction model. The second type of scheme adopts the weakly supervised learning paradigm, which avoids the problem of the difficulty in obtaining noisy-free data faced by the first type of scheme. However, the popular strategy of using image patch-level labels to achieve pixel-level segmentation (Diao et al., 2022; G. Xu et al., 2019; Y. Xu et al., 2014) has the drawbacks in MHWSIA that the work load of image patch-level labelling can still be very massive due to the large size of whole slide images and the prediction results can be very coarse at high resolution.

To alleviate this situation, in this paper, we applied the proposed OSAMTL-DiNS to TSfBC in MHWSIA. As OSAMTL-DiNS only requires very inaccurately labelled (complex noisy) samples, the difficulty in pixel-level labelling for the task of TSfBC in MHWSIA is considerably reduced. Referring to the proposed OSAMTL-DiNS, we implemented an OSAMTL-DiNS-based image semantic segmentation solution for TSfBC in MHWSIA and conducted extensive experiments to demonstrate the potentials of OSAMTL-DiNS in MHWSIA. More details about the application of OSAMTL-DiNS to TSfBC and corresponding strategies for experimental conduction can be found in section 2.3 and section 2.4.

Experiment results show that OSAMTL-DiNS is able to enable various existing approaches for learning from noisy labels (Algan & Ulusoy, 2021; Frénay & Verleysen, 2014; Sukhbaatar & Fergus, 2014), including naively learning from noisy labels, Forward, Backward (Patrini et al., 2017), Boost-Hard, Boost-Soft (Arazo et al., 2019; Reed et al., 2015), D2L (Ma et al., 2018), SCE (Wang et al., 2019), Peer (Liu & Guo, 2020), DT-Forward (Yao et al., 2020), and NCE-SCE (Ma et al., 2020), to achieve more rational predictions. We also released a predictive model pre-trained with OSAMTL-DiNS for tumour segmentation in HE-stained pre-treatment biopsy images in breast cancer, which has been successfully applied as a pre-processing tool to extract tumour-associated stroma compartment for predicting the pathological complete response to neoadjuvant chemotherapy in breast cancer (F. Li et al., 2022).

In summary, the contributions of this work are as follows:

- One-step abductive multi-target learning with diverse noisy samples (OSAMTL-DiNS) is proposed, which only require very inaccurately labelled (complex noisy) samples to produce a predictive model.
- The proposed OSAMTL-DiNS is applied to address the task of tumour segmentation for breast cancer (TSfBC) in medical histopathology whole slide image analysis (MHWSIA).
- Extensive experiments show that the proposed OSAMTL-DiNS is able to enable various existing approaches for learning from noisy labels to achieve more rational predictions in the task of TSfBC in MHWSIA, which reflects the potential effectiveness of OSAMTL-DiNS in handling complex noisy labels in MHWSIA.
- A predictive model pre-trained with OSAMTL-DiNS for tumour segmentation in HE-stained pre-treatment biopsy images in breast cancer is released and has been successfully applied as a pre-processing tool to extract tumour-associated stroma compartment for predicting the pathological complete response to neoadjuvant chemotherapy in breast cancer, which reflects the potentials of using OSAMTL-DiNS to help building basic tools for MHWSIA.

## 2 Material and Methods

This section is structured as follows. In section 2.1, we briefly formalize existing methodologies for combining machine learning and logical reasoning and summarize their differences. In section 2.2, we present the methodology of the proposed OSAMTL-DiNS and corresponding summaries about its properties. In section 2.3, we apply the proposed OSAMTL-DiNS to tumour segmentation for breast cancer (TSfBC) in MHWSIA. In section 2.4, on the basis of the implemented application of OSAMTL-DiNS to TSfBC, we present the experimental strategies for conducting extensive experiments to investigate the contributions of OSAMTL-DiNS in handling complex noisy labels.

### *2.1 Combination of machine learning and logical reasoning*

In this subsection, we briefly formalize the methodologies for combining of machine learning and logical reasoning and summarize their differences. Firstly, in section 2.1.1, we give the preliminary formalizations about machine learning and logical reasoning. Secondly, we formalize existing methodologies of combining machine learning and logical reasoning respective in sections 2.1.2-5. Finally, in section 2.1.6, we summarize the differences of existing methodologies based on their formalizations.

#### 2.1.1 Preliminary

**Machine learning** Commonly for a machine learning task in artificial intelligence research, a collected dataset containing certain instances ($x$) and corresponding labels ($y$) for the task is provided. The objective of the task here is to estimate a predictive function $f$ parameterized by $\theta$ ($f(\theta)$) that can map $x$ to corresponding predictions ($f(x;\theta)$) which are as correct as possible compared with $y$. Formally, the objective of a machine learning task can be expressed as

$$\tilde{f}(\tilde{\theta}) = \underset{f \in \Theta_f, \theta \in \Theta_\theta}{\arg min} \ ||f(x;\theta) - y||,$$

where $\Theta_f$ is the function space of $f$, $\Theta_\theta$ is the parameter space of $\theta$ corresponding to $f$ and $||f(x;\theta) - y||$ denotes the error between $f(x;\theta)$ and $y$.

**Logical reasoning** Commonly for a logical reasoning task in artificial intelligence research, a collected dataset containing certain instances ($x$) and corresponding labels ($y$) for the task and a collected knowledge base ($kb$) containing various prior knowledge or facts about the task are both provided. The objective of the task here is to search a reasoning path $r$ that can from $x$ and $y$ draw conclusions ($r(< x, y >)$) consistent with some knowledge or facts in $kb$. Formally, the objective of a logical reasoning task can be expressed as

$$\tilde{r} = \underset{r \in \Theta_r}{\arg search} \ r(< x, y >) \cong kb,$$

where $\Theta_r$ is the reasoning path space of $r$ and $r(< x, y >) \cong kb$ denotes $r(< x, y >)$ is consistent with $kb$.

### 2.1.2 Data-driven logical reasoning

Data-driven logical reasoning (DDLR) (D'Amato et al., 2012) adapts machine learning to logical reasoning. Formally, the objective of a DDLR task can be expressed as

$$\tilde{r} = \underset{r \in \Theta_r}{\arg search} \ r\left(< x, \tilde{f}(x;\tilde{\theta}) >\right) \cong kb, \quad (1)$$

where $\tilde{f}(\tilde{\theta}) = \underset{f \in \Theta_f, \theta \in \Theta_\theta}{\arg min} \ ||f(x;\theta) - y||$.

### 2.1.3 Knowledge-driven machine learning

Knowledge-driven machine learning (KDML) (D. Li et al., 2020) adapts logical reasoning to machine learning. Formally, the objective of a task can be expressed as

$$\tilde{f}(\tilde{\theta}) = \underset{f \in \Theta_f, \theta \in \Theta_\theta}{\arg min} \ ||f(\tilde{r}(< x, y >);\theta) - y||, \quad (2)$$

where $\tilde{r} = \underset{r \in \Theta_r}{\arg search} \, r(< x, y >) \cong kb$.

### 2.1.4 Abductive learning

Abductive learning (ABL) (Zhou, 2019) has two important concepts: 1) unifying machine learning and logic reasoning in a balanced way, and 2) targeting at achieving a mutual benefit between machine learning and logical reasoning in an iterative strategy. Formally, the objective of a ABL task can be expressed as

$$\begin{cases} \tilde{r}_k = \underset{r \in \Theta_r}{\arg search} \ r\left(< x, \tilde{y}_{ml,k} >\right) \cong kb \,, \ \tilde{y}_{ml,k} = \tilde{f}_{k-1}\left(x; \tilde{\theta}_{k-1}\right) \\ \tilde{f}_k(\tilde{\theta}) = \underset{f \in \Theta_f, \theta \in \Theta_\theta}{\arg min} \ ||f(x;\theta) - \tilde{y}_{lr,k}||, \ \tilde{y}_{lr,k} = \tilde{r}_k\left(< x, \tilde{y}_{ml,k} >\right) \end{cases}, \quad (3)$$

where $k = 1, \cdots, N$ and $\tilde{f}_0(\tilde{\theta}_0)$ can be a pre-trained machine learning model.

### 2.1.5 One-step abductive multi-target learning

One-step abductive multi-target learning (OSAMTL) (Yang et al., 2024) only combines machine learning and logical reasoning in a one-step balanced multi-target learning way. Formally, the objective of a OSAMTL task can be expressed as

$$\begin{cases} \tilde{r} = \underset{r \in \Theta_r}{\arg search}\ r(< x, y >) \cong kb \\ \tilde{f}(\tilde{\theta}) = \underset{f \in \Theta_f, \theta \in \Theta_\theta}{\arg min}\ ||f(x;\theta) - \tilde{y}_{lr}||,\ \tilde{y}_{lr} = \tilde{r}(< x, y >) = \{t_1, \cdots, t_n\} \end{cases}, \quad (4)$$

where $y$ can be rough labels provided by experts or predictions of a pre-trained machine learning model, and $\tilde{y}_{lr}$ is a set of multiple inaccurate targets which can be expressed as $\{t_1, \cdots, t_n\}$.

#### 2.1.6 Summary

From formula (1), we can note that DDLR prioritizes logical reasoning over machine learning by maintaining logical reasoning as the dominant paradigm for information processing, in which the results produced by machine learning serves as some inputs. From formula (2), we can note that KDML prioritizes machine learning over logical reasoning by maintaining machine learning as the dominant paradigm for information processing, in which the results produced by logical reasoning serves as some inputs. In the paradigms of both DDLR and KDML, machine learning and logical reasoning are not fully exploited by adapting one to the other.

From formula (3), we can note that the results produced by machine learning serve as some inputs of logical reasoning and the results produced by logical reasoning serve as the target of machine learning. This indicates that machine learning and logical reasoning are placed in equal positions in the objective of a ABL task. Iteratively, let $k = 1, \cdots, N$, ABL is able to fully exploit machine learning and logical reasoning, which cannot be fulfilled by DDLR or KDML.

From formula (4), we can note that it can be regarded as a special case of formula (3) when $k = 1$. In formula (4), the objective of logical reasoning is independent of the objective of machine learning and the results produced by logical reasoning serve as a set of multiple targets of machine learning. These two indicate, being a special case of ABL, OSAMTL only combines machine learning and logical reasoning in a one-step balanced way without targeting at a mutual benefit in an iterative strategy.

Overall speaking, the differences of DDLR, KDML, ABL and OSAMTL for the combination of machine learning and logical reasoning can be summarized as Fig. 1.

### *2.2 Proposed OSAMTL-DiNS*

In this subsection, we propose one-step abductive multi-target learning with diverse noisy samples (OSAMTL-DiNS) to expand the original OSAMTL to handle complex noisy labels of diverse noisy samples. The outline for the methodology of OSAMTL-DiNS is shown as Fig. 2. Specifically, section 2.2.1 gives the definition for DiNS; section 2.2.2 presents the formalizations for the proposed OSAMTL-DiNS; and section 2.2.3 summarizes some properties of the proposed OSAMTL-DiNS.

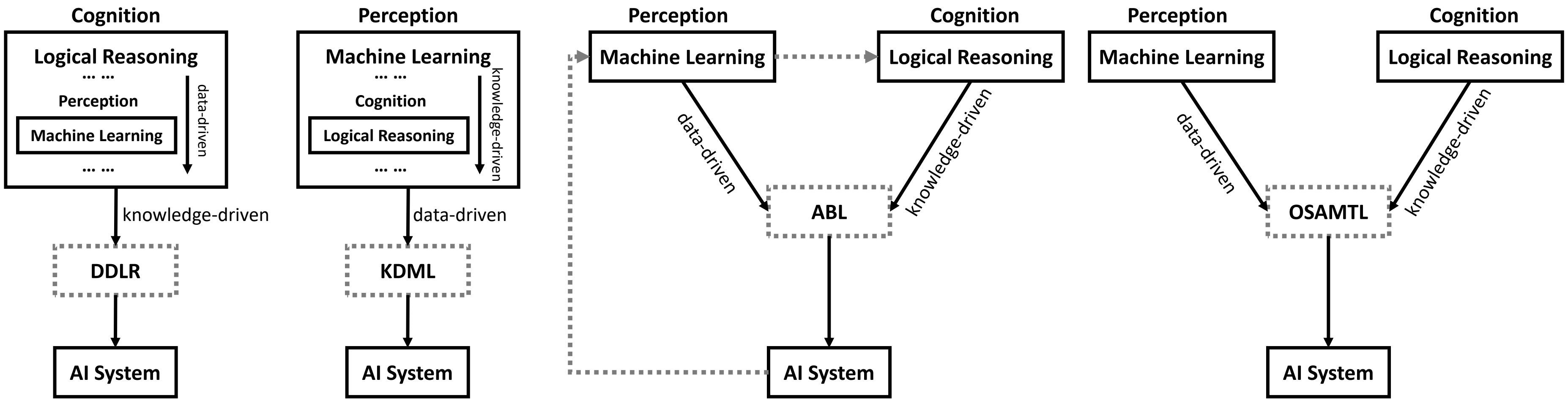

Figure. 1. Summarization for differences of DDLR, KDML, ABL and OSAMTL for the combination of machine learning and logical reasoning.

### 2.2.1 Definition of diverse noisy samples

A noisy sample ($NS_*$) consists of an instance sample ($IS_*$) and a noisy label sample ($NLS_*$). An instance sample contains a number of instances ($I$) and a noisy label sample contains a number of noisy labels ( $NL$ ). The instances of $IS_*$ are one-to-one corresponding to the noisy labels of $NLS_*$. Formally, a noisy sample can be denoted by

$$NS_* = \{IS_*, NLS_*\} = \left\{\{I_{*,1}, \cdots, I_{*,n_*}\}, \{NL_{*,1}, \cdots, NL_{*,n_*}\}\right\}$$
$$= \left\{\left(I_{*,1}, NL_{*,1}\right), \cdots \left(I_{*,n_*}, NL_{*,n_*}\right)\right\},$$

where $n_*$ is the number for instances or noisy labels of $NS_*$.

The diversity of two noisy samples ($Div_{a,b}$) can be evaluated by the differences between their instances and corresponding noisy labels. Formally, the diversity of two noisy samples can be denoted by

$$Div_{a,b} = Differenciate\ (NS_a, NS_b)$$
$$= Differenciate(IS_a, IS_b) * Differenciate(NLS_a, NLS_b).$$

For simplicity, we define $Div_{a,b} \in [0,1]$, where 1 signifies diversity exists between $NS_a$ and $NS_b$ while 0 indicates the opposite.

Diverse noisy samples (DiNS) have at least two noisy samples, where diversity exists between any two non-repeating noisy samples. Due to this property, DiNS can constitute a noisy data that have very complex noise. Formally, diverse noisy samples can be defined as

$$DiNS = \{NS_1, \cdots, NS_d\} = \left\{\{IS_1, NLS_1\}, \cdots, \{IS_d, NLS_d\}\right\}$$
$$= \left\{\left\{\{I_{1,1}, \cdots, I_{1,n_1}\}, \{NL_{1,1}, \cdots, NL_{1,n_1}\}\right\}, \cdots, \left\{\{I_{d,1}, \cdots, I_{d,n_d}\}, \{NL_{d,1}, \cdots, NL_{d,n_d}\}\right\}\right\}$$
$$= \left\{\left\{\left(I_{1,1}, NL_{1,1}\right), \cdots \left(I_{1,n_1}, NL_{1,n_1}\right)\right\}, \cdots, \left\{\left(I_{d,1}, NL_{d,1}\right), \cdots \left(I_{d,n_d}, NL_{d,n_d}\right)\right\}\right\}$$
$$s.t. \quad \forall a, \forall b \in \{1, \cdots, d\}\ and\ a \neq b,\ \exists\ Div_{a,b} = 1.$$

### 2.2.2 One-step abductive multi-target learning with DiNS

With the given definition of DiNS, we propose one-step abductive multi-target learning with diverse noisy samples (OSAMTL-DiNS). OSAMTL-DiNS constitutes of four components, including input materials, one-step logical reasoning, target rearrangement and multi-target learning.

**Input materials** The input materials of OSAMTL-DiNS include some given diverse noisy samples ($DiNS$) and a knowledge base ($KB$) containing a list of domain knowledge about the true target of a specific task. Referring to the formulations of $DNS$ presented in Section 3.1, the input materials of OSAMTL can be more specifically denoted as follows

$$DiNS = \{NS_1, \cdots, NS_d\},$$
$$KB = \{K_1, \cdots, K_b\}.$$

**One-step abductive logical reasoning with DiNS** With the given $DiNS$ and $KB$, the one-step logical reasoning procedure of OSAMTL-DiNS, which consists of four substeps, abduces multiple targets containing information consistent with the domain knowledge about the true target of a specific task.

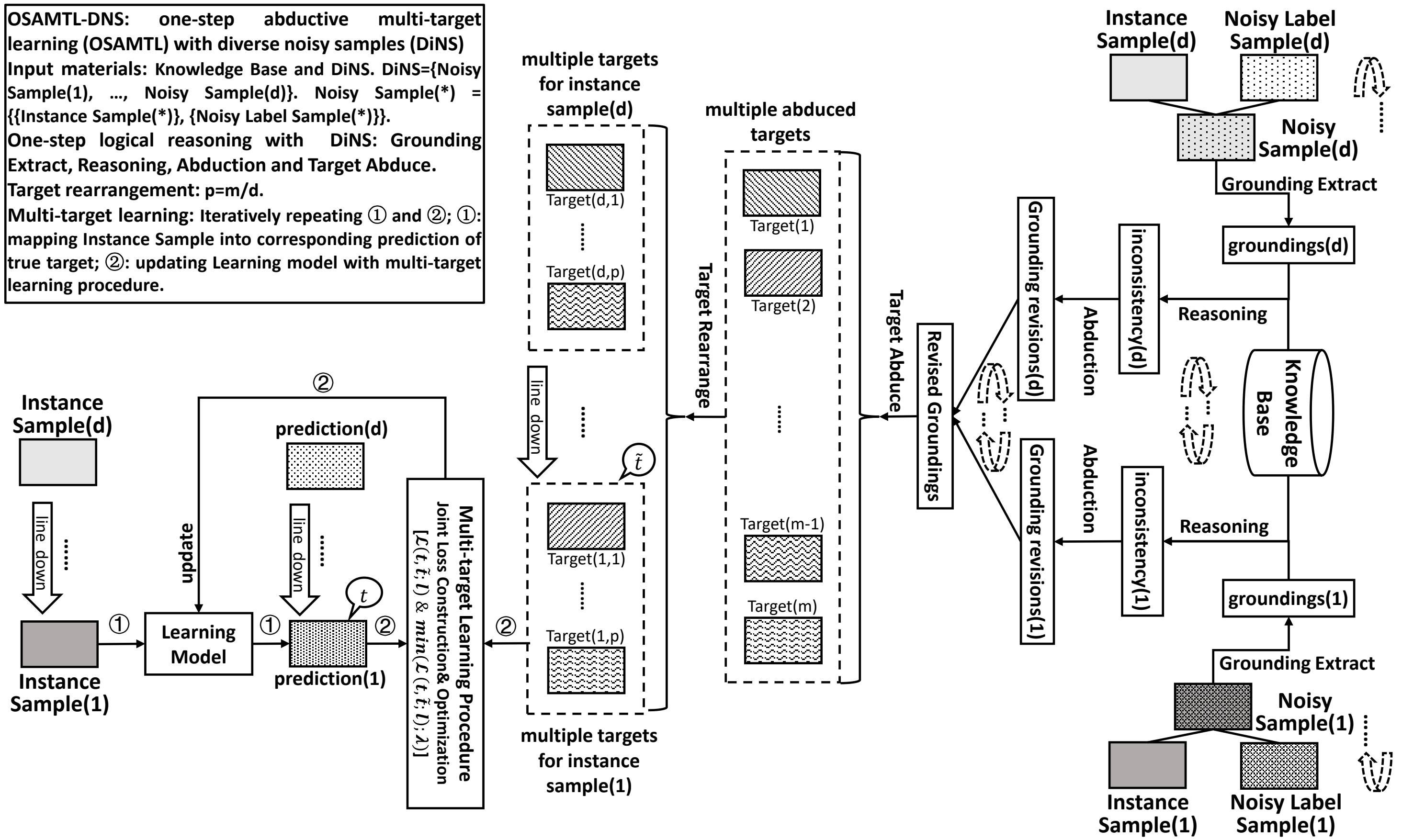

Figure. 2. The outline for the methodology of OSAMTL-DiNS. For simplicity of elaborating the methodology of OSAMTL-DiNS, we assume that each instance sample ($IS_*$) and each noisy sample ($NLS_*$) of the given diverse noisy samples ($DNS$) only have one instance and one noisy label respectively. This simplified elaboration can be deduced to the situation where each instance sample ($IS_*$) and each noisy sample ($NLS_*$) of the given diverse noisy samples ($DNS$) have a set of instances and a set of noisy labels respectively.

The substep one extracts a list of groundings from the given set of noisy label samples that can describe the logical facts contained in the given diverse noisy samples. Formally, this grounding extract ($GE$) step can be expressed as

$$\begin{aligned} G &= GE(DiNS; p^{GE}) = \{GE(NS_1; p^{GE_1}), \cdots, GE(NS_d; p^{GE_d})\} \\ &= \left\{G_1 = \{G_{1,1}, \cdots, G_{1,r_1}\}, \cdots, G_d = \{G_{d,1}, \cdots, G_{d,r_d}\}\right\}. \end{aligned} \quad (5)$$

Via logical reasoning, the substep two estimates the inconsistencies between the extracted groundings $G$ and the domain knowledge in the knowledge base $KB$. Formally, this reasoning ($R$) step can be expressed as

$$\begin{aligned} IC &= R(G, KB; p^{R}) = \{R(G_1, KB; p^{R_1}), \cdots, R(G_d, KB; p^{R_d})\} \\ &= \left\{IC_1 = \{IC_{1,1}, \cdots, IC_{1,i_1}\}, \cdots, IC_d = \{IC_{d,1}, \cdots, IC_{d,i_d}\}\right\}. \end{aligned} \quad (6)$$

The substep three revises the groundings of the given set of noisy label samples by logical abduction based on reducing the estimated inconsistency $IC$. Formally, this logical abduction ($LA$) step can be expressed as

$$\begin{aligned} RG &= LA(IC; p^{LA}) = \{LA(IC_1; p^{LA_1}), \cdots, LA(IC_d; p^{LA_d})\} \\ &= \left\{\{GR_{1,1}, \cdots, GR_{1,z_1}\}, \cdots, \{GR_{d,1}, \cdots, GR_{d,z_d}\}\right\} \\ &= \{RG_1(GR_{1,1}), \cdots, RG_s(GR_{d,z_d})\} \quad s.t. \quad s = \textstyle\sum_{i=1}^{d} z_i. \end{aligned} \quad (7)$$

Finally, the substep four leverages the revised groundings $RG$ to abduce multiple targets containing information consistent with our domain knowledge about the true target for the instance sample of the given noisy smaples. Formally, this target abduce ($TA$) step can be expressed as

$$\begin{aligned} T &= TA(\{RG\}; p^{TA}) \\ &= \left\{\begin{matrix} TA(\{RG_{1,1}, \cdots, RG_{1,e_1}\}; p^{TA_1}), \cdots, \\ TA(\{RG_{m,1}, \cdots, RG_{m,e_m}\}; p^{TA_m}) \end{matrix}\right\} \\ &= \{T_1, \cdots, T_m\} \quad s.t. \ \{RG_{*,1}, \cdots, RG_{*,e_*}\} \in RG. \end{aligned} \quad (8)$$

In the four formulas (1)-(4), each $p^*$ denotes the hyper-parameters corresponding to the implementation of respective expression.

**Target rearrangement** The target rearrangement procedure of OSAMTL-DiNS rearranges the multiple targets abduced by the one-step logical reasoning with DiNS into ordered multiple targets that are corresponding to each instance sample of the given diverse noisy samples. Formally, the target rearrangement ( $TR$ ) procedure of OSAMTL-DiNS can be expressed as

$$\begin{aligned} \widetilde{ts} &= TR(T, p^{TR}) \\ &= \{TR(\{T_i | i \in [1, m], T_i \in T\}, p^{TR_1}), \cdots, TR(\{T_i | i \in [1, m], T_i \in T\}, p^{TR_d})\} \\ &= \left\{\widetilde{ts}_1 = \{\widetilde{ts}_{1,1}, \cdots, \widetilde{ts}_{1,r_1}\}, \cdots, \widetilde{ts}_d = \{\widetilde{ts}_{d,1}, \cdots, \widetilde{ts}_{d,r_d}\}\right\} \quad s.t. \ \ \widetilde{ts}_* \in T \ and \ r_* > 1. \end{aligned} \quad (9)$$

Here, $p^{TR}$ denotes the hyper-parameters corresponding to the implementation of a target rearrangement procedure.

The formula (5) reflects that each instance sample of the given diverse noisy samples ($IS_*$) has corresponding $r$ number of targets ($\tilde{t}_* = \left\{\{\tilde{t}_{*,1}, \cdots, \tilde{t}_{*,r_*}\}\right\}$) abduced by the one-step logical reasoning. Referring to the situation where each instance sample of the given diverse noisy samples only has one instance, the formular (5) can be deduced to imply that each instance contained in an instance sample of the given diverse noisy samples has corresponding multiple targets abduced by one-step logical reasoning. With this implication, we can rewrite the rearranged targets for an instance sample of the given diverse noisy samples as

$$\widetilde{ts}_* = \{\widetilde{ts}_{*,1}, \cdots, \widetilde{ts}_{*,r_*}\} = \left\{\{\tilde{t}_{(*,1),1}, \cdots, \tilde{t}_{(*,1),r_*}\}, \cdots, \{\tilde{t}_{(*,n_*),1}, \cdots, \tilde{t}_{(*,n_*),r_*}\}\right\}. \quad (10)$$

As a result, the instances ($I$) contained in the given diverse noisy samples and corresponding rearranged multiple targets ($\tilde{t}$) can be denoted as

$$I = \{IS_1 \cup \cdots \cup IS_d\} = \left\{\{I_{1,1}, \cdots, I_{1,n_1}\} \cup \cdots \cup \{I_{d,1}, \cdots, I_{d,n_d}\}\right\} = \{I_1, \cdots, I_n\},$$

$$\tilde{t} = \{\widetilde{ts}_1 \cup \cdots \cup \widetilde{ts}_d\} = \left\{\begin{matrix} \left\{\{\tilde{t}_{(1,1),1}, \cdots, \tilde{t}_{(1,1),r_1}\}, \cdots, \{\tilde{t}_{(1,n_1),1}, \cdots, \tilde{t}_{(1,n_1),r_1}\}\right\} \cup \cdots \cup \\ \left\{\{\tilde{t}_{(d,1),1}, \cdots, \tilde{t}_{(d,1),r_d}\}, \cdots, \{\tilde{t}_{(d,n_d),1}, \cdots, \tilde{t}_{(d,n_d),r_d}\}\right\} \end{matrix}\right\}$$

$$= \left\{\tilde{t}_1 = \{\tilde{t}_{1,1}, \cdots, \tilde{t}_{1,r_1}\}, \cdots, \tilde{t}_n = \{\tilde{t}_{n,1}, \cdots, \tilde{t}_{n,r_n}\}\right\}, \quad s.t. \quad n = \sum_{i=1}^{d} n_i. \quad (11)$$

**Multi-target learning** The multi-target learning procedure of OSAMTL-DiNS is carried out on the basis of a specifically constructed learning model that maps input instances ($I$) into its corresponding target prediction ($t$), which can be expressed as

$$t = LM(I, \omega) = \{t_1, \cdots, t_n\}. \quad (12)$$

Here, $LM$ is short for learning model, and $\omega$ denotes the hyper-parameters corresponding to the construction of a specific learning model.

The multi-target learning procedure of OSAMTL-DiNS, which constitutes of a joint loss construction and optimization, imposes the rearranged multiple targets ($\tilde{t}$) upon machine learning to constrain the prediction of the learning model ($t$). The joint loss is constructed by estimating the error between $t_*$ and $\tilde{t}_*$, which can be expressed as

$$\mathcal{L}(t, \tilde{t}; \ell) = \frac{1}{n}\sum_{j=1}^{n}\sum_{i=1}^{r_j} \alpha_i \ell(t_j, \tilde{t}_{j,i}) \ \ s.t. \ \sum_{i=1}^{r_j} \alpha_i = 1. \quad (13)$$

Here, $\ell$ denotes the hyper-parameters corresponding to the construction of the basic loss function, and $\alpha_i$ is the weight for estimating the loss between $t_j$ and an abduced target ($\tilde{t}_{j,i}$) contained in $\tilde{t}_j$. Then, the objective can be expressed as

$$\min_t(\mathcal{L}(t, \tilde{t}; \ell); \lambda). \quad (14)$$

Here, $\lambda$ denotes the hyper-parameters corresponding to the implementation of an optimization approach.

### 2.2.3 Summary

OSAMTL-DiNS inherits properties from OSAMTL (Yang et al., 2024), including the difference of OSAMTL from abductive learning (ABL) (Zhou, 2019) and the distinctiveness of OSAMTL from various state-of-the-art approaches that are based on pre-assumptions about noisy-labelled instances (Arazo et al., 2019; Liu & Guo, 2020; Ma et al., 2018, 2020; Reed et al., 2015; Wang et al., 2019; Yao et al., 2020) or need premised requirements (Acuna et al., 2019; Li, J., Socher, R., & Hoi, 2020; Xiao et al., 2015) to be carried out to handle noisy labels. OSAMTL-DiNS also inherits the essence of the multi-target learning procedure of OSAMTL, which is that the multi-garget learning procedure can enable the learning model to learn from a weighted summarization of multiple targets that contain information consistent to our prior knowledge about the true target of a specific task. For more details of these properties of OSAMTL-DiNS inherited from OSAMTL, readers can refer to (Yang et al., 2024).

OSAMTL-DiNS improves OSAMTL. The one-step logical reasoning procedure of OSAMTL-DiNS abduces multiple targets using given diverse noisy samples and knowledge base, while the logical reasoning procedure of OSAMTL abduces multiple targets using given one noisy sample and knowledge base. OSAMTL-DiNS is suitable to address tasks where a knowledge base and multiple noisy samples are available and each noisy sample has a different noisy distribution in labels. From this side, OSAMTL can only handle a subset of the tasks for which OSAMTL-DiNS are suitable, since it can only handle the situation where the available noisy sample has one noisy distribution in labels. Thus OSAMTL-DiNS expands the generalization of original OSAMTL to a wider range of tasks.

Besides, OSAMTL-DiNS possesses an extra target rearrangement procedure that rearranges the multiple targets abduced by the one-step logical reasoning with DiNS into ordered multiple targets corresponding to the instances contained in the given diverse noisy samples. As a result, the instances contained in the given diverse noisy samples and corresponding rearranged multiple targets can be conveniently employed by the multi-target learning procedure of OSAMTL-DiNS.

As OSAMTL-DiNS is proposed to handle DiNS, it can produce a predictive model based on very inaccurately labelled (complex noisy) samples. This property forms the major advantage of OSAMTL-DiNS, since the data preparation can be much less expensive and less labour-tensive though at least two types of noisy samples are needed. Being able to handle DiNS, OSAMTL-DiNS is suitable to address some tasks in the field of medical analysis where the problem of low consistency always exists. Low consistency, here in the context of DiNS, can refer to that large is the difference between the noisy distributions of two different noisy samples prepared by experts for a same medical analysis task, which usually results in more complex noisy in data annotations.

Notably, OSAMTL-DiNS is independent from the machine learning architectures, which is the same as the original OSAMTL. This property of OSAMTL-DiNS makes it possible to be combined with various state-of-the-art deep learning architectures chosen for specific tasks, such as deep convolutional neural networks, transformers or very large models.

## *2.3 OSAMTL-DiNS applied on tumour segmentation for breast cancer*

In this section, we apply OSAMTL-DiNS on tumour segmentation for breast cancer (TSfBC) in medical histopathology whole slide image analysis (MHWSIA). Firstly, in section 2.3.1 we introduce the background of TSfBC, and in section 2.3.2 we give application settings of applying OSAMTL-DiNS on TSfBC. Secondly, in sections 2.3.3-6, we implement the OSAMTL-DiNS based solution for TSfBC. Finally, in section 2.3.7 we summarize the outline of the implementation of the OSAMTL-DiNS based solution for TSfBC.

### 2.3.1 Tumour segmentation for breast cancer

Fig. 2 shows the illustrations for two tasks of TSfBC. The two tasks include a task that aims to segment tumour in HE-stained pre-treatment biopsy images and a task that aims to segment residual tumour in HE-stained post-treatment surgical resection images. From the illustrations presented in Fig. 3.A and Fig. 3.B, we can note that it is indeed difficulty to accurately annotate the true targets for both segmentation tasks. Referring to these illustrations and additional suggestions from pathology experts, we here claim that the tumour segmentation task in HE-stained post-treatment surgical resection images is more difficult than the tumour segmentation task in HE-stained pre-treatment biopsy images.

Existing deep learning (LeCun et al., 2015) enhanced approaches that can be leveraged to achieve TSfBC can be classified into two schemes: learning with noisy-free/accurate labels (Bhattacharjee et al., 2022; Priego-Torres et al., 2020, 2022) and learning with noisy/inaccurate labels (Diao et al., 2022; G. Xu et al., 2019; Y. Xu et al., 2014). The first type of scheme adopts the supervised learning paradigm. However, due to the difficulty in accurately labelling the tumour in breast cancer on whole slide images (Hanna et al., 2020), very limited noisy-free data is often available, which will inevitably limit the generalization of the prediction model. The second type of scheme adopts the weakly supervised learning paradigm, which avoids the problem of the difficulty in obtaining noisy-free data faced by the first type of scheme. However, the popular strategy of using image patch-level labels to achieve pixel-level segmentation (Diao et al., 2022; G. Xu et al., 2019; Y. Xu et al., 2014) has the drawback in MHWSIA that the work load of image patch-level labelling can still be very massive due to the large size of whole slide images. To alleviate this situation, we apply the proposed OSAMTL-DiNS to TSfBC.

**2.3.2 Application setting**

On one hand, due to the difficulty to visually annotate the true target for TSfBC on HE-stained images, we asked two pathology experts to provide weak annotations: one pathology expert only aims to as accurate as possible exclude the non-true target on one data set; and another pathology expert only aims to as accurate as possible include the target on another dataset. As a result, two diverse noisy samples (DiNS) are provided for TSfBC from the vision perspective. On the other hand, existing knowledge of pathology can semantically give clear descriptions, that is a list of semantic sentences from pathological knowledge can present what is the true target for TSfBC. As a result, we also asked the two pathology experts to provide a knowledge base (KB) about the true target for TSfBC from the semantic perspective.

The noisy labels contained in visual DiNS are inaccurate but can be easily transformed into learnable target, meanwhile, the sentences contained in semantic KB are clear but cannot be easily transformed into learnable target. It is desirable to take the advantages of both visual DiNS and semantic KB into machine learning to achieve more reasonable predictions. Fortunately, the proposed OSAMTL-DiNS framework can take advantages of both visual DiNS and semantic KB by transforming visual DiNS into multiple learnable inaccurate targets containing information consistent with the knowledge of semantic KB for the true target via one-step abductive logical reasoning. Thus, on the basis of the provided visual DiNS and semantic KB, we employ

OSAMTL-DiNS to address tumour segmentation in HE-stained pre-treatment biopsy images and tumour segmentation in HE-stained post-treatment surgical resection images.

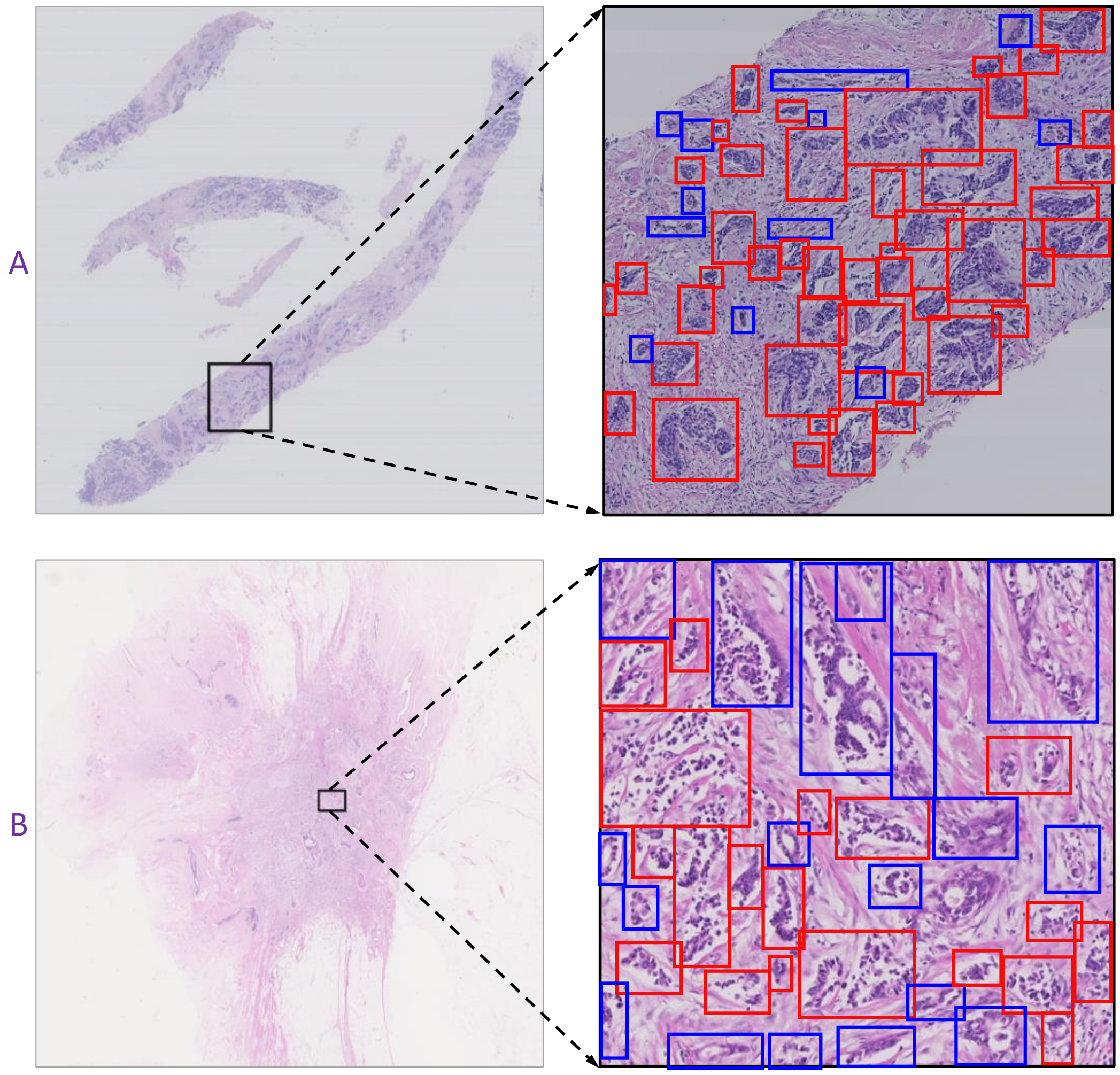

Figure. 3. Illustrations for two tumour segmentation tasks for breast cancer. A: tumour segmentation in HE-stained pre-treatment biopsy images. B: tumour segmentation in HE-stained post-treatment surgical resection images. A-Left: A 1× magnification shown medical histopathology whole slide image digitalized from a HE-stained pre-treatment biopsy slide; A-Right: A 10×magnification shown image patch cropped from the left whole slide image at the boxed area. B-Left: A 0.5×magnification shown medical histopathology whole slide image digitalized from a HE-stained post-treatment surgical resection slide; B-Right: A 10×magnification shown image patch cropped from the left whole slide image at the boxed area. Red boxes in A-Right image or B-Right image: areas that confidently contain tumour. Blue boxes in A-Right image or B-Right image: areas that possibly (not sure) contain tumour. Rest of A-Right image or B-Right image: areas that confidently do not contain tumour. Pathology experts annotated these boxes shown in A-Right image and B-Right image.

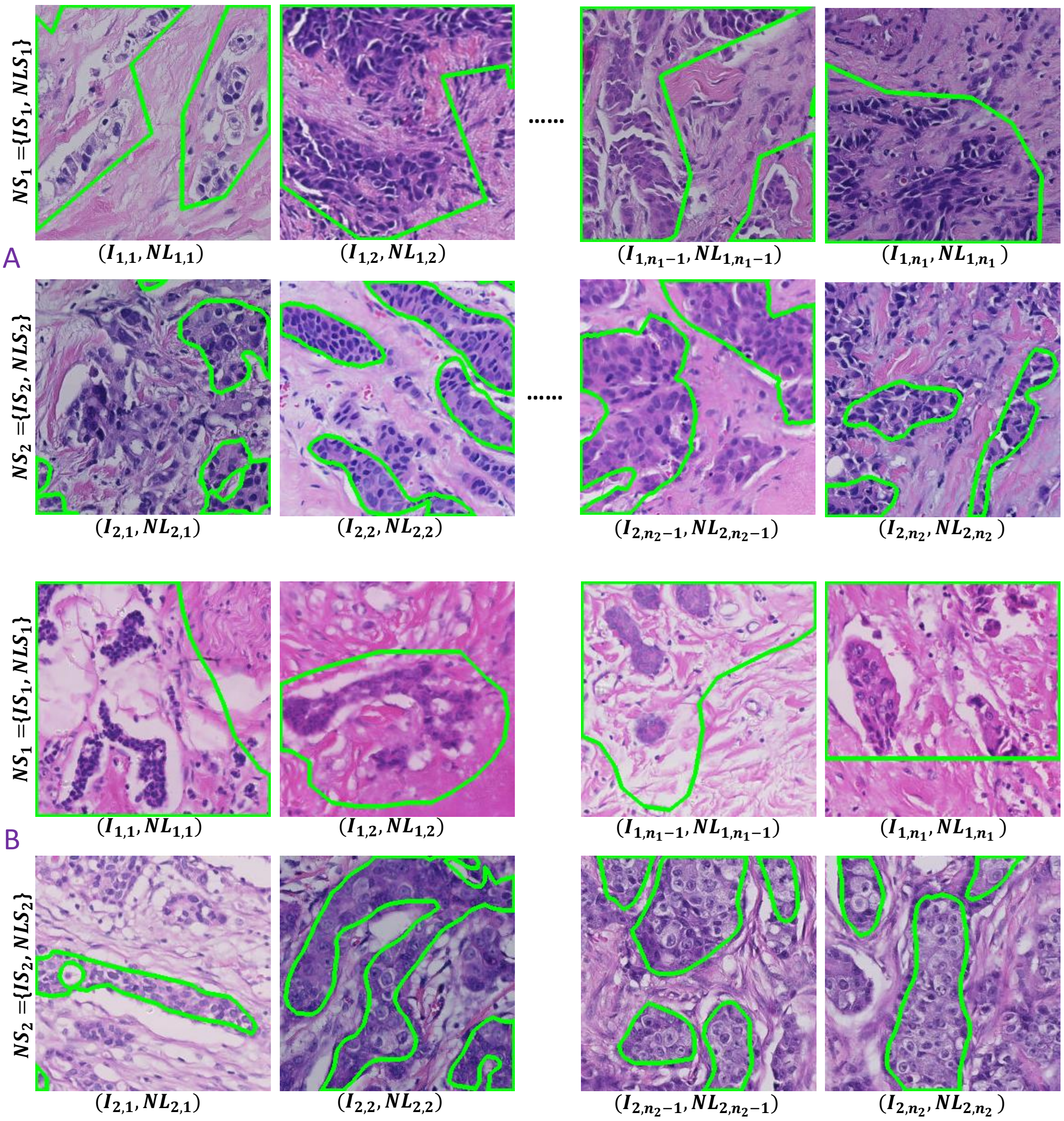

Figure. 4. Examples of diverse noisy samples provided for the two tumour segmentation tasks for breast cancer. A: diverse noisy samples for tumour segmentation in HE-stained pre-treatment biopsy images. B: diverse noisy samples for tumour segmentation in HE-stained post-treatment surgical resection images.

### 2.3.3 Input materials

**Diverse noisy samples** The diverse noisy samples (DiNS) provided for the two tumour segmentation tasks for breast cancer are denoted as follows.

$$DNS = \{NS_1, NS_2\} = \{\{IS_1, NLS_1\}, \{IS_2, NLS_2\}\}$$

$$= \left\{\left\{\{I_{1,1}, \cdots, I_{1,n_1}\}, \{NL_{1,1}, \cdots, NL_{1,n_1}\}\right\}, \left\{\{I_{2,1}, \cdots, I_{2,n_2}\}, \{NL_{2,1}, \cdots, NL_{2,n_2}\}\right\}\right\}$$

$$= \left\{\left\{(I_{1,1}, NL_{1,1}), \cdots (I_{1,n_1}, NL_{1,n_1})\right\}, \left\{(I_{2,1}, NL_{2,1}), \cdots (I_{2,n_2}, NL_{2,n_2})\right\}\right\}.$$

Some examples of the provided DiNS for the two tasks are shown as Fig. 4. The noisy labels contained in the provided DiNS can significantly alleviate the mission for accurate ground-truth labels, however, these noisy labels also suffer from severe inaccuracy compared with the true target for TSfBC. From Fig. 4, we can observe that many non-tumour areas are included as tumour by the labels of $NS_1$ while many tumour areas are excluded as non-tumour by the labels of $NS_2$.

**Knowledge base** Regarding to existing knowledge of pathology, the knowledge base (KB) provided for the two tumour segmentation tasks for breast cancer are shown as Table 1.

$$KB = \{K_1, K_2, K_3, K_4, K_5, K_6\}.$$

Table 1. Knowledge provided for tumour segmentation for breast cancer

| Knowledge Base |
|---|
| $K_1$: Tumour is composed of tumour cells. |
| $K_2$: Tumour cells may be arranged in cords, clusters, and trabeculae. |
| $K_3$: Some tumours are characterized by predominantly solid or syncytial infiltrative pattern with little associated stroma. |
| $K_4$: Cytoplasm of tumour cell is eosinophilic and vacuolated. |
| $K_5$: Nuclei of tumour cell is enlarged and chromatin of tumour cell is vacuolated. |
| $K_6$: Nuclei of tumour cell is degenerated. |

### 2.3.4 One-step abductive logical reasoning with DiNS

**Grounding Extract** The Grounding Extract step takes the provided $DNS$ (shown in Fig. 2) as input and produces a list of groundings that describe the logical facts of the provided DiNS. Referring to Eq. (5), we use the semantics contained in the provided $DNS$ as $p^{GE}$ to implement the Grounding Extract step, which can produce groundings as follows

$$\begin{aligned} G &= GE(DNS; \{semantics\ contained\ in\ DNS\}) \\ &= \left\{ \begin{array}{l} GE\left( \begin{array}{c} NS_1 = \{IS_1, NLS_1\}; \\ \{semantics\ contained\ in\ NLS_1\ for\ labeling\ IS_1\} \end{array} \right), \\ GE\left( \begin{array}{c} NS_2 = \{\{IS_2, NLS_2\}\}; \\ \{semantics\ contained\ in\ NLS_2\ for\ labeling\ IS_2\} \end{array} \right) \end{array} \right\}, \\ &= \left\{ \begin{array}{l} G_1 = \{G_{1,1}, G_{1,2}\}, \\ G_2 = \{G_{2,1}, G_{2,2}\} \end{array} \right\}. \end{aligned}$$

Details of the extracted groundings are provided in Table 2.

Table 2. Details of the extracted groundings

| Extracted Groundings |
|---|
| $G_{1,1}$: pixels of $IS_1$ outside the polygons of $NLS_1$ are tumour negatives |
| $G_{1,2}$: pixels of $IS_1$ inside the polygons of $NLS_1$ are tumour positives |
| $G_{2,1}$: pixels of $IS_2$ inside the polygons of $NLS_2$ are tumour positives |
| $G_{2,2}$: pixels of $IS_2$ outside the polygons of $NLS_2$ are tumour negatives |

**Reasoning** The Reasoning step takes the $G$ extracted by the Grounding Extract step and the provided $KB$ as inputs and produces a list of inconsistencies that describe the gap between the extracted groundings $G$ and the provided $KB$ from various perspectives. On the basis of the extracted groundings $G$ and the provided $KB$, we derive two reasonings (Reasoning 1 and Reasoning 2). The validity of the two derived reasonings are respectively proved by Proof-R1 and Proof-R2 which are provided in Supplementary 1.

**Reasoning 1**. *If $G_1$ is given and $KB$ is given, then inconsistency between the false tumour negatives covered by $G_{1,1}$ and the false tumour negatives covered by $KB$ is low, inconsistency between the false tumour positives covered by $G_{1,2}$ and the false tumour positives covered by $KB$ is high, inconsistency between the true tumour positives covered by $G_{1,2}$ and the true tumour positives covered by $KB$ is low, and inconsistency between the true tumour negatives covered by $G_{1,1}$ and the true tumour negatives covered by $KB$ is high*.

**Reasoning 2**. *If $G_2$ is given and $KB$ is given, then inconsistency between the false tumour positives covered by $G_{2,1}$ and the false tumour positives covered by $KB$ is low, inconsistency between the false tumour negatives covered by $G_{2,2}$ and the false tumour negatives covered by $KB$ is high, inconsistency between the true tumour negatives covered by $G_{2,2}$ and the true tumour negatives covered by $KB$ is low, and inconsistency between the true tumour positives covered by $G_{1,1}$ and the true tumour positives covered by $KB$ is high.*

Referring to Eq. (6), we use Reasoning 1 and Reasoning 2 as $p^R$ to implement the Reasoning step, which can produce estimated inconsistencies between the extracted groundings $G$ and the provided $KB$ as follows

$$\begin{aligned} IC &= R(G, KB; \{Reasoning\ 1, Reasoning\ 2\}) \\ &= \{R(G_1, KB; Reasoning\ 1), R(G_2, KB; Reasoning\ 2)\} \\ &= \begin{Bmatrix} IC_1 = \{IC_{1,1}, IC_{1,2}, IC_{1,3}, IC_{1,4}\}, \\ IC_2 = \{IC_{2,1}, IC_{2,2}, IC_{2,3}, IC_{2,4}\} \end{Bmatrix} \end{aligned}$$

Details of the estimated inconsistencies are provided in Table 3.

Table 3. Details of the estimated inconsistencies

| Estimated Inconsistencies |
|---|
| $IC_{1,1}$: inconsistency between the false tumour negatives covered by $G_{1,1}$ and the false tumour negatives covered by $KB$ is low |
| $IC_{1,2}$: inconsistency between the false tumour positives covered by $G_{1,2}$ and the false tumour positives covered by $KB$ is high |
| $IC_{1,3}$: inconsistency between the true tumour positives covered by $G_{1,2}$ and the true tumour positives covered by $KB$ is low |
| $IC_{1,4}$: inconsistency between the true tumour negatives covered by $G_{1,1}$ and the true tumour negatives covered by $KB$ is high |
| $IC_{2,1}$: inconsistency between the false tumour positives covered by $G_{2,1}$ and the false tumour positives covered by $KB$ is low |
| $IC_{2,2}$: inconsistency between the false tumour negatives covered by $G_{2,2}$ and the false tumour negatives covered by $KB$ is high |
| $IC_{2,3}$: inconsistency between the true tumour negatives covered by $G_{2,2}$ and the true tumour negatives covered by $KB$ is low |
| $IC_{2,4}$: inconsistency between the true tumour positives covered by $G_{2,1}$ and the true tumour positives covered by $KB$ is high |

**Abduction** The Abduction step takes the $IC$ estimated by Reasoning step and the $G$ extracted by the Grounding Extract step as inputs and produces a list of revised groundings that reduce the inconsistencies in $IC$. On the basis of the estimated $IC$ and the extracted $G$, we derive two reasonings (Reasoning 3 and Reasoning 4). The validity of the two derived reasonings are respectively proved by Proof-R3 and Proof-R4 which are provided in Supplementary 1.

**Reasoning 3**. *If* $IC_1$ *is given, then* $G_{1,1}$ *should not be revised to remain* $IC_{1,1}$, $G_{1,2}$ *should not be revised to remain* $IC_{1,3}$, $G_{1,1}$ *should be revised to reduce* $IC_{1,4}$, *and* $G_{1,2}$ *should be revised to reduce* $IC_{1,2}$.

**Reasoning 4**. *If* $IC_2$ *is given, then* $G_{2,1}$ *should not be revised to remain* $IC_{2,1}$, *then* $G_{2,2}$ *should not be revised to remain* $IC_{2,3}$, $G_{2,1}$ *should be revised to reduce* $IC_{2,2}$, *and* $G_{2,2}$ *should be revised to reduce* $IC_{2,4}$.

Referring to Eq. (7), we use Reasoning 3 and Reasoning 4 as $p^{LA}$ to implement the Abduction step, which can produce revised groundings as follows

$$\begin{aligned} RG &= LA(\{IC\};\{Reasoning\ 3, Reasoning 4\}) \\ &= \begin{Bmatrix} LA(\{IC_1\}; Reasoning\ 3), \\ LA(\{IC_2\}; Reasoning\ 4) \end{Bmatrix} \\ &= \begin{Bmatrix} \{GR_{1,1}, GR_{1,2}, GR_{1,3}, GR_{1,4}\}, \\ \{GR_{2,1}, GR_{2,2}, GR_{2,3}, GR_{2,4}\} \end{Bmatrix} \\ &= \begin{Bmatrix} RG_1(GR_{1,1}), RG_2(GR_{1,2}), RG_3(GR_{1,3}), RG_4(GR_{1,4}), \\ RG_5(GR_{2,1}), RG_6(GR_{2,2}), RG_7(GR_{2,3}), RG_8(GR_{2,4}) \end{Bmatrix} \end{aligned}$$

Details of the grounding revisions and the revised groundings are respectively provided in Table 4 and Table 5.

Table 4. Details of the grounding revisions

| Grounding Revisions |
|---|
| $GR_{1,1}$: $G_{1,1}$ should not be revised to remain $IC_{1,1}$ |
| $GR_{1,2}$: $G_{1,2}$ should not be revised to remain $IC_{1,3}$ |
| $GR_{1,3}$: $G_{1,1}$ should be revised to reduce $IC_{1,4}$ |
| $GR_{1,4}$: $G_{1,2}$ should be revised to reduce $IC_{1,2}$ |
| $GR_{2,1}$: $G_{2,1}$ should not be revised to remain $IC_{2,1}$ |
| $GR_{2,2}$: $G_{2,2}$ should not be revised to remain $IC_{2,3}$ |
| $GR_{2,3}$: $G_{2,1}$ should be revised to reduce $IC_{2,4}$ |
| $GR_{2,4}$: $G_{2,2}$ should be revised to reduce $IC_{2,2}$ |

Table 5. Details of the revised groundings

| Revised Groundings |
|---|
| $RG_1(GR_{1,1})$: $= G_{1,1}$, pixels of $IS_1$ outside the polygons of $NLS_1$ are tumour negatives |
| $RG_2(GR_{1,2})$: $= G_{1,2}$, pixels of $IS_1$ inside the polygons of $NLS_1$ are tumour positives |

$RG_3(GR_{1,3})$: pixels of $IS_1$ outside the polygons of $NLS_1$ are not exactly true tumour negatives
$RG_4(GR_{1,4})$: pixels of $IS_1$ inside the polygons of $NLS_1$ are not exactly true tumour positives
$RG_5(GR_{2,1})$: $= G_{2,1}$, pixels of $IS_2$ inside the polygons of $NLS_2$ are tumour positives
$RG_6(GR_{2,2})$: $= G_{2,2}$, pixels of $IS_2$ outside the polygons of $NLS_2$ are tumour negatives
$RG_7(GR_{2,3})$: pixels of $IS_2$ inside the polygons of $NLS_2$ are not exactly true tumour positives
$RG_8(GR_{2,4})$: pixels of $IS_2$ outside the polygons of $NLS_2$ are not exactly true tumour negatives

**Target Abduce** The target abduce step takes the $RG$ produced by the Abduction step as input and abduces a list of multiple targets to more appropriately represent the true target of tumour for breast cancer. On the basis of the input $RG$, we derive four reasonings (Reasoning 5, Reasoning 6, Reasoning 7 and Reasoning 8). The validity of the four derived reasonings are respectively proved by Proof-R5-9 which are provided in Supplementary 1.

**Reasoning 5**. *If $RG_1$ is given and $RG_2$ is given, then a target ($T_1$) can be abduced from the union of $RG_1$ and $RG_2$, and $T_1$ has a high recall of true tumour positives and a high precision of true tumour negatives.*

**Reasoning 6**. *If $RG_5$ is given and $RG_6$ is given, then a target ($T_2$) can be abduced from the union of $RG_5$ and $RG_6$, and $T_2$ has a high precision of true tumour positives and a high recall of true tumour negatives.*

**Reasoning 7**. *If the target ($T_1$) abduced from the union of $RG_1$ and $RG_2$ is given, $RG_3$ is given and $RG_4$ is given, then $T_1$ has a low precision of true tumour positives and a low recall of true tumour negatives.*

**Reasoning 8**. *If the target ($T_2$) abduced from the union of $RG_5$ and $RG_6$ is given, $RG_7$ is given and $RG_8$ is given, then $T_2$ has a low recall of true tumour positives and a low precision of true tumour negatives.*

**Reasoning 9**. *If the target ($T_1$) abduced from the union of $RG_1$ and $RG_2$ is given and the target ($T_2$) abduced from the union of $RG_5$ and $RG_6$ is given, then a target ($T_3$) can be abduced by improving $T_1$ with $T_2$, a target ($T_4$) can be abduced by improving $T_2$ with $T_1$, $T_3$ can have a relatively higher precision of true tumour positives than $T_1$ and a relatively higher recall of true tumour negatives than $T_1$, and $T_4$ can have a relatively higher recall of true tumour positives than $T_2$ and a relatively higher precision of true tumour negatives than $T_2$.*

Referring to Eq. (8), we use Reasoning 5-9 as $p^{TA}$ to implement the Target Abduce step, which can be denoted as follows

$$
\begin{aligned}
T &= TA(RG; \{Reasoning\ 5, \cdots, Reasoning\ 9\}) \\
&= \left\{ \begin{array}{c} TA(\{RG_1, RG_2\}; Reasoning\ 5), TA(\{RG_4, RG_5\}; Reasoning\ 6), \\ TA\left( \begin{array}{c} \{RG_1, RG_2, RG_3, RG_4, RG_5, RG_6\}; \\ \{Reasoning\ 5, Reasoning\ 7, Reasoning\ 9\} \end{array} \right), \\ TA\left( \begin{array}{c} \{RG_4, RG_5, RG_6, RG_7, RG_1, RG_2\}; \\ \{Reasoning\ 6, Reasoning\ 8, Reasoning\ 9\} \end{array} \right) \end{array} \right\} \\
&= \{T_1, T_2, T_3, T_4\}.
\end{aligned}
$$

With Reasoning 5, we generate the target $T_1$ by the union of $RG_1$ and $RG_2$ to keep a high recall of true tumour positives and a high precision of true tumour negatives. With Reasoning 6, we generate the target $T_2$ by the union of $RG_4$ and $RG_5$ to keep high precision of true tumour positives and a high recall of true tumour negatives. With Reasoning 5-9, we generate the target $T_3$ to keep a relatively higher precision of true tumour positives than $T_1$ and a relatively higher recall of true tumour negatives than $T_1$. Specifically, we first employed $NLS_1$ to train an image semantic segmentation model and then tested it on the instance images of $NLS_2$ to produce $T_3$. With Reasoning 6-9, we generate the target $T_4$ to keep a relatively higher recall of true tumour positives than $T_2$ and a relatively higher precision of true tumour negatives than $T_2$. Specifically, we first employed $NLS_2$ to train an image semantic segmentation model and then tested it on the instance images of $NLS_1$ to produce $T_4$. Some examples of the abduced multiple targets for the two tumour segmentation tasks for breast cancer illustrated in Fig. 2 are provided in Supplementary 2.

**2.3.5 Target rearrangement**

The target rearrangement step takes the $T$ produced by the Target Abduce step as input and produce ordered multiple targets that are corresponding to each of the two given diverse noisy samples. On the basis of the input $T$, we derive two reasonings (Reasoning 10 and Reasoning 11). The validity of the two derived reasonings are respectively proved by Proof-R10 and Proof-R11 which are provided in Supplementary 1.

**Reasoning 10**. *If $T_1$ is given and $T_3$ is given, then $T_1$ and $T_3$ can be combined to approximate the true target for $NS_1$.*

**Reasoning 11**. *If $T_2$ is given and $T_4$ is given, then $T_2$ and $T_4$ can be combined to approximate the true target for $NS_2$.*

Referring to Eq. (9), we use Reasoning 10 and Reasoning 11 as $p^{TR}$ to implement the Target Rearrangement step, which can produce rearranged targets as follows

$$\widetilde{ts} = TR(T, \{Reasoning10, Reasoning11\})$$
$$= \begin{Bmatrix} TR(\{T_1, T_3\}, Reasoning10), \\ TR(\{T_2, T_4\}, Reasoning11) \end{Bmatrix}$$
$$= \left\{\widetilde{ts}_1 = \left\{\widetilde{ts}_{1,1}, \widetilde{ts}_{1,2}\right\} = \{T_1, T_3\}, \widetilde{ts}_2 = \left\{\widetilde{ts}_{2,1}, \widetilde{ts}_{2,2}\right\} = \{T_4, T_2\}\right\}.$$

On the basis of the rearranged $\widetilde{ts}$ and referring to Eq. (10)-(11), the instances ($I$) contained in the given $DNS$ and their corresponding rearranged multiple targets are denoted as follows.

$$I = \{I_1, \cdots, I_n\}, \quad \tilde{t} = \left\{\tilde{t}_1 = \left\{\tilde{t}_{1,1}, \tilde{t}_{1,2}\right\}, \cdots, \tilde{t}_n = \left\{\tilde{t}_{n,1}, \tilde{t}_{n,2}\right\}\right\}, \quad s.t.\ n = n_1 + n_2.$$

Some examples of the rearranged multiple targets are provided in Supplementary 2.

**2.3.6 Multi-target learning**

Referring to Eq. (12), we employ deep convolutional neural network (DCNN) to implement the learning model ($LM$) that maps input instances ($I$) into its corresponding target predictions ($t$) by

$$t = LM(I, DCNN) = \{t_1, \cdots, t_n\}.$$

The DCNN employed for the learning model of tumour segmentation here is a symmetric image semantic segmentation architecture that we constructed in (Yang et al., 2024) by referring to the commonly used FCN (Shelhamer et al., 2017), which is the representative for the fully convolutional networks based solutions and has inspired various other solutions (Badrinarayanan et al., 2017; Chen et al., 2018; Falk et al., 2019; Fu et al., 2019; Zhao et al., 2017) achieving state-of-the-art performances in image semantic segmentation.

On the basis of the rearranged targets $\tilde{t}$ and the target prediction of implemented learning model and referring to Eq. (13), we employ cross entropy (CE) to implement the multi-target learning procedure by

$$\mathcal{L}(t,\tilde{t};CE)=\frac{1}{n}\sum_{j=1}^{n}\left(\alpha_1 CE\left(t_j,\tilde{t}_{j,1}\right)+\alpha_2 CE\left(t_j,\tilde{t}_{j,2}\right)\right)\ \ s.t.\ \ \alpha_1+\alpha_2=1.$$

Referring to Eq. (14), we employ stochastic gradient descent (SGD) to implement the objective by

$$\min_{t}(\mathcal{L}(t,\tilde{t};CE);SGD).$$

**2.3.7 Summary**

Referring to the contents of sections 2.3.3-6, the outline of the OSAMTL-DiNS solution implemented for the TSfBC in HE-stained pre-treatment biopsy images is shown as Fig. 5 and Fig. 6. On the basis of Fig. 5 and Fig. 6, the simplified outline of the OSAMTL-DiNS solution implemented for residual TSfBC in HE-stained post-treatment surgical resection images is simplistically shown as Fig. 7.

## *2.4 Experimental strategies*

On the basis of the OSAMTL-DiNS solutions implemented for the two tumour segmentation tasks for breast cancer, we conduct extensive experiments to demonstrate the contributions of OSAMTL-DiNS in handling complex noisy labels. In this subsection, we give descriptions about the overall design, data preparation, evaluation metrics and experimental details for conducting the experiments. In section 2.4.1, we give the overall design for the conducted experiments. The details about the data preparation for the conducted experiments are provided in section 2.4.2. The section 2.4.3 gives the metrics used to evaluation the results of the conducted experiments. Finally, in section 2.4.4, we describe the experimental details for the conducted experiments.

**2.4.1 Overall design**

Due to the fact that accurate/noisy-free ground-truth labels for the two tumour segmentation tasks for breast cancer illustrated in Fig. 2 (tumour segmentation task in HE-stained pre-treatment biopsy images and tumour segmentation task in HE-stained post-treatment surgical resection images) are quite difficult to collect, we prepare a small noisy-free dataset for each of the two tumour segmentation tasks for usual evaluations with accurate ground truth labels (AGTLs). The prepared AGTLs can be regarded as golden standards for performance evaluation.

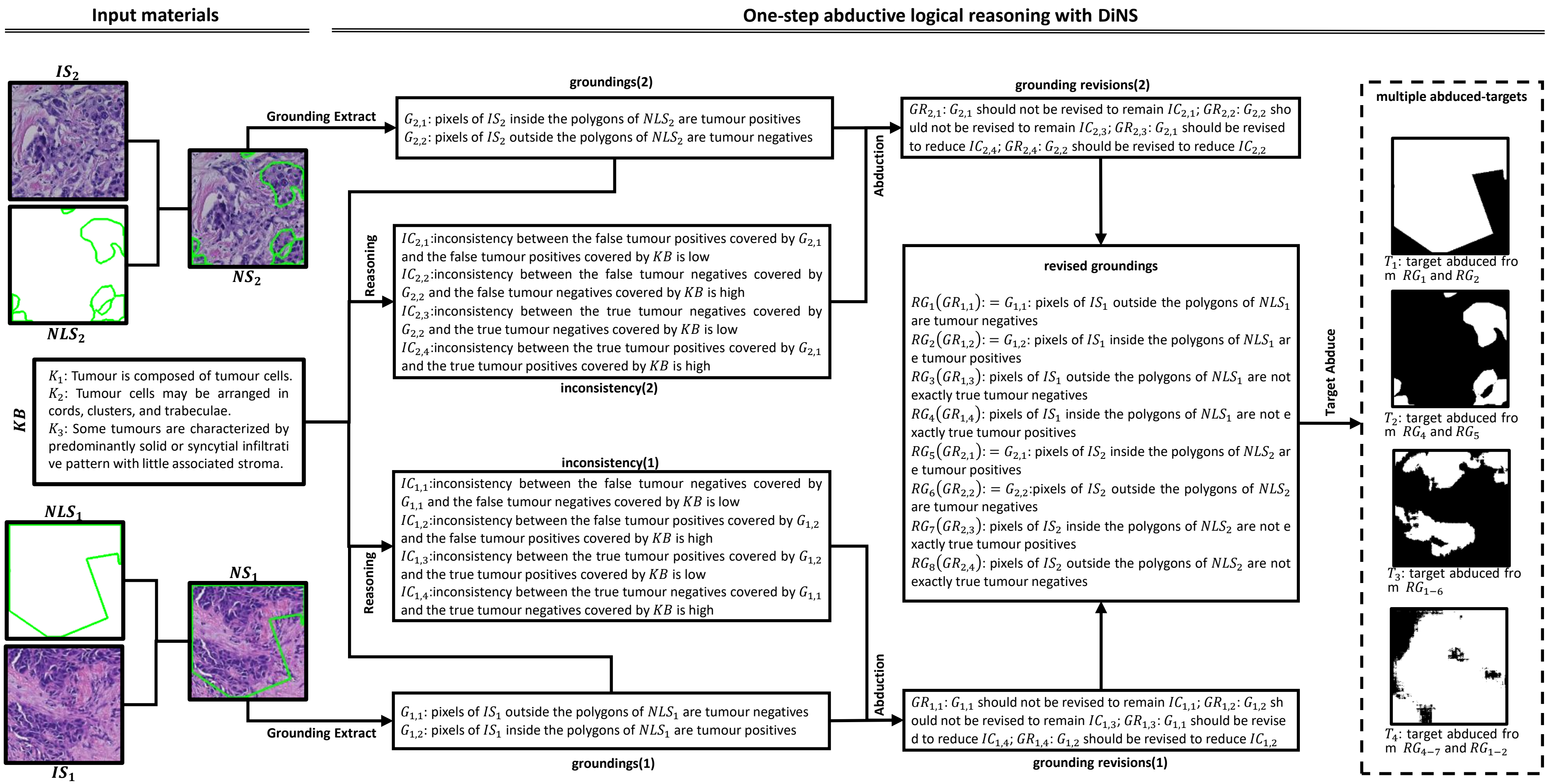

Figure. 5. The outline for the input materials and one-step abductive logical reasoning with DiNS for the OSAMTL-DiNS solution implemented for the TSfBC in HE-stained pre-treatment biopsy images.

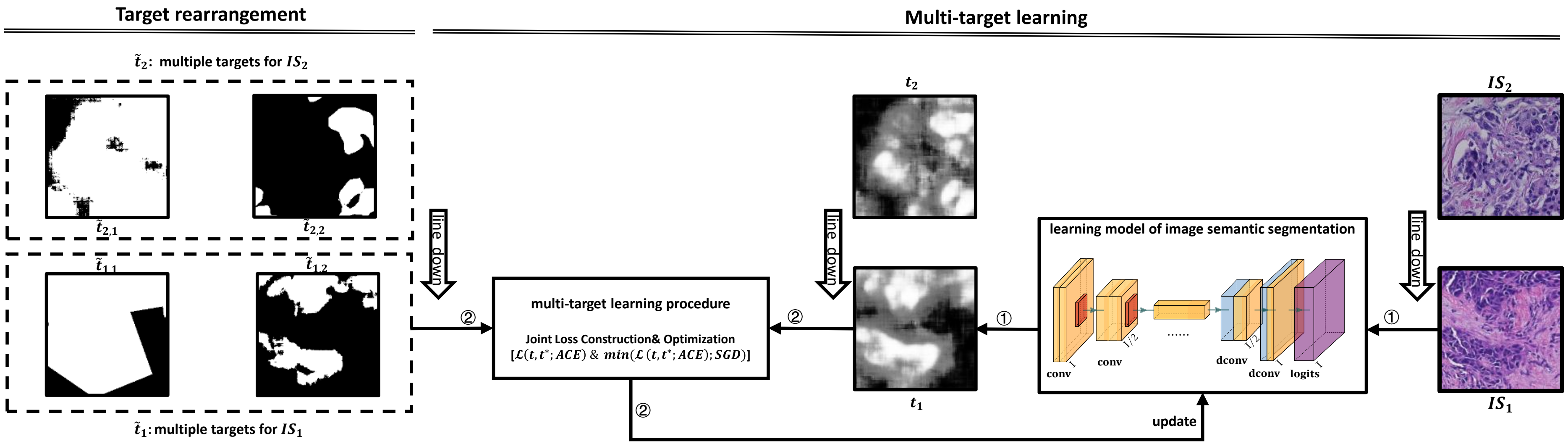

Figure. 6. The outline for the target rearrangement and multi-target learning for the OSAMTL-DiNS solution implemented for the TSfBC in HE-stained pre-treatment biopsy images.

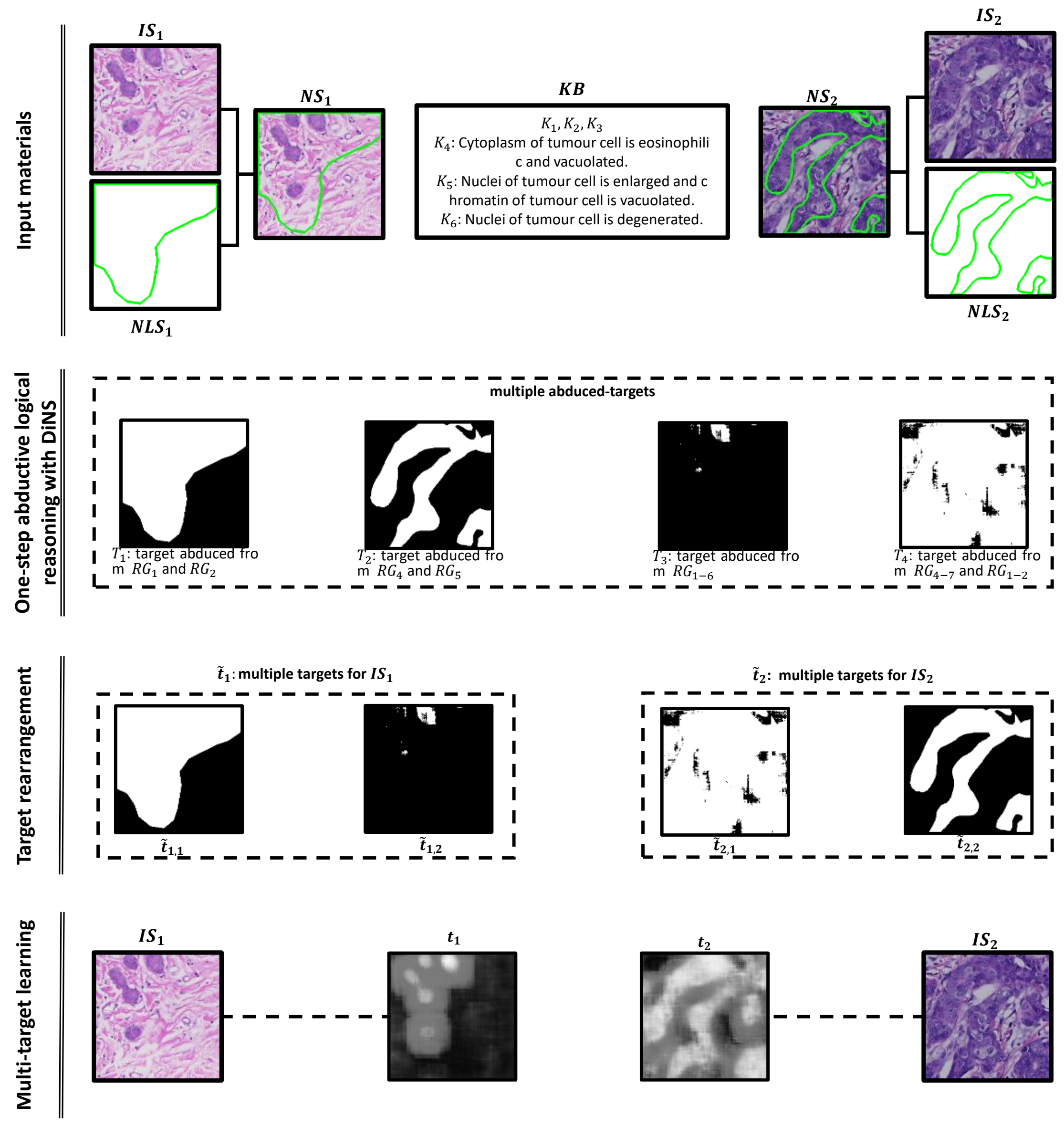

Figure. 7. The simplified outline for the OSAMTL-DiNS solution implemented for residual TSfBC in HE-stained post-treatment surgical resection images referring to Fig. 5 and Fig. 6.

We first respectively conducted experiments of various existing solutions that learn from complex noisy labels and experiments of various existing solutions that introduce OSAMTL-DiNS for enhancement in handling complex noisy labels. Then, we compare the results of these two series of approaches to show the contributions of OSAMTL-DiNS in handling complex noisy labels. In addition, we also provide some qualitative example results as visualized proofs to show the contributions of OSAMTL-DiNS in handling complex noisy labels.

For all experiments, we first use a training dataset to learn segmentation models, then we use a validation dataset to select the model for testing. The experimental results of segmentation models are averaged on corresponding dataset. The preparation for the training, validation and testing is described in section 2.4.2.

### 2.4.2 Dataset preparation

For the tumour segmentation task in HE-stained pre-treatment biopsy images, we collected a total amount of 144 WSIs. Among the collected WSIs, 105 WSIs were used by pathology expert to produce noisy sample one ($NS_1$ shown in the first row of Fig. 3.A), 20 WSIs were used by pathology expert to produce noisy sample two ($NS_2$ shown in the second row of Fig. 3.A), and the rest 19 WSIs were used by pathology expert to produce a relatively noisy-free sample ($RNFS$). Both $NS_1$and $NS_2$ were prepared for training without AGTLs, and $RNFS$ was prepared for validation and testing with AGTLs. $NS_1$ contains 992 pairs of images and corresponding noisy labels for training without AGTLs. $NS_2$ contains 142 pairs of images and corresponding noisy labels for training without AGTLs. $RNFS$ contains 158 pairs of images and corresponding accurate labels, among which 79 pairs are for validation and 79 pairs are for testing with AGTLs.

Similarly, for the tumour segmentation task in HE-stained post-treatment surgical resection images, we collected a total amount of 126 WSIs. Among the collected WSIs, 94 WSIs were used by pathology expert to produce noisy sample one ($NS_1$ shown in the first row of Fig. 3.B), 20 WSIs were used by pathology expert to produce noisy sample two ($NS_2$ shown in the second row of Fig. 3.B), and the rest 12 WSIs were used by pathology expert to produce a relatively noisy-free sample ($RNFS$). Both $NS_1$and $NS_2$ were prepared for training without AGTLs. $NS_1$ contains 2944 pairs of images and corresponding noisy labels for training without AGTLs. $NS_2$ contains 1431 pairs of images and corresponding noisy labels for training without AGTLs. $RNFS$ contains 242 pairs of images and corresponding accurate labels, among which 121 pairs are for validation and 121 pairs are for testing with AGTLs.

The image patches prepared for experiments were cropped at 10× magnification of the given WSIs and the size of each cropped image patch was at 256 × 256 pixels (width × height) to remain distinguishable morphologies of tumour.

**2.4.3 Evaluation metrics**

We employed usual metrics for image semantic segmentation evaluation. Let TP (true positive) be the number of pixels correctly predicted to belong to the H. pylori class, FP (false positive) be the number of pixels incorrectly predicted to belong to the H. pylori class, and FN (false negative) be the number of missing pixels predicted to belong to the background class. These metrics are tightly related to the foreground class which we are interested the most. Based on TP, FP and FN, we further employed precision, recall, f1 and foreground intersection over union (fIoU) for evaluation of the performance for image segmentation. These evaluation metrics were chosen for evaluation because they are most commonly used for evaluating image segmentation.

**2.4.4 Experimental details**

All of our experiments were performed on an Intel core Xeon E5–2630 v4s with a memory capacity of 128GB and eight NVIDIA GTX 1080Ti GPUs. Our developing environment is based on Tensorflow 1.10.1 and Python 3.5.

We started the training procedures of the employed DCNN for the learning model of tumour segmentation with the same initialization and hyperparameters including Adam (Kingma & Ba, 2015) selected as the optimizer, batch size set to 32, learning rate set to 0.0001, and online augmentations involving vertical and horizontal flips and random brightness. These hyperparameters were chosen because they have been the default setting in many opensource developing libraries of deep learning.

Various existing approaches for learning from noisy labels (LfNLs) (Frénay & Verleysen, 2014; Karimi et al., 2020; Song et al., 2022), including naively learning from noisy labels (BaseLine), Forward, Backward (Patrini et al., 2017), Boost-Hard, Boost-Soft (Arazo et al., 2019; Reed et al., 2015), D2L (Ma et al., 2018), SCE (Wang et al., 2019), Peer (Liu & Guo, 2020), DT-Forward (Yao et al., 2020), and NCE-SCE (Ma et al., 2020), were chosen for experimental investigations. These approaches were chosen due to their flexibility to be applied to the situation where no clean dataset is available, the targeted object cannot be clearly defined, and any of the given labels cannot be confidently regarded as clean. We respectively set the hyper parameters of these approaches as suggested by corresponding papers.

In default, we set the weights for the multi-task learning procedure of OSAMTL-DiNS to fifty-fifty in this TSfBC case, considering that both targets are equally important. When calculating the evaluation metrics, we used 0.5 to thresh the logits of the optimized DCNN for segmentation, as it is a default value to separate the predictions into target and non-target which can balance the bias and variance of the optimized DCNN.

## 3. Results

We show the results of various existing methods for learning from noisy labels (LfNLs) and the results of various existing methods with OSAMTL-DiNS introduced for LfNLs respectively on the two tumour segmentation tasks for breast cancer, respectively in sections 3.1 and 3.2. Based on the results presented in sections 3.1 and 3.2, we further show the contributions of OSAMTL-DiNS in handling complex noisy labels in section 3.3. On the basis of the contents indicated in section 3.3, we show the generalization contributions of OSAMTL-DiNS from validation to testing in section 3.4. In section 3.5, we give some qualitative testing results to show the contributions of OSAMTL-DiNS. Finally in section 3.6, we give information of a released predictive model pre-trained with OSAMTL-DiNS for tumour segmentation in HE-stained pre-treatment biopsy images in breast cancer and its application as a pre-processing tool to extract tumour-associated stroma compartment for predicting the pathological complete response to neoadjuvant chemotherapy in breast cancer (F. Li et al., 2022).

### *3.1 Results of various existing methods for LfNLs*

For simplicity, we denote Task1 as the tumour segmentation task in HE-stained pre-treatment biopsy images and Task2 as the tumour segmentation task in HE-stained post-treatment surgical resection images. In this subsection, we respectively show the experimental evaluations of various existing approaches for LfNLs mentioned in section 5.1.3 on Task1 and Task2.

#### 3.1.1 Evaluations on Task1 with AGTLs

We evaluate the results of employing various existing methods for LfNLs on Task1 using usual evaluations with AGTLs. Experimental statistics for validation and testing are respectively shown as Table 6 and Table 7.

Table 6. Evaluations of various existing method for validation on Task1

| Solution | TP | FP | FN | precision | recall | f1 | fIoU |
|---|---|---|---|---|---|---|---|
| BaseLine | 19278 | 15955 | 2807 | 54.72 | 87.29 | 67.27 | 50.68 |
| Forward | 20243 | 16860 | 1843 | 54.56 | 91.66 | 68.40 | 51.98 |
| Backward | 19069 | 14239 | 3016 | 57.25 | 86.34 | 68.85 | 52.50 |
| Boost-Hard | 18806 | 14959 | 3279 | 55.70 | 85.15 | 67.34 | 50.77 |
| Boost-Soft | 20138 | 17912 | 1948 | 52.93 | 91.18 | 66.97 | 50.35 |
| D2l | 19676 | 17020 | 2410 | 53.62 | 89.09 | 66.95 | 50.31 |
| SCE | 19500 | 15153 | 2585 | 56.27 | 88.30 | 68.74 | 52.37 |
| Peer | 19775 | 14660 | 2310 | 57.43 | 89.54 | 69.98 | 53.82 |
| DT-Forward | 19806 | 16229 | 2280 | 54.96 | 89.68 | 68.15 | 51.69 |
| NCE-SCE | 20015 | 16580 | 2071 | 54.69 | 90.62 | 68.22 | 51.76 |

Table 7. Evaluations of various existing method for testing on Task1

| Solution | TP | FP | FN | precision | recall | f1 | fIoU |
|---|---|---|---|---|---|---|---|
| BaseLine | 22707 | 13298 | 3249 | 63.07 | 87.48 | 73.29 | 57.85 |
| Forward | 23494 | 15160 | 2462 | 60.78 | 90.51 | 72.73 | 57.14 |
| Backward | 21858 | 13453 | 4098 | 61.90 | 84.21 | 71.35 | 55.46 |
| Boost-Hard | 22184 | 12652 | 3771 | 63.68 | 85.47 | 72.98 | 57.46 |
| Boost-Soft | 23724 | 15849 | 2231 | 59.95 | 91.40 | 72.41 | 56.75 |
| D2l | 23068 | 14632 | 2888 | 61.19 | 88.87 | 72.48 | 56.83 |
| SCE | 22753 | 13499 | 3203 | 62.76 | 87.66 | 73.15 | 57.67 |
| Peer | 22658 | 12704 | 3298 | 64.07 | 87.29 | 73.90 | 58.61 |
| DT-Forward | 23280 | 14239 | 2676 | 62.05 | 89.69 | 73.35 | 57.92 |
| NCE-SCE | 23395 | 14452 | 2561 | 61.81 | 90.13 | 73.34 | 57.90 |

### 3.1.2 Evaluations on Task2 with AGTLs

We evaluate the results of employing various existing methods for LfNLs on Task2 using usual evaluations with AGTLs. Experimental statistics for validation and testing are respectively shown as Table 8 and Table 9.

Table 8. Evaluations of various existing method for validation on Task2

| Solution | TP | FP | FN | precision | recall | f1 | fIoU |
|---|---|---|---|---|---|---|---|
| BaseLine | 18507 | 12766 | 11120 | 59.18 | 62.47 | 60.78 | 43.66 |
| Forward | 18176 | 11995 | 11451 | 60.24 | 61.35 | 60.79 | 43.67 |
| Backward | 19632 | 14650 | 9995 | 57.27 | 66.26 | 61.44 | 44.34 |
| Boost-Hard | 20238 | 14820 | 9389 | 57.73 | 68.31 | 62.57 | 45.53 |
| Boost-Soft | 23277 | 17557 | 6351 | 57.00 | 78.56 | 66.07 | 49.33 |
| D2l | 19617 | 12892 | 10010 | 60.34 | 66.21 | 63.14 | 46.13 |
| SCE | 18807 | 12118 | 10821 | 60.81 | 63.48 | 62.12 | 45.05 |
| Peer | 21279 | 18131 | 8349 | 53.99 | 71.82 | 61.64 | 44.55 |
| DT-Forward | 18509 | 13330 | 11119 | 58.13 | 62.47 | 60.22 | 43.09 |
| NCE-SCE | 19801 | 14619 | 9826 | 57.53 | 66.83 | 61.83 | 44.75 |

Table 9. Evaluations of various existing method for testing on Task2

| Solution | TP | FP | FN | precision | recall | f1 | fIoU |
|---|---|---|---|---|---|---|---|
| BaseLine | 15446 | 13831 | 8467 | 52.76 | 64.59 | 58.08 | 40.92 |
| Forward | 15129 | 13409 | 8783 | 53.01 | 63.27 | 57.69 | 40.54 |
| Backward | 16373 | 17083 | 7540 | 48.94 | 68.47 | 57.08 | 39.94 |
| Boost-Hard | 16599 | 15904 | 7313 | 51.07 | 69.42 | 58.85 | 41.69 |
| Boost-Soft | 19000 | 18353 | 4912 | 50.87 | 79.46 | 62.03 | 44.95 |
| D2l | 16331 | 14876 | 7581 | 52.33 | 68.30 | 59.26 | 42.10 |
| SCE | 15604 | 13286 | 8309 | 54.01 | 65.25 | 59.10 | 41.95 |
| Peer | 17366 | 19348 | 6546 | 47.30 | 72.62 | 57.29 | 40.14 |
| DT-Forward | 15374 | 15525 | 8538 | 49.76 | 64.29 | 56.10 | 38.98 |
| NCE-SCE | 16356 | 16574 | 7556 | 49.67 | 68.40 | 57.55 | 40.40 |

## *3.2 Results of various existing methods with OSAMTL-DiNS introduced for LfNLs*

In this subsection, we respectively give experimental evaluations of OSAMTL-DiNS introduced to various existing approaches for LfNLs mentioned in section 5.1.3 on Task1 and Task2.

### 3.2.1 Evaluations on Task1

We evaluate the results of employing various existing methods with OSAMTL-DiNS introduced for LfNLs on Task1 using usual evaluations with AGTLs. Experimental statistics for validation and testing are respectively shown as Table 10 and Table 11.

Table 10. Evaluations of various existing method with OSAMTL-DiNS introduced for validation on Task1

| Solution | TP | FP | FN | precision | recall | f1 | fIoU |
|---|---|---|---|---|---|---|---|
| BaseLine | 18026 | 8651 | 4059 | 67.57 | 81.62 | 73.93 | 58.65 |
| Forward | 17266 | 7505 | 4819 | 69.70 | 78.18 | 73.70 | 58.35 |
| Backward | 17814 | 8371 | 4272 | 68.03 | 80.66 | 73.81 | 58.49 |
| Boost-Hard | 17348 | 7606 | 4738 | 69.52 | 78.55 | 73.76 | 58.43 |
| Boost-Soft | 17855 | 8059 | 4231 | 68.90 | 80.84 | 74.40 | 59.23 |
| D2l | 17597 | 8128 | 4489 | 68.40 | 79.67 | 73.61 | 58.24 |
| SCE | 16774 | 7302 | 5311 | 69.67 | 75.95 | 72.68 | 57.08 |
| Peer | 17681 | 8557 | 4404 | 67.39 | 80.06 | 73.18 | 57.70 |
| DT-Forward | 17218 | 7393 | 4868 | 69.96 | 77.96 | 73.74 | 58.41 |
| NCE-SCE | 16305 | 6562 | 5781 | 71.30 | 73.83 | 72.54 | 56.91 |

Table 11. Evaluations of various existing method with OSAMTL-DiNS introduced for testing on Task1

| Solution | TP | FP | FN | precision | recall | f1 | fIoU |
|---|---|---|---|---|---|---|---|
| BaseLine | 21010 | 6381 | 4946 | 76.70 | 80.94 | 78.77 | 64.97 |
| Forward | 20215 | 5579 | 5740 | 78.37 | 77.88 | 78.13 | 64.11 |
| Backward | 20818 | 6124 | 5137 | 77.27 | 80.21 | 78.71 | 64.90 |
| Boost-Hard | 20230 | 5732 | 5725 | 77.92 | 77.94 | 77.93 | 63.84 |
| Boost-Soft | 20657 | 5936 | 5298 | 77.68 | 79.59 | 78.62 | 64.77 |
| D2l | 20348 | 5981 | 5608 | 77.28 | 78.39 | 77.83 | 63.71 |
| SCE | 19719 | 5651 | 6236 | 77.73 | 75.97 | 76.84 | 62.39 |
| Peer | 20379 | 6634 | 5577 | 75.44 | 78.51 | 76.95 | 62.53 |
| DT-Forward | 19958 | 5347 | 5998 | 78.87 | 76.89 | 77.87 | 63.76 |
| NCE-SCE | 18712 | 4594 | 7244 | 80.29 | 72.09 | 75.97 | 61.25 |

### 3.2.2 Evaluations on Task2

We evaluate the results of various existing methods with OSAMTL-DiNS introduced for LfNLs on Task2 using usual evaluations with AGTLs. Experimental statistics for validation and testing are respectively shown as Table 12 and Table 13.

Table 12. Evaluations of various existing method with OSAMTL-DiNS introduced for validation on Task2

| Solution | TP | FP | FN | precision | recall | f1 | fIoU |
|---|---|---|---|---|---|---|---|
| BaseLine | 19291 | 4878 | 10337 | 79.82 | 65.11 | 71.72 | 55.91 |
| Forward | 18257 | 3530 | 11370 | 83.80 | 61.62 | 71.02 | 55.06 |
| Backward | 18966 | 4648 | 10661 | 80.32 | 64.02 | 71.25 | 55.33 |
| Boost-Hard | 18990 | 3964 | 10637 | 82.73 | 64.10 | 72.23 | 56.53 |
| Boost-Soft | 18850 | 5732 | 10777 | 76.68 | 63.62 | 69.55 | 53.31 |
| D2l | 18438 | 3154 | 11190 | 85.39 | 62.23 | 71.99 | 56.24 |
| SCE | 18449 | 5022 | 11178 | 78.60 | 62.27 | 69.49 | 53.24 |
| Peer | 20269 | 6413 | 9358 | 75.97 | 68.41 | 71.99 | 56.24 |
| DT-Forward | 18371 | 4117 | 11256 | 81.69 | 62.01 | 70.50 | 54.44 |
| NCE-SCE | 16177 | 2620 | 13451 | 86.06 | 54.60 | 66.81 | 50.16 |

Table 13. Evaluations of various existing method with OSAMTL-DiNS introduced for testing on Task2

| Solution | TP | FP | FN | precision | recall | f1 | fIoU |
|---|---|---|---|---|---|---|---|
| BaseLine | 16000 | 5649 | 7912 | 73.91 | 66.91 | 70.24 | 54.13 |
| Forward | 14825 | 3948 | 9088 | 78.97 | 62.00 | 69.46 | 53.21 |
| Backward | 15441 | 5648 | 8471 | 73.22 | 65.57 | 68.62 | 52.24 |
| Boost-Hard | 15713 | 4611 | 8200 | 77.31 | 65.71 | 71.04 | 55.09 |
| Boost-Soft | 15799 | 6017 | 8114 | 72.42 | 66.07 | 69.10 | 52.79 |
| D2l | 15109 | 3599 | 8803 | 80.76 | 63.18 | 70.90 | 54.92 |
| SCE | 15168 | 5151 | 8744 | 74.65 | 63.43 | 68.59 | 52.19 |
| Peer | 16954 | 7478 | 6958 | 69.39 | 70.90 | 70.14 | 54.01 |
| DT-Forward | 15175 | 4483 | 8737 | 77.20 | 63.46 | 69.66 | 53.44 |
| NCE-SCE | 13101 | 2749 | 10811 | 82.66 | 54.79 | 65.90 | 49.14 |

## *3.3 Contributions of OSAMTL-DiNS in handling complex noisy labels*

In this subsection, we show the contributions of OSAMTL-DiNS in handling complex noisy labels on Task1 and Task2.

### 3.3.1 Contributions of OSAMTL-DiNS on Task1

We reflect the contributions of OSAMTL-DiNS on Task1 by the differences between Table 6 and Table 10 for validation and the differences between Table 7 and Table 11 for testing, which are respectively shown as Fig. 8 and Fig. 9. More specifically, the contributions of OSAMTL-DiNS on Task1 are quantitatively evaluated by statistics of Table 10 minus statistics of Table 6 for validation and statistics of Table 11 minus statistics of Table 7 for testing, which are respectively shown as Table 14 and Table 15. And the confident intervals for the contributions of OSAMTL-DiNS on Task1 are shown in Table 16.

Table 14. Contributions of OSAMTL-DiNS to various existing method for validation on Task1

| Solution | TP | FP | FN | precision | recall | f1 | fIoU |
|---|---|---|---|---|---|---|---|
| BaseLine | -1252 | -7304 | 1252 | 12.85 | -5.67 | 6.66 | 7.97 |
| Forward | -2977 | -9355 | 2976 | 15.14 | -13.48 | 5.3 | 6.37 |
| Backward | -1255 | -5868 | 1256 | 10.78 | -5.68 | 4.96 | 5.99 |
| Boost-Hard | -1458 | -7353 | 1459 | 13.82 | -6.6 | 6.42 | 7.66 |
| Boost-Soft | -2283 | -9853 | 2283 | 15.97 | -10.34 | 7.43 | 8.88 |
| D2l | -2079 | -8892 | 2079 | 14.78 | -9.42 | 6.66 | 7.93 |
| SCE | -2726 | -7851 | 2726 | 13.4 | -12.35 | 3.94 | 4.71 |
| Peer | -2094 | -6103 | 2094 | 9.96 | -9.48 | 3.2 | 3.88 |
| DT-Forward | -2588 | -8836 | 2588 | 15 | -11.72 | 5.59 | 6.72 |
| NCE-SCE | -3710 | -10018 | 3710 | 16.61 | -16.79 | 4.32 | 5.15 |

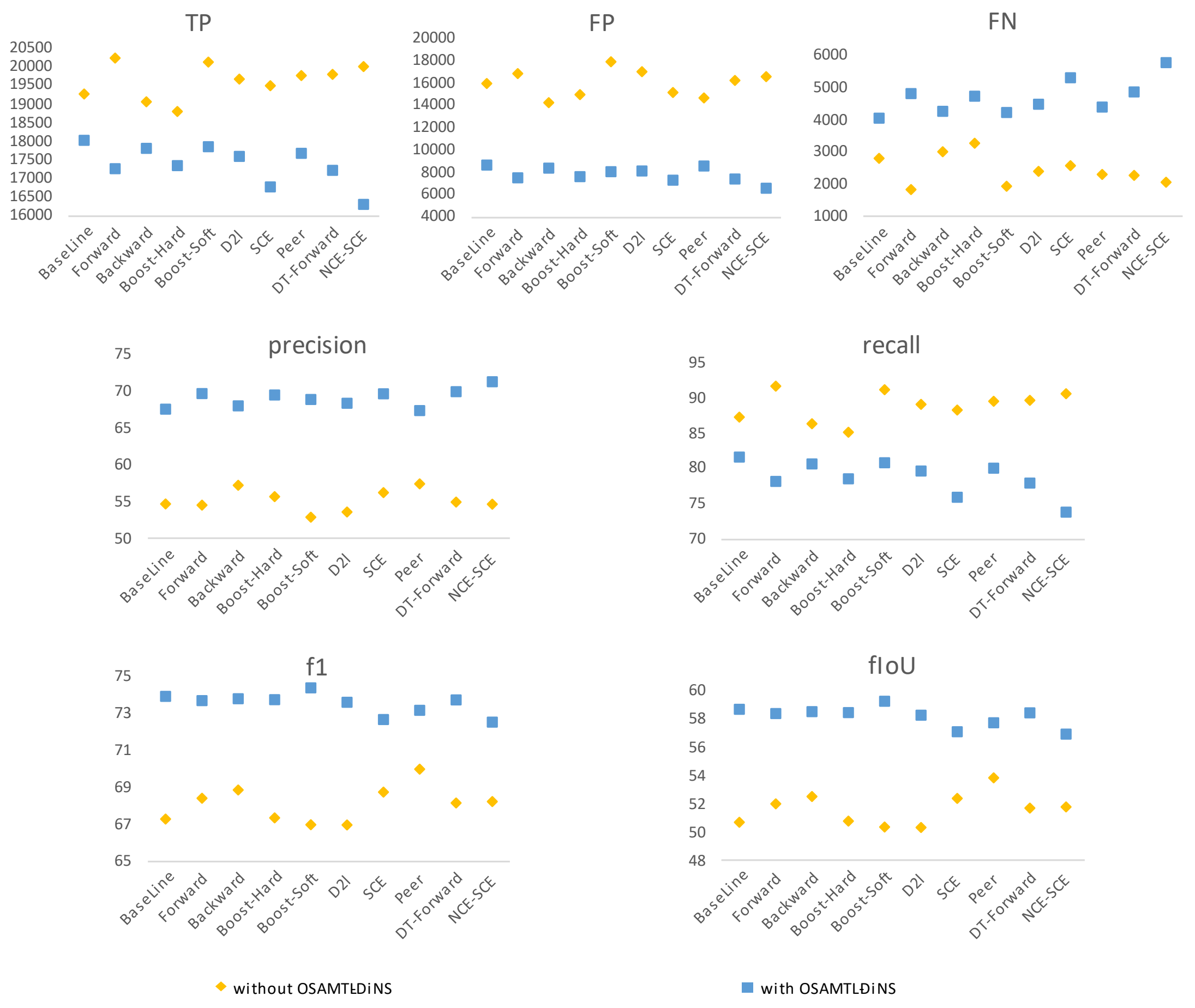

Fig. 8. Differences between Table 6 and Table 10 for validation of Task1.

Table 15. Contributions of OSAMTL-DiNS to various existing method for testing on Task1

| Solution | TP | FP | FN | precision | recall | f1 | fIoU |
|---|---|---|---|---|---|---|---|
| BaseLine | -1697 | -6917 | 1697 | 13.63 | -6.54 | 5.48 | 7.12 |
| Forward | -3279 | -9581 | 3278 | 17.59 | -12.63 | 5.4 | 6.97 |
| Backward | -1040 | -7329 | 1039 | 15.37 | -4 | 7.36 | 9.44 |
| Boost-Hard | -1954 | -6920 | 1954 | 14.24 | -7.53 | 4.95 | 6.38 |
| Boost-Soft | -3067 | -9913 | 3067 | 17.73 | -11.81 | 6.21 | 8.02 |
| D2l | -2720 | -8651 | 2720 | 16.09 | -10.48 | 5.35 | 6.88 |
| SCE | -3034 | -7848 | 3033 | 14.97 | -11.69 | 3.69 | 4.72 |
| Peer | -2279 | -6070 | 2279 | 11.37 | -8.78 | 3.05 | 3.92 |
| DT-Forward | -3322 | -8892 | 3322 | 16.82 | -12.8 | 4.52 | 5.84 |
| NCE-SCE | -4683 | -9858 | 4683 | 18.48 | -18.04 | 2.63 | 3.35 |

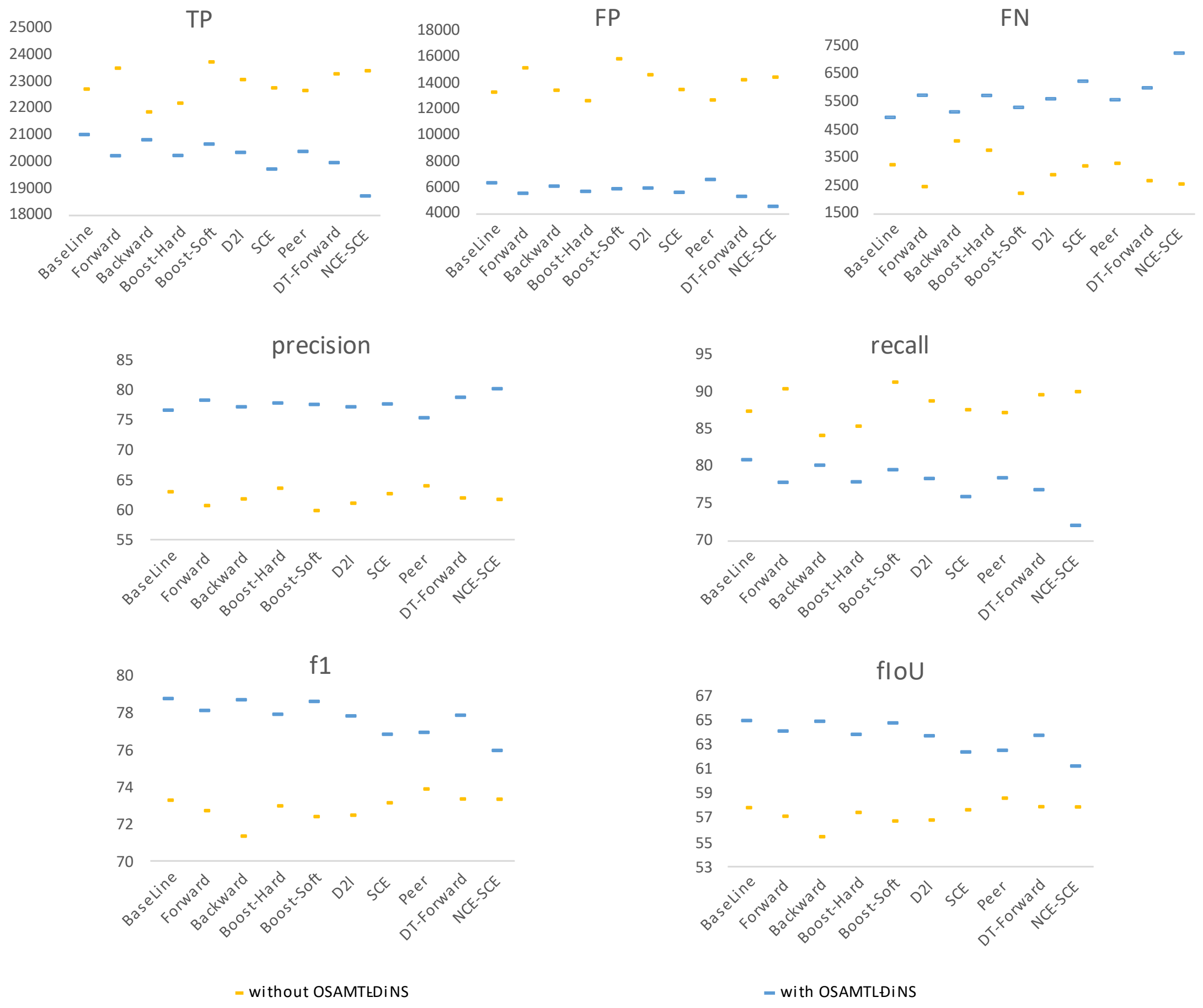

Fig. 9. Differences between Table 7 and Table 11 for testing of Task1.

Table 16. Confident intervals for contributions of OSAMTL-DiNS to various existing method on Task1

| Dataset | | TP | FP | FN | precision | recall | f1 | fIoU |
|---|---|---|---|---|---|---|---|---|
| Validation | lower limit | -2810 | -9202 | 1675 | 12.29 | -12.72 | 4.47 | 5.37 |
| | mean | -2242 | -8143 | 2242 | 13.83 | -10.15 | 5.45 | 6.53 |
| | upper limit | -1674 | -7085 | 2810 | 15.37 | -7.58 | 6.42 | 7.68 |
| Testing | lower limit | -3099 | -9027 | 1876 | 13.74 | -11.94 | 4.13 | 5.32 |
| | mean | -2488 | -8013 | 2488 | 15.31 | -9.58 | 5.11 | 6.59 |
| | upper limit | -1877 | -7000 | 3099 | 16.89 | -7.23 | 6.10 | 7.86 |

### 3.3.2 Contributions of OSAMTL-DiNS on Task2

Similarly, we reflect the contributions of OSAMTL-DiNS on Task2 by the differences between Table 8 and Table 12 for validation and the differences between Table 9 and Table 13 for testing, which are respectively shown as Fig. 10 and Fig. 11. More specifically, the contributions of OSAMTL-DiNS on Task1 are quantitatively evaluated by statistics of Table 12 minus statistics of Table 8 for validation and statistics of Table 13 minus statistics of Table 9 for testing, which are respectively shown as Table 17 and Table 18. And the confident intervals for the contributions of OSAMTL-DiNS on Task2 are shown in Table 19.

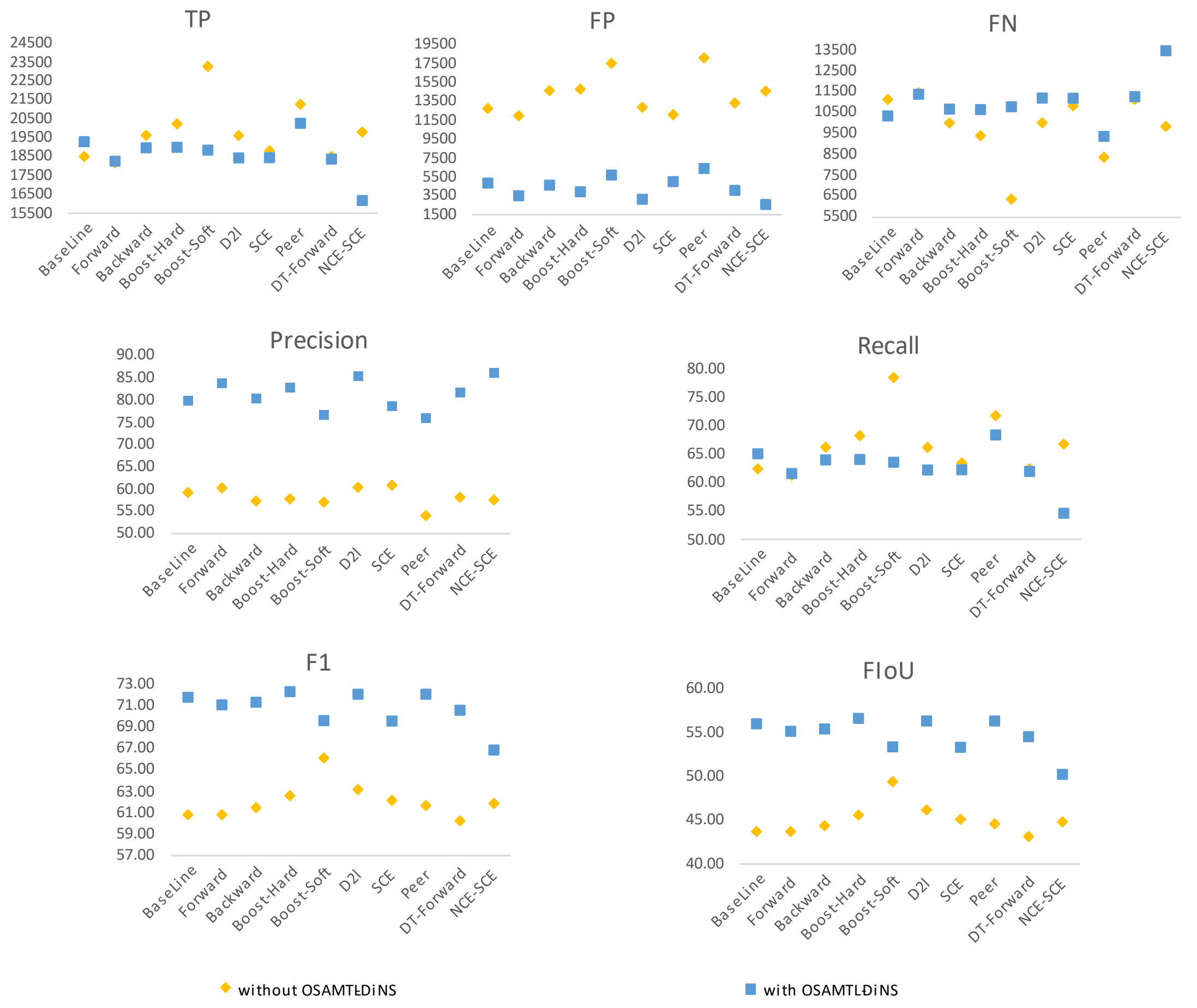

Fig. 10. Differences between Table 8 and Table 12 for validation of Task2.

Table 17. Contributions of OSAMTL-DiNS to various existing method for validation on Task2

| Solution | TP | FP | FN | precision | recall | f1 | fIoU |
|---|---|---|---|---|---|---|---|
| BaseLine | 784 | -7888 | -783 | 20.64 | 2.64 | 10.94 | 12.25 |
| Forward | 81 | -8465 | -81 | 23.56 | 0.27 | 10.23 | 11.39 |
| Backward | -666 | -10002 | 666 | 23.05 | -2.24 | 9.81 | 10.99 |
| Boost-Hard | -1248 | -10856 | 1248 | 25.00 | -4.21 | 9.66 | 11.00 |
| Boost-Soft | -4427 | -11825 | 4426 | 19.68 | -14.94 | 3.48 | 3.98 |
| D2l | -1179 | -9738 | 1180 | 25.05 | -3.98 | 8.85 | 10.11 |
| SCE | -358 | -7096 | 357 | 17.79 | -1.21 | 7.37 | 8.19 |
| Peer | -1010 | -11718 | 1009 | 21.98 | -3.41 | 10.35 | 11.69 |
| DT-Forward | -138 | -9213 | 137 | 23.56 | -0.46 | 10.28 | 11.35 |
| NCE-SCE | -3624 | -11999 | 3625 | 28.53 | -12.23 | 4.98 | 5.41 |

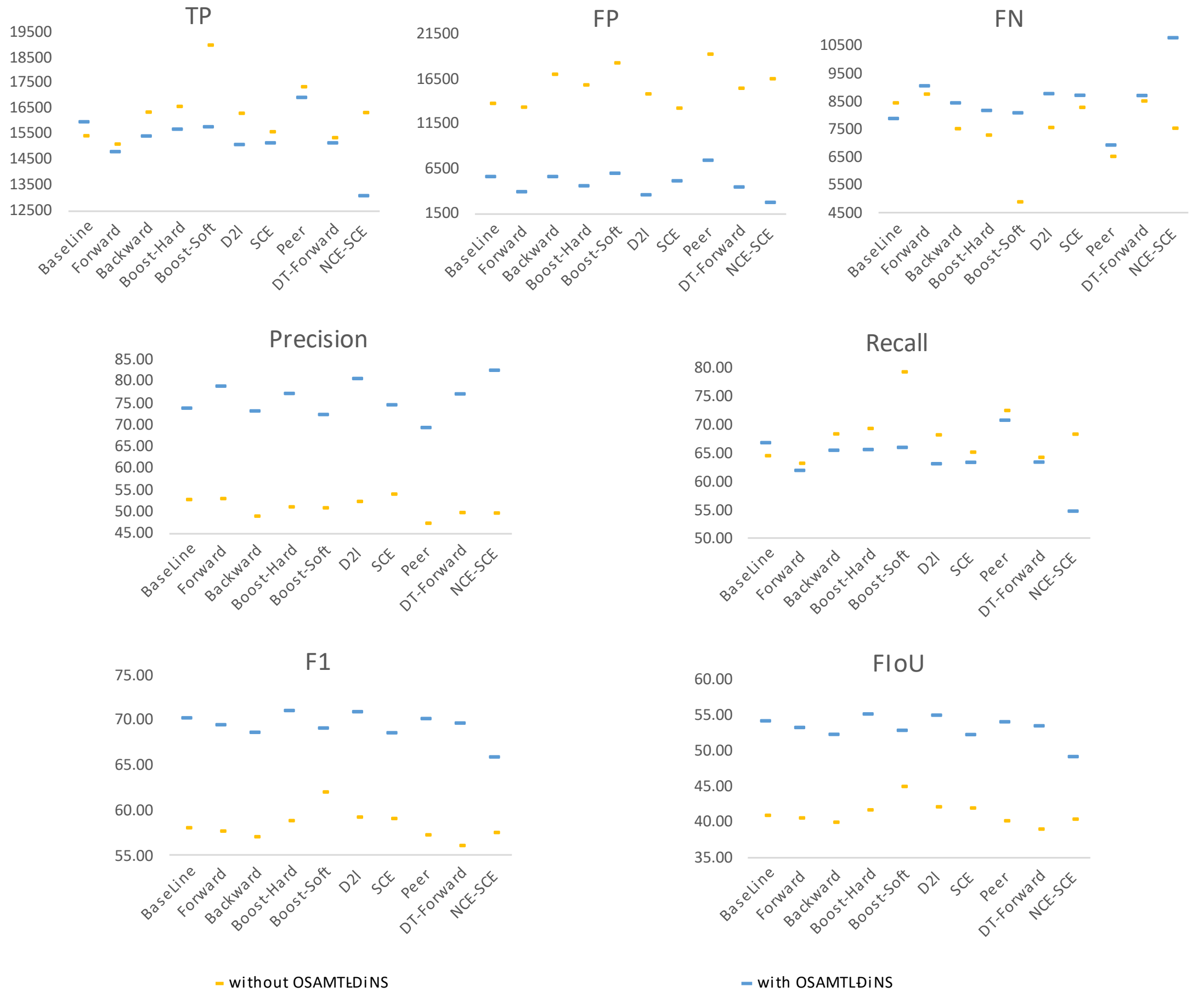

Fig. 11. Differences between Table 9 and Table 13 for testing of Task2.

Table 18. Contributions of OSAMTL-DiNS to various existing method for testing on Task2

| Solution | TP | FP | FN | precision | recall | f1 | fIoU |
|---|---|---|---|---|---|---|---|
| BaseLine | 554 | -8182 | -555 | 21.15 | 2.32 | 12.16 | 13.21 |
| Forward | -304 | -9461 | 305 | 25.96 | -1.27 | 11.77 | 12.67 |
| Backward | -932 | -11435 | 931 | 24.28 | -2.90 | 11.54 | 12.30 |
| Boost-Hard | -886 | -11293 | 887 | 26.24 | -3.71 | 12.19 | 13.40 |
| Boost-Soft | -3201 | -12336 | 3202 | 21.55 | -13.39 | 7.07 | 7.84 |
| D2l | -1222 | -11277 | 1222 | 28.43 | -5.12 | 11.64 | 12.82 |
| SCE | -436 | -8135 | 435 | 20.64 | -1.82 | 9.49 | 10.24 |
| Peer | -412 | -11870 | 412 | 22.09 | -1.72 | 12.85 | 13.87 |
| DT-Forward | -199 | -11042 | 199 | 27.44 | -0.83 | 13.56 | 14.46 |
| NCE-SCE | -3255 | -13825 | 3255 | 32.99 | -13.61 | 8.35 | 8.74 |

Table 19. Confident intervals for contributions of OSAMTL-DiNS to various existing method on Task2

| Dataset | | TP | FP | FN | precision | recall | f1 | fIoU |
|---|---|---|---|---|---|---|---|---|
| Validation | lower limit | -2347 | -11115 | 10 | 20.70 | -7.92 | 6.79 | 7.60 |
| | mean | -1179 | -9880 | 1178 | 22.88 | -3.98 | 8.60 | 9.64 |
| | upper limit | -10 | -8645 | 2347 | 25.07 | -0.03 | 10.40 | 11.67 |
| Testing | lower limit | -1928 | -12176 | 131 | 22.27 | -7.98 | 9.58 | 10.35 |
| | mean | -1029 | -10886 | 1029 | 25.08 | -4.21 | 11.06 | 11.96 |
| | upper limit | -131 | -9596 | 1928 | 27.89 | -0.43 | 12.55 | 13.56 |

### *3.4 Generalization contributions of OSAMTL-DiNS from validation to testing*

In this subsection, we show the generalization contributions of OSAMTL-DiNS by statistics between the contributions of OSAMTL-DiNS on validation and the contributions of OSAMTL-DiNS on testing. The results for Task1 and Task2 are respectively shown as Table 20 and Table 21.

Table 20. Statistics between the contributions of OSAMTL-DiNS on validation and the contributions of OSAMTL-DiNS on testing for Task 1

| Statistics | | TP | FP | FN | precision | recall | f1 | fIoU |
|---|---|---|---|---|---|---|---|---|
| Mean | validation | -2242 | -8143 | 2242 | 13.83 | -10.15 | 5.45 | 6.53 |
| | testing | -2708 | -8198 | 2707 | 15.63 | -10.43 | 4.86 | 6.26 |
| P-value | | 0.270 | 0.933 | 0.271 | 0.080 | 0.871 | 0.364 | 0.741 |
| Similarity | | 0.960 | 0.932 | 0.960 | 0.862 | 0.960 | 0.628 | 0.645 |

Table 21. Statistics between the contributions of OSAMTL-DiNS on validation and the contributions of OSAMTL-DiNS on testing for Task 2

| Statistics | | TP | FP | FN | precision | recall | f1 | fIoU |
|---|---|---|---|---|---|---|---|---|
| Mean | validation | -1179 | -9880 | 1178 | 22.88 | -3.98 | 8.60 | 9.64 |
| | testing | -1029 | -10886 | 1029 | 25.08 | -4.21 | 11.06 | 11.96 |
| P-value | | 0.821 | 0.219 | 0.822 | 0.180 | 0.926 | 0.028 | 0.058 |
| Similarity | | 0.976 | 0.936 | 0.976 | 0.944 | 0.979 | 0.957 | 0.957 |

### *3.5 Qualitative testing results*

Some typical testing results of various state-of-the-art approaches (respectively without OSAMTL-DiNS introduced and with OSAMTL-DiNS introduced) for handling complex noisy labels on Task1 and Task2 are respectively shown as Fig. 12 and Fig. 13. More testing results are provided in Supplementary 3.

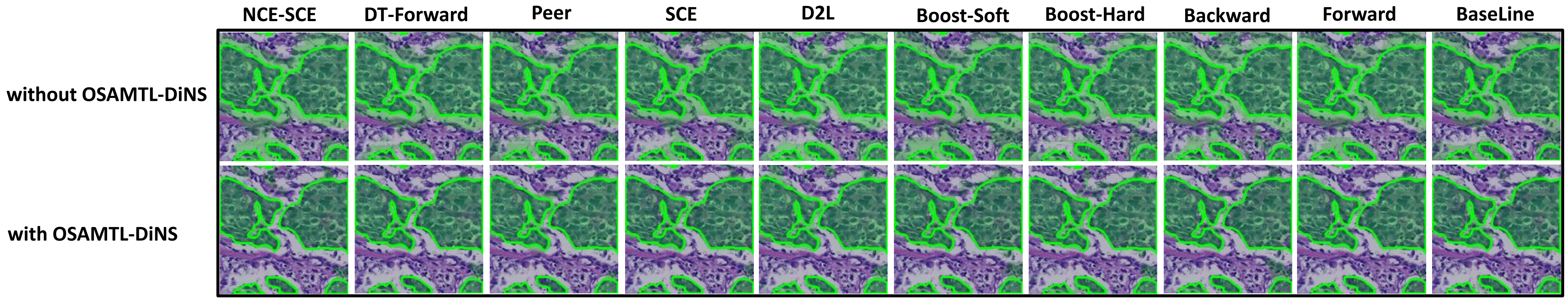

Figure. 12. Typical testing results of various state-of-the-art approaches (respectively without OSAMTL-DiNS introduced and with OSAMTL-DiNS introduced) for handling complex noisy labels on Task1. Top row: testing results of various state-of-the-art approaches without OSAMTL-DiNS introduced. Bottom row: testing results of state-of-the-art approaches with OSAMTL-DiNS introduced. The green solid polygons are the relatively accurately labelled tumour areas provided by pathology experts and the green transparent masks are the tumour areas predicted by image semantic segmentation models.

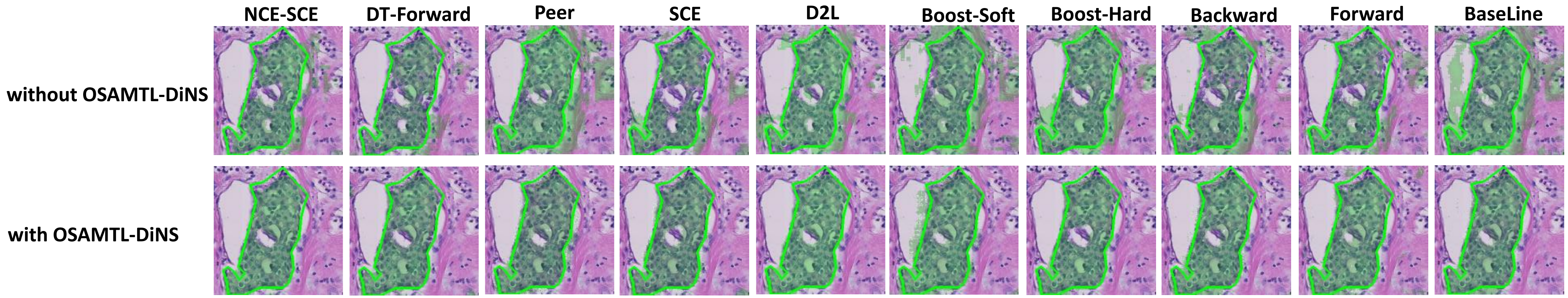

Fig. 13. Typical testing results of various state-of-the-art approaches (respectively without OSAMTL-DiNS introduced and with OSAMTL-DiNS introduced) for handling complex noisy labels on Task2. Top row: testing results of various state-of-the-art approaches without OSAMTL-DiNS introduced. Bottom row: testing results of state-of-the-art approaches with OSAMTL-DiNS introduced. The green solid polygons are the relatively accurately labelled tumour areas provided by pathology experts and the green transparent masks are the tumour areas predicted by image semantic segmentation models.

### *3.6 Released predictive model*

We released a predictive model pre-trained with OSAMTL-DiNS for tumour segmentation in HE-stained pre-treatment biopsy images in breast cancer. The released model was based on the BaseLine solution with OSAMTL-DiNS introduced, since it showed the best generalization performance for Task 1. The release model has also been successfully applied as a pre-processing tool to extract tumour-associated stroma compartment for predicting the pathological complete response to neoadjuvant chemotherapy in breast cancer. For more details about how the released predictive model was utilized, readers can refer to our paper (F. Li et al., 2022). The released model is available at https://github.com/YongQuanYang/TS-Score.

## 4 Discussion

In the application of the proposed OSAMTL-DiNS to the TSfBC, we primarily conducted two series of experiments: 1) simply using various state-of-the-art approaches for handling complex noisy labels to produce predictive models for TSfBC; 2) introducing OSAMTL-DiNS to various state-of-the-art approaches for handling complex noisy labels to produce predictive models for TSfBC. We compared the respective results of these two series of predictive models, which were obtained from the testing dataset with AGTLs provided by pathological experts. As the compared results of experiments are obtained from AGTLs which can be regarded as golden standards, the comparison between the two series of experiments can be effective to investigate the effectiveness of the proposed OSAMTL-DiNS in handling complex noisy labels for TSfBC. Further, the investigation of the effectiveness of OSAMTL-DiNS applied to TSfBC can also reflect the potentials of applying OSAMTL-DiNS to address specific challenges like TSfBC in MHWSIA, since the situation, where it is often difficult for experts to manually achieve the accurate ground-truth labels, is quite normal in MHWSIA.

Specifically, OSAMTL-DiNS was able to enable various state-of-the-art approaches for handling complex noisy labels to achieve significantly more rational predictions for Task1 of TSfBC on both validation and testing, by appropriately increasing the precision performance while reducing the recall performance (see respective metrics in Fig. 8, Fig. 9, Table 14 and Table 15). For the Task1 of TSfBC, the achieved mean improvements in overall performances of f1 and fIoU on validation were respectively 5.45%(4.47%, 6.42%) and from 6.53% (5.37%, 7.68%), while the achieved mean improvements in overall performances of f1 and fIoU on testing were respectively 5.11%(4.13%, 6.10%) and 6.59%(5.32%, 7.86%) (see statistics in Table 16). Identically, OSAMTL-DiNS was also able to enable various state-of-the-art approaches for handling complex noisy labels to achieve significantly more rational predictions for Task2 of TSfBC on both validation and testing, by appropriately increasing the precision performance while reducing the recall performance (see respective metrics in Fig. 10, Fig. 11, Table 17 and Table 18). For the Task2 of TSfBC, the achieved mean improvements in overall performances of f1 and fIoU on validation were respectively 8.60%(6.79%, 10.40%) and from 9.64% (7.60%, 11.67%), while the achieved mean improvements in overall performances of f1 and fIoU on testing were respectively 11.06%(9.58%, 12.55%) and 11.96%(10.35%, 13.56%) (see statistics in Table 19). As the experimental results were obtained from dataset with AGTLs provided by pathological experts, these contributions of OSAMTL-DiNS to various state-of-the-art approaches for handling complex noisy labels imply that OSAMTL-DiNS can help to improve the accuracy and consistency of the predictive model for TSfBC regarding to the results of pathological experts.

The statistics for the comparison between the contributions of OSAMTL-DiNS respective on validation and testing show that OSAMTL-DiNS can robustly improve the accuracy and consistency of the predictive model for TSfBC (see statistics in Table 20 and Table 21). The P-values of the statistics indicate that there are no significant differences between the contributions of OSAMTL-DiNS respectively on validation and testing, as all P-values are greater than 0.01 and most P-values are greater than 0.05. The similarity-values of the statistic indicate that the contributions of OSAMTL-DiNS respectively on validation and testing are quite analogical, as all similarity-values are greater than 0.62 and most similarity-values are greater than 0.95. As a result, the contributions of OSAMTL-DiNS on validation can be well generalized to testing, which implies the robustness of OSAMTL-DiNS in improving the accuracy and consistency of the predictive model for TSfBC.

The visualized testing results for the two series of experiments show that, the testing results of predictive models produced with OSAMTL-DiNS introduced, are much more accurate than the results of the predictive models simply produced with various state-of-the-art approaches as the former results have much more clean predictions compared with the labels provided by pathological experts (see Fig. 12 and Fig. 13). This visual comparison indicates that OSAMTL-DiNS can be helpful to produce predictive model that can achieve predictions visually more consistent with AGTLs provided by pathology experts for TSfBC.

In addition to the contributions of OSAMTL-DiNS to TSfBC, which reflect the potentials of OSAMTL-DiNS to address specific challenges like TSfBC in MHWSIA, the usage of the released model pre-trained with OSAMTL-DiNS as a pre-processing tool to extract tumour-associated stroma compartment for predicting the pathological complete response to neoadjuvant chemotherapy in breast cancer (F. Li et al., 2022) further proves that the proposed OSAMTL-DiNS has the potentials to facilitate medical image analytics in MHWSIA.

Although the results of the proposed OSAMTL-DiNS applied on TSfBC have shown favourable advantages, it is foreseeable the released predictive model pre-trained with OSAMTL-DiNS for tumour segmentation in HE-stained pre-treatment biopsy images in breast cancer will have limited generalization to some unseen testing HE-stained pre-treatment biopsy images in breast cancer. This is because only a very small amount of inaccurately labelled data was prepared and the machine learning architecture selected to produce the released predictive model was a relatively shallow deep learning architecture.

Notably, recently proposed large models (Kirillov et al., 2023) (mostly based on the machine learning architectures of transformers (Khan et al., 2022)) have appeared to be very effective for achieving the general purpose of image segmentation. However, recent reports have demonstrated that these large models for the general purpose of image segmentation have limited generalization to data of specific fields, such as medical images (Y. Huang et al., 2024) and images of concealed scenes (Ji et al., 2023). This situation proves that it is still worthwhile to produce relatively small models that are appropriate for particular image segmentation tasks in specific fields, though they definitely will be not as strong as large models to segment anything.

## 5 Conclusions and Future Work

In this paper, we proposed one-step abductive multi-target learning (OSAMTL) with diverse noisy samples (DiNS) (OSAMTL-DiNS) to expand the original OSAMTL to handle complex noisy labels of DiNS. We applied OSAMTL-DiNS to tumour segmentation for breast cancer (TSfBC) in medical histopathology whole slide image analysis (MHWSIA). Results of extensive experiments showed that OSAMTL-DiNS is able to enable various state-of-the-art approaches for learning from noisy labels to achieve more rational predictions for TSfBC. We also released a model pre-trained with OSAMTL-DiNS for tumour segmentation in HE-stained pre-treatment biopsy images in breast cancer.

In this paper, giving definition of diverse noisy samples (DiNS), we proposed one-step abductive multi-target learning with DiNS (OSAMTL-DiNS) that expands the original OSAMTL to handle complex noisy labels of DiNS. The major advantage of OSAMTL-DiNS is that it can produce a predictive model based on a chosen machine learning architecture simply using very inaccurately labelled (complex noisy) data, which makes OSAMTL-DiNS suitable to address some tasks in the field of medical analysis where the problem of complex noisy in data always exists.

Based on the proposed OSAMTL-DiNS, we implemented solutions for tumour segmentation for breast cancer in medical histopathology whole slide image analysis (MHWSIA) and conducted extensive experiments. Experiment results for contributions of OSAMTL-DiNS in handling complex noisy labels show that introducing OSAMTL-DiNS to existing various methods for learning from noisy labels can significantly enhance the abilities of these methods in handling complex noisy labels. At the meantime, experiment results for the generalization of OSAMTL-DiNS show that the contributions of OSAMTL-DiNS in handling complex noisy labels can be well generalized form validation to testing. Additionally, visualized qualitative results as well show that, with OSAMTL-DiNS introduced, various methods for learning from noisy labels can achieve predictions that are more consistent with the relatively accurate labels provided by experts. These results reflect the potential effectiveness of the proposed OSAMTL-DiNS in handling complex noisy labels in MHWSIA.

Based on the conducted experiments, we also chose to release a model pre-trained with OSAMTL-DiNS for tumour segmentation in HE-stained pre-treatment biopsy images in breast cancer. The released model has been successfully applied as a pre-processing tool to extract tumour-associated stroma compartment for predicting the pathological complete response to neoadjuvant chemotherapy in breast cancer, which further reflects the potentials of using OSAMTL-DiNS to build basic tools to facilitate medical image analytics in MHWSIA..

Although the application of OSAMTL-DiNS on TSfBC has reflected the potentials of the proposed OSAMTL-DiNS in handling complex noisy labels in MHWSIA, the more rational predictions of OSAMTL-DiNS in TSfBC was achieved by appropriately increase the precision performance while reducing the recall performance to obtain a better overall performance in fIoU, compared with various state-of-the-art approaches for handling complex noisy labels. As the recall performance for a target is very important to medical evaluations, in future works, we will continue, with limited annotations, to improve the solution for tumour segmentation of breast cancer to fulfil the goal that increases the recall performance while being able to promote or at least maintain the fIoU performance.

As the proposed OSAMTL-DiNS is independent from the machine learning architectures chosen for specific tasks and recently proposed large models have shown limited generalization to some tasks of specific fields, it is also interesting to combine OSAMTL-DiNS with very large models to enhance their generalization to particular tasks. In fact, a recent work, which employed weaker learners to supervise a pre-trained strong learner to enhance its generalization ability (Burns et al., 2023), has shown the promise of this idea. Intrinsically, the essence of the proposed OSAMTL-DiNS and the strategy of weak-to-strong generalization is identical, since the multiple inaccurate targets abduced from DiNS in this paper can be regarded as predictions of multiple weaker learners (see the details of how the multiple inaccurate targets are generated in the last paragraph of the section 2.3.4). The main difference is that we are focusing more on revealing the science behind the weak-to-strong generalization strategy, instead of chasing the state-of-the-art results. Our focus is to prove whether we can use multiple weak labels to validate a predictive model reasonably (Yang, 2024a), because if we can validate a predictive model with multiple weak labels then we can also use them to train a predictive model. We will continue this focus, as we believe it can scientifically contribute to part of the artificial intelligence alignment problem (Yang, 2023), especially when the state-of-the-art deep neural network of machine learning for building predictive model has been becoming standardized and reaching its limits in some specific AI applications (Yang, 2024b).

## Acknowledgments

This work was supported by the 1·3·5 project for disciplines of excellence (ZYGD18012) and the Technological Innovation Project of Chengdu New Industrial Technology Research Institute (2017-CY02–00026-GX). We really appreciate the very constructive comments received from the anonymous reviewers and editors of Expert Systems with Applications, which provided the authors with good opportunities to improve the quality of this paper. We acknowledge ZHONG JIU Flash Medical Technology Co., Ltd., Mianyang, China for providing some technical supports for revisions of this paper. The author, Yongquan Yang, especially thanks his Yang Family,

Chengdu, China for providing him with the financial supports and mental encouragements to do the major revisions during the publication processes of this paper.

# Supplementary 1

## Preliminary of Logical Reasoning

We introduce some propositional connectives and rules for proof of propositional logical reasoning, which are respectively shown as Table 1 and Table 2, for the logical reasonings conducted in this paper.

Table 1. Propositional connectives

| Connective | Meaning |
|---|---|
| $\wedge$ | conjunction |
| $\rightarrow$ | implication |

Table 2. Rules for proof of propositional logical reasoning, $\vdash$ denotes 'bring out'

| Rule | Meaning |
|---|---|
| $\wedge -$ | reductive law of conjunction: $A \wedge B$, $\vdash$ A or B. |
| $\wedge +$ | additional law of conjunction: $A, B$, $\vdash A \wedge B$. |
| MP | modus ponens: $A \rightarrow B$, A, $\vdash$ B. |
| HS | hypothetical syllogism: $A \rightarrow B$, $B \rightarrow C$, $\vdash A \rightarrow C$. |

## Proof of Reasoning 1

**Reasoning 1**. *If $G_1$ is given and $KB$ is given, then inconsistency between the false tumour negatives covered by $G_{1,1}$ and the false tumour negatives covered by $KB$ is low, inconsistency between the false tumour positives covered by $G_{1,2}$ and the false tumour positives covered by $KB$ is high, inconsistency between the true tumour positives covered by $G_{1,2}$ and the true tumour positives covered by $KB$ is low, and inconsistency between the true tumour negatives covered by $G_{1,1}$ and the true tumour negatives covered by $KB$ is high.*

**Proof-R1**. Firstly, with $G_1$ and $KB$, we have following derived preconditions for Reasoning 1.

1. If $G_1$ is given, then $NS_1 = \{IS_1, NLS_1\}$ exists. ($G_1$ is associated with $NS_1$ in Grounding Extract)
2. If $NS_1 = \{IS_1, NLS_1\}$ exists, then $NLS_1$ can exclude tumour negatives of $IS_1$ as accurate as possible, and $NLS_1$ can include many tumour negatives of $IS_1$ as tumour positives. (facts contained in the existing $NS_1$)
3. If $G_1$ is given, then $G_{1,1}$: pixels of $IS_1$ outside the polygons of $NLS_1$ are tumour negatives is given, and $G_{1,2}$: pixels of $IS_1$ inside the polygons of $NLS_1$ are tumour positives is given. (groundings extracted from $NS_1$ by Grounding Extract)
4. If $G_{1,1}$: pixels of $IS_1$ outside the polygons of $NLS_1$ are tumour negatives is given, and $G_{1,2}$: pixels of $IS_1$ inside the polygons of $NLS_1$ are tumour positives is given, then $G_{1,1}$ and $G_{1,2}$ are complementary to each other.

5. If $NLS_1$ can exclude tumour negatives of $IS_1$ as accurate as possible, and $G_{1,1}$: pixels of $IS_1$ outside the polygons of $NLS_1$ are tumour negatives is given, then false tumour negatives covered by $G_{1,1}$ are rare.
6. If $NLS_1$ can include many tumour negatives of $IS_1$ as tumour positives, and $G_{1,2}$: pixels of $IS_1$ inside the polygons of $NLS_1$ are tumour positives is given, then false tumour positives covered by $G_{1,2}$ are many.
7. If false tumour negatives covered by $G_{1,1}$ are rare, and $G_{1,1}$ and $G_{1,2}$ are complementary to each other, then $G_{1,2}$ covers almost all true tumour positives.
8. If false tumour positives covered by $G_{1,2}$ are many, and $G_{1,1}$ and $G_{1,2}$ are complementary to each other, then $G_{1,1}$ covers only a part of true tumour negatives.
9. If $KB$ is given, then the information of $KB$ can clearly describe what are the true tumour positives, and the opposite information of $KB$ can clearly describe what are the true tumour negatives. (facts contained in the given $KB$)
10. If the information of $KB$ can clearly describe what are the true tumour positives, then true tumour positives covered by $KB$ are all-sided, and false tumour negatives covered by $KB$ are none.
11. If the opposite information of $KB$ can clearly describe what are the true tumour negatives, then true tumour negatives covered by $KB$ are all-sided, and false tumour positives covered by $KB$ are none.
12. If false tumour negatives covered by $G_{1,1}$ are rare, and false tumour negatives covered by $KB$ are none, then inconsistency between the false tumour negatives covered by $G_{1,1}$ and the false tumour negatives covered by $KB$ is low.
13. If false tumour positives covered by $G_{1,2}$ are many, and false tumour positives covered by $KB$ are none, then inconsistency between false tumour positives covered by $G_{1,2}$ and the false tumour positives covered by $KB$ is high.
14. If $G_{1,2}$ covers almost all true tumour positives, and true tumour positives covered by $KB$ are all-sided, then inconsistency between the true tumour positives covered by $G_{1,2}$ and the true tumour positives covered by $KB$ is low.
15. If $G_{1,1}$ covers only a part of true tumour negatives, and true tumour negatives covered by $KB$ are all-sided, then inconsistency between true tumour negatives covered by $G_{1,1}$ and the true tumour negatives covered by $KB$ is high.

Secondly, we give the propositional symbols for the above preconditions 1-15 for Reasoning 1, which are shown in Table 3.

Table 3. Propositional symbols of preconditions for Reasoning 1

| Symbol | Meaning |
|---|---|
| $a$ | $G_1$ is given |
| $b$ | $NS_1 = \{IS_1, NLS_1\}$ exists |
| $c$ | $NLS_1$ can exclude tumour negatives of $IS_1$ as accurate as possible |
| $d$ | $NLS_1$ can include many tumour negatives of $IS_1$ as tumour positives |
| $e$ | $G_{1,1}$: pixels of $IS_1$ outside the polygons of $NLS_1$ are tumour negatives is given |
| $f$ | $G_{1,2}$: pixels of $IS_1$ inside the polygons of $NLS_1$ are tumour positives is given |
| $g$ | $G_{1,1}$ and $G_{1,2}$ are complementary to each other |
| $h$ | false tumour negatives covered by $G_{1,1}$ are rare |
| $i$ | false tumour positives covered by $G_{1,2}$ are many |
| $j$ | $G_{1,2}$ covers almost all true tumour positives |
| $k$ | $G_{1,1}$ covers only a part of true tumour negatives |
| $l$ | $KB$ is given |
| $m$ | the information of $KB$ can clearly describe what are the true tumour positives |
| $n$ | the opposite information of $KB$ can clearly describe what are the true tumour negatives |
| $o$ | true tumour positives covered by $KB$ are all-sided |
| $p$ | false tumour negatives covered by $KB$ are none |
| $q$ | true tumour negatives covered by $KB$ are all-sided |
| $r$ | false tumour positives covered by $KB$ are none |
| $s$ | inconsistency between the false tumour negatives covered by $G_{1,1}$ and the false tumour negatives covered by $KB$ is low |
| $t$ | inconsistency between false tumour positives covered by $G_{1,2}$ and the false tumour positives covered by $KB$ is high |
| $u$ | inconsistency between the true tumour positives covered by $G_{1,2}$ and the true tumour positives covered by $KB$ is low |
| $v$ | inconsistency between true tumour negatives covered by $G_{1,1}$ and the true tumour negatives covered by $KB$ is high |

Thirdly, referring to Table 3, we signify the propositional formalizations of the preconditions 1-15 for Reasoning 1 and Reasoning 1 via the propositional connectives listed in Table 1 as follows.

| | |
|---|---|
| 1) $a \rightarrow b$ | Precondition |
| 2) $b \rightarrow (c \wedge d)$ | Precondition |
| 3) $a \rightarrow (e \wedge f)$ | Precondition |
| 4) $(e \wedge f) \rightarrow g$ | Precondition |
| 5) $(c \wedge e) \rightarrow h$ | Precondition |
| 6) $(d \wedge f) \rightarrow i$ | Precondition |
| 7) $(h \wedge g) \rightarrow j$ | Precondition |
| 8) $(i \wedge g) \rightarrow k$ | Precondition |
| 9) $l \rightarrow (m \wedge n)$ | Precondition |
| 10) $m \rightarrow (o \wedge p)$ | Precondition |
| 11) $n \rightarrow (q \wedge r)$ | Precondition |
| 12) $(h \wedge p) \rightarrow s$ | Precondition |
| 13) $(i \wedge r) \rightarrow t$ | Precondition |
| 14) $(j \wedge o) \rightarrow u$ | Precondition |
| 15) $(k \wedge q) \rightarrow v$ | Precondition |
| $(a \wedge l) \rightarrow (s \wedge t \wedge u \wedge v)$ | Reasoning 1 |

Fourthly, we show the validity of Reasoning 1 via the rules for proof of propositional logical reasoning listed in Table 3 as follows.

$\therefore \boldsymbol{(a \wedge l) \rightarrow (s \wedge t \wedge u \wedge v)}$

| | |
|---|---|
| 16) $a \wedge l$ | Hypothesis |
| 17) $a$ | 16); $\wedge -$ |
| 18) $l$ | 16); $\wedge -$ |
| 19) $a \rightarrow (c \wedge d)$ | 1),2); HS |
| 20) $c \wedge d$ | 19),17); MP |
| 21) $e \wedge f$ | 3),17); MP |
| 22) $c$ | 20); $\wedge -$ |
| 23) $d$ | 20); $\wedge -$ |
| 24) $e$ | 21); $\wedge -$ |
| 25) $f$ | 21); $\wedge -$ |
| 26) $c \wedge e$ | 22),24); $\wedge +$ |
| 27) $d \wedge f$ | 23),25); $\wedge +$ |
| 28) $h$ | 5),26); MP |
| 29) $i$ | 6),27); MP |
| 30) $a \to g$ | 3),4); HS |
| 31) $g$ | 30),17); MP |
| 32) $h \wedge g$ | 28),31); $\wedge +$ |
| 33) $i \wedge g$ | 29),31); $\wedge +$ |
| 34) $j$ | 7),32); MP |
| 35) $k$ | 8),33); MP |
| 36) $m \wedge n$ | 9),18); MP |
| 37) $m$ | 36); $\wedge -$ |
| 38) $n$ | 36); $\wedge -$ |
| 39) $o \wedge p$ | 10),37); MP |
| 40) $q \wedge r$ | 11),38); MP |
| 41) $o$ | 39); $\wedge -$ |
| 42) $p$ | 39); $\wedge -$ |
| 43) $q$ | 40); $\wedge -$ |
| 44) $r$ | 40); $\wedge -$ |
| 45) $h \wedge p$ | 28),42); $\wedge +$ |
| 46) $i \wedge r$ | 29),44); $\wedge +$ |
| 47) $j \wedge o$ | 34),41); $\wedge +$ |
| 48) $k \wedge q$ | 35),43); $\wedge +$ |
| 49) $s$ | 12),45); MP |
| 50) $t$ | 13),46); MP |
| 51) $u$ | 14),47); MP |
| 52) $v$ | 15),48); MP |
| 53) $s \wedge t \wedge u \wedge v$ | 49),50),51),52); $\wedge +$ |
| 54) $(a \wedge l) \to (s \wedge t \wedge u \wedge v)$ | 16)-53); Conditional Proof |

Since the hypothesis $a \wedge l$ of the 16) step can be fulfilled by the input materials of OSAMTL-DiNS applied on tumour segmentation for breast cancer and Grounding Extract, Reasoning 1 is proved to be valid.

## Proof of Reasoning 2

**Reasoning 2**. *If $G_2$ is given and $KB$ is given, then inconsistency between the false tumour positives covered by $G_{2,1}$ and the false tumour positives covered by $KB$ is low, inconsistency between the false tumour negatives covered by $G_{2,2}$ and the false tumour* *negatives covered by $KB$ is high, inconsistency between the true tumour negatives covered by $G_{2,2}$ and the true tumour negatives covered by $KB$ is low, and inconsistency between the true tumour positives covered by $G_{1,1}$ and the true tumour positives covered by $KB$ is high.*

**Proof-R2**. Firstly, with $G_2$ and $KB$, we have following derived preconditions for Reasoning 2.

1. If $G_2$ is given, then $NS_2 = \{IS_2, NLS_2\}$ exists. ($G_2$ is associated with $NS_2$ in Grounding Extract)
2. If $NS_2 = \{IS_2, NLS_2\}$ exists, then $NLS_2$ can include tumour negatives of $IS_2$ as accurate as possible, and $NLS_2$ can exclude many tumour positives of $IS_2$ as tumour negatives. (facts contained in the existing $NS_2$)
3. If $G_2$ is given, then $G_{2,1}$: pixels of $IS_2$ inside the polygons of $NLS_2$ are tumour positives is given, and $G_{2,2}$: pixels of $IS_2$ outside the polygons of $NLS_2$ are tumour negatives is given. (groundings extracted from $NS_2$ by Grounding Extract)
4. If $G_{2,1}$: pixels of $IS_2$ inside the polygons of $NLS_2$ are tumour positives is given, and $G_{2,2}$: pixels of $IS_2$ outside the polygons of $NLS_2$ are tumour negatives is given, then $G_{2,1}$ and $G_{2,2}$ are complementary to each other.
5. If $NLS_2$ can include tumour negatives of $IS_2$ as accurate as possible, and $G_{2,1}$: pixels of $IS_2$ inside the polygons of $NLS_2$ are tumour positives is given, then false tumour positives covered by $G_{2,1}$ are rare.
6. If $NLS_2$ can exclude many tumour positives of $IS_2$ as tumour negatives, and $G_{2,2}$: pixels of $IS_2$ outside the polygons of $NLS_2$ are tumour negatives is given, then false tumour negatives covered by $G_{2,2}$ are many.
7. If false tumour positives covered by $G_{2,1}$ are rare, and $G_{2,1}$ and $G_{2,2}$ are complementary to each other, then $G_{2,2}$ covers almost all true tumour negatives.
8. If false tumour negatives covered by $G_{2,2}$ are many, and $G_{2,1}$ and $G_{2,2}$ are complementary to each other, then $G_{2,1}$ covers only a part of true tumour positives.
9. If $KB$ is given, then the information of $KB$ can clearly describe what are the true tumour positives, and the opposite information of $KB$ can clearly describe what are the true tumour negatives. (facts contained in the given $KB$)
10. If the information of $KB$ can clearly describe what are the true tumour positives, then true tumour positives covered by $KB$ are all-sided, and false tumour negatives covered by $KB$ are none.
11. If the opposite information of $KB$ can clearly describe what are the true tumour negatives, then true tumour negatives covered by $KB$ are all-sided, and false tumour positives covered by $KB$ are none.
12. If false tumour positives covered by $G_{2,1}$ are rare, and false tumour positives covered by $KB$ are none, then inconsistency between the false tumour positives covered by $G_{2,1}$ and the false tumour positives covered by $KB$ is low.

13. If false tumour negatives covered by $G_{2,2}$ are many, and false tumour negatives covered by $KB$ are none, then inconsistency between false tumour negatives covered by $G_{2,2}$ and the false tumour negatives covered by $KB$ is high.
14. If $G_{2,2}$ covers almost all true tumour negatives, and true tumour negatives covered by $KB$ are all-sided, then inconsistency between the true tumour negatives covered by $G_{2,2}$ and the true tumour negatives covered by $KB$ is low.
15. If $G_{2,1}$ covers only a part of true tumour positives, and true tumour positives covered by $KB$ are all-sided, then inconsistency between true tumour positives covered by $G_{2,1}$ and the true tumour positives covered by $KB$ is high.

Secondly, we give the propositional symbols for the above preconditions 1-15 for Reasoning 2, which are shown in Table 3.

Table 3. Propositional symbols of preconditions for Reasoning 2

| Symbol | Meaning |
|---|---|
| $a$ | $G_2$ is given |
| $b$ | $NS_2 = \{IS_2, NLS_2\}$ exists |
| $c$ | $NLS_2$ can include tumour negatives of $IS_2$ as accurate as possible |
| $d$ | $NLS_2$ can exclude many tumour positives of $IS_2$ as tumour negatives |
| $e$ | $G_{2,1}$: pixels of $IS_2$ inside the polygons of $NLS_2$ are tumour positives is given |
| $f$ | $G_{2,2}$: pixels of $IS_2$ outside the polygons of $NLS_2$ are tumour negatives is given |
| $g$ | $G_{2,1}$ and $G_{2,2}$ are complementary to each other |
| $h$ | false tumour positives covered by $G_{2,1}$ are rare |
| $i$ | false tumour negatives covered by $G_{2,2}$ are many |
| $j$ | $G_{2,2}$ covers almost all true tumour negatives |
| $k$ | $G_{2,1}$ covers only a part of true tumour positives |
| $l$ | $KB$ is given |
| $m$ | the information of $KB$ can clearly describe what are the true tumour positives |
| $n$ | the opposite information of $KB$ can clearly describe what are the true tumour negatives |
| $o$ | true tumour positives covered by $KB$ are all-sided |
| $p$ | false tumour negatives covered by $KB$ are none |
| $q$ | true tumour negatives covered by $KB$ are all-sided |
| $r$ | false tumour positives covered by $KB$ are none |
| $s$ | inconsistency between the false tumour positives covered by $G_{2,1}$ and the false tumour positives covered by $KB$ is low |
| $t$ | inconsistency between false tumour negatives covered by $G_{2,2}$ and the false tumour negatives covered by $KB$ is high |
| $u$ | inconsistency between the true tumour negatives covered by $G_{2,2}$ and the true tumour negatives covered by $KB$ is low |
| $v$ | inconsistency between true tumour positives covered by $G_{2,1}$ and the true tumour positives covered by $KB$ is high |

Thirdly, referring to Table 3, we signify the propositional formalizations of the preconditions 1-15 for Reasoning 2 and Reasoning 2 via the propositional connectives listed in Table 1 as follows.

1) $a \to b$ Precondition
2) $b \to (c \wedge d)$ Precondition
3) $a \to (e \wedge f)$ Precondition
4) $(e \wedge f) \to g$ Precondition

| | |
|---|---|
| 5) $(c \wedge e) \rightarrow h$ | Precondition |
| 6) $(d \wedge f) \rightarrow i$ | Precondition |
| 7) $(h \wedge g) \rightarrow j$ | Precondition |
| 8) $(i \wedge g) \rightarrow k$ | Precondition |
| 9) $l \rightarrow (m \wedge n)$ | Precondition |
| 10) $m \rightarrow (o \wedge p)$ | Precondition |
| 11) $n \rightarrow (q \wedge r)$ | Precondition |
| 12) $(h \wedge r) \rightarrow s$ | Precondition |
| 13) $(i \wedge p) \rightarrow t$ | Precondition |
| 14) $(j \wedge q) \rightarrow u$ | Precondition |
| 15) $(k \wedge o) \rightarrow v$ | Precondition |
| $(a \wedge l) \rightarrow (s \wedge t \wedge u \wedge v)$ | Reasoning 2 |

Fourthly, we show the validity of Reasoning 1 via the rules for proof of propositional logical reasoning listed in Table 3 as follows.

$\therefore (\boldsymbol{a} \wedge \boldsymbol{l}) \rightarrow (\boldsymbol{s} \wedge \boldsymbol{t} \wedge \boldsymbol{u} \wedge \boldsymbol{v})$

| | |
|---|---|
| 16) $a \wedge l$ | Hypothesis |
| 17) $a$ | 16); $\wedge -$ |
| 18) $l$ | 16); $\wedge -$ |
| 19) $a \rightarrow (c \wedge d)$ | 1),2); HS |
| 20) $c \wedge d$ | 19),17); MP |
| 21) $e \wedge f$ | 3),17); MP |
| 22) $c$ | 20); $\wedge -$ |
| 23) $d$ | 20); $\wedge -$ |
| 24) $e$ | 21); $\wedge -$ |
| 25) $f$ | 21); $\wedge -$ |
| 26) $c \wedge e$ | 22),24); $\wedge +$ |
| 27) $d \wedge f$ | 23),25); $\wedge +$ |
| 28) $h$ | 5),26); MP |
| 29) $i$ | 6),27); MP |
| 30) $a \rightarrow g$ | 3),4); HS |
| 31) $g$ | 30),17); MP |
| 32) $h \wedge g$ | 28),31); $\wedge +$ |
| 33) $i \wedge g$ | 29),31); $\wedge +$ |
| 34) $j$ | 7),32); MP |
| 35) $k$ | 8),33); MP |
| 36) $m \wedge n$ | 9),18); MP |
| 37) $m$ | 36); $\wedge -$ |
| 38) $n$ | 36); $\wedge -$ |
| 39) $o \wedge p$ | 10),37); MP |
| 40) $q \wedge r$ | 11),38); MP |
| 41) $o$ | 39); $\wedge -$ |
| 42) $p$ | 39); $\wedge -$ |
| 43) $q$ | 40); $\wedge -$ |
| 44) $r$ | 40); $\wedge -$ |
| 45) $h \wedge r$ | 28),44); $\wedge +$ |
| 46) $i \wedge p$ | 29),42); $\wedge +$ |
| 47) $j \wedge q$ | 34),43); $\wedge +$ |
| 48) $k \wedge o$ | 35),41); $\wedge +$ |
| 49) $s$ | 12),45); MP |
| 50) $t$ | 13),46); MP |
| 51) $u$ | 14),47); MP |
| 52) $v$ | 15),48); MP |
| 53) $s \wedge t \wedge u \wedge v$ | 49),50),51),52); $\wedge +$ |
| 54) $(a \wedge l) \rightarrow (s \wedge t \wedge u \wedge v)$ | 16)-53); Conditional Proof |

Since the hypothesis $a \wedge l$ of the 16) step can be fulfilled by the input materials of OSAMTL-DiNS applied on tumour segmentation for breast cancer and Grounding Extract, Reasoning 2 is proved to be valid.

## Proof of Reasoning 3

**Reasoning 3**. *If $IC_1$ is given, then $G_{1,1}$ should not be revised to remain $IC_{1,1}$, $G_{1,2}$ should not be revised to remain $IC_{1,3}$, $G_{1,1}$ should be revised to reduce $IC_{1,4}$, and $G_{1,2}$ should be revised to reduce $IC_{1,2}$.*

**Proof-R3**. Firstly, with $IC_1$, we have following derived preconditions for Reasoning 3.

1. If $IC_1$ is given, then $G_1$ exists. ($IC_1$ is associated with $G_1$ in Reasoning)
2. If $IC_1$ is given, then $IC_{1,1}$ is given, $IC_{1,2}$ is given, $IC_{1,3}$ is given, and $IC_{1,4}$ is given.
3. If $G_1$exists, then $G_{1,1}$ exists, and $G_{1,2}$ exists.
4. If $IC_{1,1}$ is given, then inconsistency between the false tumour negatives covered by $G_{1,1}$ and the false tumour negatives covered by $KB$ is low. (inconsistency estimated by Reasoning)
5. If $G_{1,1}$ exists, and inconsistency between the false tumour negatives covered by $G_{1,1}$ and the false tumour negatives covered by $KB$ is low, then $G_{1,1}$ should not be revised to remain $IC_{1,1}$.
6. If $IC_{1,3}$ is given, then inconsistency between the true tumour positives covered by $G_{1,2}$ and the true tumour positives covered by $KB$ is low. (inconsistency estimated by Reasoning)
7. If $G_{1,2}$ is given, and inconsistency between the true tumour positives covered by $G_{1,2}$ and the true tumour positives covered by $KB$ is low, then $G_{1,2}$ should not be revised to remain $IC_{1,3}$.
8. If $IC_{1,2}$ is given, then inconsistency between the false tumour positives covered by $G_{1,2}$ and the false tumour positives covered by $KB$ is high. (inconsistency estimated by Reasoning)

9. If $G_{1,2}$ exists, and inconsistency between the false tumour positives covered by $G_{1,2}$ and the false tumour positives covered by $KB$ is high, then $G_{1,2}$ should be revised to reduce $IC_{1,2}$.
10. If $IC_{1,4}$ is given, then inconsistency between the true tumour negatives covered by $G_{1,1}$ and the true tumour negatives covered by $KB$ is high. (inconsistency estimated by Reasoning)
11. If $G_{1,1}$ exists, and inconsistency between the true tumour negatives covered by $G_{1,1}$ and the true tumour negatives covered by $KB$ is high, then $G_{1,1}$ should be revised to reduce $IC_{1,4}$.

Secondly, we give the propositional symbols for the above preconditions 1-11 for Reasoning 3, which are shown in Table 5.

Table 5. Propositional symbols of preconditions for Reasoning 3

| Symbol | Meaning |
|---|---|
| $a$ | $IC_1$ is given |
| $b$ | $G_1$ exists |
| $c$ | $IC_{1,1}$ is given |
| $d$ | $IC_{1,2}$ is given |
| $e$ | $IC_{1,3}$ is given |
| $f$ | $IC_{1,4}$ is given |
| $g$ | $G_{1,1}$ exists |
| $h$ | $G_{1,2}$ exists |
| $i$ | inconsistency between the false tumour negatives covered by $G_{1,1}$ and the false tumour negatives covered by $KB$ is low |
| $j$ | $G_{1,1}$ should not be revised to remain $IC_{1,1}$ |
| $k$ | inconsistency between the true tumour positives covered by $G_{1,2}$ and the true tumour positives covered by $KB$ is low |
| $l$ | $G_{1,2}$ should not be revised to remain $IC_{1,3}$ |
| $m$ | inconsistency between the false tumour positives covered by $G_{1,2}$ and the false tumour positives covered by $KB$ is high |
| $n$ | then $G_{1,2}$ should be revised to reduce $IC_{1,2}$ |
| $o$ | inconsistency between the true tumour negatives covered by $G_{1,1}$ and the true tumour negatives covered by $KB$ is high |
| $p$ | $G_{1,1}$ should be revised to reduce $IC_{1,4}$ |

Thirdly, referring to Table 5, we signify the propositional formalizations of the preconditions 1-11 for Reasoning 3 and Reasoning 3 via the propositional connectives listed in Table 1 as follows.

1) $a \rightarrow b$ Precondition
2) $a \rightarrow (c \wedge d \wedge e \wedge f)$ Precondition
3) $b \rightarrow (g \wedge h)$ Precondition
4) $c \rightarrow i$ Precondition
5) $(g \wedge i) \rightarrow j$ Precondition
6) $e \rightarrow k$ Precondition
7) $(h \wedge k) \rightarrow l$ Precondition
8) $d \rightarrow m$ Precondition

9) $(h \wedge m) \rightarrow n$ Precondition

10) $f \rightarrow o$ Precondition

11) $(g \wedge o) \rightarrow p$ Precondition

$a \rightarrow (j \wedge l \wedge n \wedge p)$ Reasoning 3

Fourthly, we show the validity of Reasoning 3 via the rules for proof of propositional logical reasoning listed in Table 2 as follows.

$\therefore \boldsymbol{a \rightarrow (j \wedge l \wedge n \wedge p)}$

| | |
|---|---|
| 12) $a$ | Hypothesis |
| 13) $a \rightarrow (g \wedge h)$ | 1),3); HS |
| 14) $g \wedge h$ | 13),12); MP |
| 15) $g$ | 14); $\wedge -$ |
| 16) $h$ | 14); $\wedge -$ |
| 17) $c \wedge d \wedge e \wedge f$ | 2),12); MP |
| 18) $c$ | 17); $\wedge -$ |
| 19) $d$ | 17); $\wedge -$ |
| 20) $e$ | 17); $\wedge -$ |
| 21) $f$ | 17); $\wedge -$ |
| 22) $i$ | 4),18); MP |
| 23) $g \wedge i$ | 15),22); $\wedge +$ |
| 24) $j$ | 5),23); MP |
| 25) $k$ | 6),20); MP |
| 26) $h \wedge k$ | 16),25); $\wedge +$ |
| 27) $l$ | 7),26); MP |
| 28) $m$ | 8),19); MP |
| 29) $h \wedge m$ | 16),28); $\wedge +$ |
| 30) $n$ | 9),29); MP |
| 31) $o$ | 10),21); MP |
| 32) $g \wedge o$ | 15),31); $\wedge +$ |
| 33) $p$ | 11),32); MP |
| 34) $j \wedge l \wedge n \wedge p$ | 24),27),30),33); $\wedge +$ |
| 35) $a \rightarrow j \wedge l \wedge n \wedge p$ | 12)-34); Conditional Proof |

Since the hypothesis $a$ of the 17) step can be fulfilled by the Reasoning step, Reasoning 3 is proved to be valid.

## Proof of Reasoning 4

**Reasoning 4**. *If $IC_2$ is given, then $G_{2,1}$ should not be revised to remain $IC_{2,1}$, then $G_{2,2}$ should not be revised to remain $IC_{2,3}$, $G_{2,1}$ should be revised to reduce $IC_{2,2}$, and $G_{2,2}$ should be revised to reduce $IC_{2,4}$.*

**Proof-R4**. Firstly, with $IC_2$, we have following derived preconditions for Reasoning 4.

1. If $IC_2$ is given, then $G_2$ exists. ($IC_2$ is associated with $G_2$ in Reasoning)
2. If $IC_2$ is given, then $IC_{2,1}$ is given, $IC_{2,2}$ is given, $IC_{2,3}$ is given, and $IC_{2,4}$ is given.

3. If $G_2$exists, then $G_{2,1}$ exists, and $G_{2,2}$ exists.
4. If $IC_{2,1}$ is given, then inconsistency between the false tumour positives covered by $G_{2,1}$ and the false tumour positives covered by $KB$ is low. (inconsistency estimated by Reasoning)
5. If $G_{2,1}$ exists, and inconsistency between the false tumour positives covered by $G_{2,1}$ and the false tumour positives covered by $KB$ is low, then $G_{2,1}$ should not be revised to remain $IC_{2,1}$.
6. If $IC_{2,3}$ is given, then inconsistency between the true tumour negatives covered by $G_{2,2}$ and the true tumour negatives covered by $KB$ is low. (inconsistency estimated by Reasoning)
7. If $G_{2,2}$ is given, and inconsistency between the true tumour negatives covered by $G_{2,2}$ and the true tumour negatives covered by $KB$ is low, then $G_{2,2}$ should not be revised to remain $IC_{2,3}$.
8. If $IC_{2,2}$ is given, then inconsistency between the false tumour negatives covered by $G_{2,2}$ and the false tumour negatives covered by $KB$ is high. (inconsistency estimated by Reasoning)
9. If $G_{2,2}$ exists, and inconsistency between the false tumour negatives covered by $G_{2,2}$ and the false tumour negatives covered by $KB$ is high, then $G_{2,2}$ should be revised to reduce $IC_{2,2}$.
10. If $IC_{2,4}$ is given, then inconsistency between the true tumour positives covered by $G_{2,1}$ and the true tumour positives covered by $KB$ is high. (inconsistency estimated by Reasoning)
11. If $G_{2,1}$ exists, and inconsistency between the true tumour positives covered by $G_{2,1}$ and the true tumour negatives covered by $KB$ is high, then $G_{2,1}$ should be revised to reduce $IC_{2,4}$.

Secondly, we give the propositional symbols for the above preconditions 1-11 for Reasoning 4, which are shown in Table 6.

Table 6. Propositional symbols of preconditions for Reasoning 4

| Symbol | Meaning |
|---|---|
| $a$ | $IC_2$ is given |
| $b$ | $G_2$ exists |
| $c$ | $IC_{2,1}$ is given |
| $d$ | $IC_{2,2}$ is given |
| $e$ | $IC_{2,3}$ is given |
| $f$ | $IC_{2,4}$ is given |
| $g$ | $G_{2,1}$ exists |
| $h$ | $G_{2,2}$ exists |
| $i$ | inconsistency between the false tumour positives covered by $G_{2,1}$ and the false tumour positives covered by $KB$ is low |
| $j$ | $G_{2,1}$ should not be revised to remain $IC_{2,1}$ |
| $k$ | inconsistency between the true tumour negatives covered by $G_{2,2}$ and the true tumour negatives covered by $KB$ is low |
| $l$ | $G_{2,2}$ should not be revised to remain $IC_{2,3}$ |
| $m$ | inconsistency between the false tumour negatives covered by $G_{2,2}$ and the false tumour negatives covered by $KB$ is high |
| $n$ | then $G_{2,2}$ should be revised to reduce $IC_{2,2}$ |
| $o$ | inconsistency between the true tumour positives covered by $G_{2,1}$ and the true tumour positives covered by $KB$ is high |
| $p$ | $G_{2,1}$ should be revised to reduce $IC_{2,4}$ |

Thirdly, referring to Table 6, we signify the propositional formalizations of the preconditions 1-11 for Reasoning 4 and Reasoning 4 via the propositional connectives listed in Table 1 as follows.

| | |
|---|---|
| 1) $a \rightarrow b$ | Precondition |
| 2) $a \rightarrow (c \land d \land e \land f)$ | Precondition |
| 3) $b \rightarrow (g \land h)$ | Precondition |
| 4) $c \rightarrow i$ | Precondition |
| 5) $(g \land i) \rightarrow j$ | Precondition |
| 6) $e \rightarrow k$ | Precondition |
| 7) $(h \land k) \rightarrow l$ | Precondition |
| 8) $d \rightarrow m$ | Precondition |
| 9) $(h \land m) \rightarrow n$ | Precondition |
| 10) $f \rightarrow o$ | Precondition |
| 11) $(g \land o) \rightarrow p$ | Precondition |
| $a \rightarrow (j \land l \land n \land p)$ | Reasoning 3 |

Fourthly, we show the validity of Reasoning 3 via the rules for proof of propositional logical reasoning listed in Table 2 as follows.

$\therefore \boldsymbol{a} \rightarrow (\boldsymbol{j} \land \boldsymbol{l} \land \boldsymbol{n} \land \boldsymbol{p})$

| | |
|---|---|
| 12) $a$ | Hypothesis |
| 13) $a \rightarrow (g \land h)$ | 1),3); HS |
| 14) $g \land h$ | 13),12); MP |
| 15) $g$ | 14); $\land -$ |
| 16) $h$ | 14); $\land -$ |
| 17) $c \land d \land e \land f$ | 2),12); MP |
| 18) $c$ | 17); $\land -$ |
| 19) $d$ | 17); $\land -$ |
| 20) $e$ | 17); $\land -$ |
| 21) $f$ | 17); $\land -$ |
| 22) $i$ | 4),18); MP |
| 23) $g \land i$ | 15),22); $\land +$ |
| 24) $j$ | 5),23); MP |
| 25) $k$ | 6),20); MP |
| 26) $h \land k$ | 16),25); $\land +$ |
| 27) $l$ | 7),26); MP |
| 28) $m$ | 8),19); MP |
| 29) $h \land m$ | 16),28); $\land +$ |
| 30) $n$ | 9),29); MP |

31) $o$ 10),21); MP

32) $g \wedge o$ 15),31); $\wedge +$

33) $p$ 11),32); MP

34) $j \wedge l \wedge n \wedge p$ 24),27),30),33); $\wedge +$

35) $a \rightarrow j \wedge l \wedge n \wedge p$ 12)-34); Conditional Proof

Since the hypothesis $a$ of the 17) step can be fulfilled by the Reasoning step, Reasoning 4 is proved to be valid.

## Proof of Reasoning 5

**Reasoning 5**. *If $RG_1$ is given and $RG_2$ is given, then a target ($T_1$) can be abduced from the union of $RG_1$ and $RG_2$, and $T_1$ has a high recall of true tumour positives and a high precision of true tumour negatives.*

**Proof-R5**. Firstly, with $RG_1$ and $RG_2$, we have following derived preconditions for Reasoning 5.

1. If $RG_1$ is given, then $RG_1$ is equivalent to $G_{1,1}$: pixels of $IS_1$ outside the polygons of $NLS_1$ are tumour negatives. (revised grounding produced by Abduction)
2. If $RG_2$ is given, then $RG_2$ is equivalent to $G_{1,2}$: pixels of $IS_1$ inside the polygons of $NLS_1$ are tumour positives. (revised grounding produced by Abduction)
3. If $RG_1$ is equivalent to $G_{1,1}$: pixels of $IS_1$ outside the polygons of $NLS_1$ are tumour negatives and $RG_2$ is equivalent to $G_{1,2}$: pixels of $IS_1$ inside the polygons of $NLS_1$ are tumour positives, then $RG_1$ indicates where are tumour negatives of $IS_1$ and $RG_2$ indicates where are tumour positives of $IS_1$.
4. If $RG_1$ indicates where are tumour negatives of $IS_1$ and $RG_2$ indicates where are tumour positives of $IS_1$, then a target ($T_1$) can be abduced from the union of $RG_1$ and $RG_2$.
5. If $T_1$ can be abduced from the union of $RG_1$ and $RG_2$, $RG_1$ is equivalent to $G_{1,1}$: pixels of $IS_1$ outside the polygons of $NLS_1$ are tumour negatives, and $RG_2$ is equivalent to $G_{1,2}$ pixels of $IS_1$ inside the polygons of $NLS_1$ are tumour positives, then the recall of $T_1$ for true tumour positives can be denoted by true tumour positives covered by $G_{1,2}$ (TP($G_{1,2}$)) dividing the sum of TP($G_{1,2}$) and false tumour negatives covered by $G_{1,1}$ (FN($G_{1,1}$)), i.e., TP($G_{1,2}$)/( TP($G_{1,2}$)+ FN($G_{1,1}$)); and the precision of $T_1$ for true tumour negatives can be denoted by true tumor negatives covered by $G_{1,1}$ (TN($G_{1,1}$)) dividing the sum of TN($G_{1,1}$) and FN($G_{1,1}$), i.e., TN($G_{1,1}$)/(TN($G_{1,1}$)+ FN($G_{1,1}$)).
6. If $RG_1$ is given, then $G_{1,1}$ should not be revised to remain $IC_{1,1}$. (revised grounding produced by Abduction is only associated with corresponding grounding revision in Abduction)
7. If $G_{1,1}$ should not be revised to remain $IC_{1,1}$, then $G_{1,1}$ should not be revised to remain the fact that inconsistency between FN($G_{1,1}$) and false tumour negatives covered by $KB$ (FN($KB$)) is low. (fact contained in $IC_{1,1}$)

8. If $G_{1,1}$ is not revised to remain the fact that inconsistency between FN($G_{1,1}$) and FN($KB$) is low, then FN($G_{1,1}$) can be regarded as close to FN($KB$).
9. If FN($G_{1,1}$) can be regarded as close to FN($KB$), then FN($G_{1,1}$) can be regarded as close to 0.
10. If FN($G_{1,1}$) can be regarded as close to 0, then tumour negatives covered by $G_{1,1}$ can be regarded as true tumor negatives.
11. If tumour negatives covered by $G_{1,1}$ can be regarded as true tumor negatives, then TN($G_{1,1}$) can be regarded as a constant positive integer.
12. If $RG_2$ is given, then $G_{1,2}$ should not be revised to remain $IC_{1,3}$. (revised grounding produced by Abduction is only associated with corresponding grounding revision in Abduction)
13. If $G_{1,2}$ should not be revised to remain $IC_{1,3}$, then $G_{1,2}$ should not be revised to remain the fact that inconsistency between TP($G_{1,2}$) and true tumour positives covered by $KB$ (TP($KB$)) is low.
14. If $G_{1,2}$ should not be revised to remain the fact that inconsistency between TP($G_{1,2}$) and TP($KB$) is low, then TP($G_{1,2}$) can be regarded as close to TP($KB$).
15. If TP($G_{1,2}$) can be regarded as close to TP($KB$), TP($G_{1,2}$) can be regarded as a constant positive integer.
16. If TP($G_{1,2}$) can be regarded as a constant positive integer and FN($G_{1,1}$) can be regarded as close to 0, and the recall of $T_1$for true tumour positives can be denoted by TP($G_{1,2}$)/( TP($G_{1,2}$)+ FN($G_{1,1}$)), then the recall of $T_1$ for true tumour positives is close to 1.
17. If TN($G_{1,1}$) can be regarded as a constant positive integer, FN($G_{1,1}$) can be regarded as close to 0, and the precision of $T_1$ for true tumour negatives can be denoted by TN($G_{1,1}$)/(TN($G_{1,1}$)+ FN($G_{1,1}$)), then the precision of $T_1$ for true tumour negatives is close to 1.
18. If the recall of $T_1$ for true tumour positives is close to 1 and the precision of $T_1$ for true tumour negatives is close to 1, then $T_1$ has a high recall of true tumour positives and a high precision of true tumour negatives.

Secondly, we give the propositional symbols for the above preconditions 1-18 for Reasoning 5, which are shown in Table 7.

Table 7. Propositional symbols of preconditions for Reasoning 5

| Symbol | Meaning |
|---|---|
| $a$ | $RG_1$ is given |
| $b$ | $RG_1$ is equivalent to $G_{1,1}$: pixels of $IS_1$ outside the polygons of $NLS_1$ are tumour negatives |
| $c$ | $RG_2$ is given |
| $d$ | $RG_2$ is equivalent to $G_{1,2}$ pixels of $IS_1$ inside the polygons of $NLS_1$ are tumour positives |
| $e$ | $RG_1$ indicates where are tumour negatives of $IS_1$ |
| $f$ | $RG_2$ indicates where are tumour positives of $IS_1$ |
| $g$ | a target ($T_1$) can be abduced from the union of $RG_1$ and $RG_2$ |
| $h$ | the recall of $T_1$ for true tumour positives can be denoted by true tumour positives covered by $G_{1,2}$ (TP($G_{1,2}$)) dividing the sum of TP($G_{1,2}$) and false tumour negatives covered by $G_{1,1}$ (FN($G_{1,1}$)), i.e., TP($G_{1,2}$)/( TP($G_{1,2}$)+ FN($G_{1,1}$)) |
| $i$ | the precision of $T_1$ for true tumour negatives can be denoted by true tumor negatives covered by $G_{1,1}$ (TN( $G_{1,1}$ )) dividing the sum of TN( $G_{1,1}$ ) and FN( $G_{1,1}$ ), i.e., TN($G_{1,1}$)/(TN($G_{1,1}$)+ FN($G_{1,1}$)) |
| $j$ | $G_{1,1}$ should not be revised to remain $IC_{1,1}$ |
| $k$ | $G_{1,1}$ should not be revised to remain the fact that inconsistency between FN($G_{1,1}$) and false tumour negatives covered by $KB$ (FN($KB$)) is low |
| $l$ | FN($G_{1,1}$) can be regarded as close to FN($KB$) |
| $m$ | FN($G_{1,1}$) can be regarded as close to 0 |
| $n$ | tumour negatives covered by $G_{1,1}$ can be regarded as true tumor negatives |
| $o$ | TN($G_{1,1}$) can be regarded as a constant positive integer |
| $p$ | $G_{1,2}$ should not be revised to remain $IC_{1,3}$ |
| $q$ | $G_{1,2}$ should not be revised to remain the fact that inconsistency between TP($G_{1,2}$) and true tumour positives covered by $KB$ (TP($KB$)) is low |
| $r$ | TP($G_{1,2}$) can be regarded as close to TP($KB$) |
| $s$ | TP($G_{1,2}$) can be regarded as a constant positive integer |
| $t$ | the recall of $T_1$ (TP($G_{1,2}$)/( TP($G_{1,2}$)+ FN($G_{1,1}$)))for true tumour positives is close to 1 |
| $u$ | the precision of $T_1$ (TN($G_{1,1}$)/(TN($G_{1,1}$)+ FN($G_{1,1}$))) for true tumour negatives is close to 1 |
| $v$ | $T_1$ has a high recall of true tumour positives and a high precision of true tumour negatives |

Thirdly, referring to Table 7, we signify the propositional formalizations of the preconditions 1-18 for Reasoning 5 and Reasoning 5 via the propositional connectives listed in Table 1 as follows.

| | |
|---|---|
| 1) $a \rightarrow b$ | Precondition |
| 2) $c \rightarrow d$ | Precondition |
| 3) $(b \land d) \rightarrow (e \land f)$ | Precondition |
| 4) $(e \land f) \rightarrow g$ | Precondition |
| 5) $(g \land b \land d) \rightarrow (h \land i)$ | Precondition |
| 6) $a \rightarrow j$ | Precondition |
| 7) $j \rightarrow k$ | Precondition |
| 8) $k \rightarrow l$ | Precondition |
| 9) $l \rightarrow m$ | Precondition |
| 10) $m \rightarrow n$ | Precondition |
| 11) $n \rightarrow o$ | Precondition |
| 12) $b \rightarrow p$ | Precondition |
| 13) $p \rightarrow q$ | Precondition |
| 14) $q \rightarrow r$ | Precondition |
| 15) $r \rightarrow s$ | Precondition |
| 16) $(s \land m \land h) \rightarrow t$ | Precondition |
| 17) $(o \land m \land i) \rightarrow u$ | Precondition |
| 18) $(t \land u) \rightarrow v$ | Precondition |
| $(a \land c) \rightarrow (g \land v)$ | Reasoning 5 |

Fourthly, we show the validity of Reasoning 5 via the rules for proof of propositional logical reasoning listed in Table 2 as follows.

$\therefore (\boldsymbol{a} \wedge \boldsymbol{c}) \rightarrow (\boldsymbol{g} \wedge \boldsymbol{v})$

| | |
|---|---|
| 19) $a \wedge c$ | Hypothesis |
| 20) $a$ | 19); $\wedge -$ |
| 21) $c$ | 19); $\wedge -$ |
| 22) $b$ | 1),20); MP |
| 23) $d$ | 2),21); MP |
| 24) $b \wedge d$ | 22),23); $\wedge +$ |
| 25) $(b \wedge d) \rightarrow g$ | 3),24); HS |
| 26) $g$ | 25),24); MP |
| 27) $g \wedge b \wedge d$ | 26),24); $\wedge +$ |
| 28) $h \wedge i$ | 5),27); MP |
| 29) $h$ | 28); $\wedge -$ |
| 30) $i$ | 28); $\wedge -$ |
| 31) $a \rightarrow m$ | 6),7),8),9); HS |
| 32) $m$ | 31),20); MP |
| 33) $m \rightarrow o$ | 10),11); HS |
| 34) $o$ | 33),32); MP |
| 35) $b \rightarrow s$ | 12),13),14),15); HS |
| 36) $s$ | 35),22); MP |
| 37) $s \wedge m \wedge h$ | 36),32),29); $\wedge +$ |
| 38) $o \wedge m \wedge i$ | 34),32),30); $\wedge +$ |
| 39) $t$ | 16),37); MP |
| 40) $u$ | 17),38); MP |
| 41) $t \wedge u$ | 39),40); $\wedge +$ |
| 42) $v$ | 18),41); MP |
| 43) $g \wedge v$ | 26),42); $\wedge +$ |
| 48) $(a \wedge c) \rightarrow (g \wedge v)$ | 19)-43); Conditional Proof |

Since the hypothesis $a \wedge c$ of the 19) step can be fulfilled by the Abduction step, Reasoning 5 is proved to be valid.

## Proof of Reasoning 6

**Reasoning 6**. *If $RG_5$ is given and $RG_6$ is given, then a target ($T_2$) can be abduced from the union of $RG_5$ and $RG_6$, and $T_2$ has a high precision of true tumour positives and a high recall of true tumour negatives.*

**Proof-R6**. Firstly, with $RG_5$ and $RG_6$, we have following derived preconditions for Reasoning 6.

1. If $RG_5$ is given, then $RG_5$is equivalent to $G_{2,1}$: pixels of $IS_2$ inside the polygons of $NLS_2$ are tumour positives. (revised grounding produced by Abduction)
2. If $RG_6$ is given, then $RG_6$ is equivalent to $G_{2,2}$: pixels of $IS_2$ outside the polygons of $NLS_2$ are tumour negatives. (revised grounding produced by Abduction)

3. If $RG_5$ is equivalent to $G_{2,1}$: pixels of $IS_2$ inside the polygons of $NLS_2$ are tumour positives and $RG_6$ is equivalent to $G_{2,2}$: pixels of $IS_2$ outside the polygons of $NLS_2$ are tumour negatives, then $RG_5$ indicates where are tumour positives of $IS_2$ and $RG_6$ indicates where are tumour negatives of $IS_2$.
4. If $RG_5$ indicates where are tumour positives of $IS_2$ and $RG_6$ indicates where are tumour negatives of $IS_2$, then a target ($T_2$) can be abduced from the union of $RG_5$ and $RG_6$.
5. If $T_2$ can be abduced from the union of $RG_5$ and $RG_6$, $RG_5$is equivalent to $G_{2,1}$: pixels of $IS_2$ inside the polygons of $NLS_2$ are tumour positives, and $RG_6$ is equivalent to $G_{2,2}$ pixels of $IS_2$ outside the polygons of $NLS_2$ are tumour negatives, then the precision of $T_2$ for true tumour positives can be denoted by true tumour positives covered by $G_{2,1}$ (TP($G_{2,1}$)) dividing the sum of TP($G_{2,1}$) and false tumour positives covered by $G_{2,1}$ (FP( $G_{2,1}$ )), i.e., TP($G_{2,1}$)/( TP($G_{2,1}$)+ FP($G_{2,1}$)); and the recall of $T_2$ for true tumour negatives can be denoted by true tumor negatives covered by $G_{2,2}$ (TN($G_{2,2}$)) dividing the sum of TN($G_{2,2}$) and FP($G_{2,1}$), i.e., TN($G_{2,2}$)/(TN($G_{2,2}$)+ FP($G_{2,1}$)).
6. If $RG_5$ is given, then $G_{2,1}$ should not be revised to remain $IC_{2,1}$. (revised grounding produced by Abduction is only associated with corresponding grounding revision in Abduction)
7. If $G_{2,1}$ should not be revised to remain $IC_{2,1}$, then $G_{2,1}$ should not be revised to remain the fact that inconsistency between FP($G_{2,1}$) and false tumour positives covered by $KB$ (FP($KB$)) is low. (fact contained in $IC_{2,1}$)
8. If $G_{2,1}$ should not be revised to remain the fact that inconsistency between FP($G_{2,1}$) and FP($KB$) is low, then FP($G_{2,1}$) can be regarded as close to FP($KB$).
9. If FP($G_{2,1}$) can be regarded as close to FP($KB$), then FP($G_{2,1}$) can be regarded as close to 0.
10. If FP($G_{2,1}$) can be regarded as close to 0, then tumour positives covered by $G_{2,1}$ can be regarded as true tumor positives.
11. If tumour positives covered by $G_{2,1}$ can be regarded as true tumor positives, then TP($G_{2,1}$) can be regarded as a constant positive integer.
12. If $RG_6$ is given, then $G_{2,2}$ should not be revised to remain $IC_{2,3}$. (revised grounding produced by Abduction is only associated with corresponding grounding revision in Abduction)
13. If $G_{2,2}$ should not be revised to remain $IC_{2,3}$, then $G_{2,2}$ should not be revised to remain the fact that inconsistency between TN($G_{2,2}$) and true tumour negatives covered by $KB$ (TN($KB$)) is low.
14. If $G_{2,2}$ should not be revised to remain the fact that inconsistency between TN($G_{2,2}$) and TN($KB$) is low, then TN($G_{2,2}$) can be regarded as close to TN($KB$).
15. If TN($G_{2,2}$) can be regarded as close to TN($KB$), TN($G_{2,2}$) can be regarded as a constant positive integer.

16. If TP($G_{2,1}$) can be regarded as a constant positive integer, FP($G_{2,1}$) can be regarded as close to 0, and the precision of $T_2$ for true tumour positives can be denoted by TP($G_{2,1}$)/( TP($G_{2,1}$)+ FP($G_{2,1}$)), then the precision of $T_2$ for true tumour positives is close to 1.
17. If TN($G_{2,2}$) can be regarded as a constant positive integer, FP($G_{2,1}$) can be regarded as close to 0, and the recall of $T_2$ for true tumour negatives can be denoted by TN($G_{2,2}$)/(TN($G_{2,2}$)+ FP($G_{2,1}$)), then the recall of $T_2$ for true tumour negatives is close to 1.
18. If the precision of $T_2$ for true tumour positives is close to 1 and the precision of $T_2$ for true tumour negatives is close to 1, then $T_2$ has a high precision of true tumour positives and a high recall of true tumour negatives.

Secondly, we give the propositional symbols for the above preconditions 1-18 for Reasoning 6, which are shown in Table 8.

Table 8. Propositional symbols of preconditions for Reasoning 6

| Symbol | Meaning |
|---|---|
| $a$ | $RG_5$ is given |
| $b$ | $RG_5$ is equivalent to $G_{2,1}$: pixels of $IS_2$ inside the polygons of $NLS_2$ are tumour positives |
| $c$ | $RG_6$ is given |
| $d$ | $RG_6$ is equivalent to $G_{2,2}$: pixels of $IS_2$ outside the polygons of $NLS_2$ are tumour negatives |
| $e$ | $RG_5$ indicates where are tumour positives of $IS_2$ |
| $f$ | $RG_6$ indicates where are tumour negatives of $IS_2$ |
| $g$ | a target ($T_2$) can be abduced from the union of $RG_5$ and $RG_6$ |
| $h$ | the precision of $T_2$ for true tumour positives can be denoted by true tumour positives covered by $G_{2,1}$ (TP($G_{2,1}$)) dividing the sum of TP($G_{2,1}$) and false tumour positives covered by $G_{2,1}$ (FP($G_{2,1}$)), i.e., TP($G_{2,1}$)/( TP($G_{2,1}$)+ FP($G_{2,1}$)) |
| $i$ | the recall of $T_2$ for true tumour negatives can be denoted by true tumor negatives covered by $G_{2,2}$ (TN($G_{2,2}$)) dividing the sum of TN($G_{2,2}$) and FP($G_{2,1}$), i.e., TN($G_{2,2}$)/(TN($G_{2,2}$)+ FP($G_{2,1}$)) |
| $j$ | $G_{2,1}$ should not be revised to remain $IC_{2,1}$ |
| $k$ | $G_{2,1}$ should not be revised to remain the fact that inconsistency between FP($G_{2,1}$) and false tumour positives covered by $KB$ (FP($KB$)) is low |
| $l$ | FP($G_{2,1}$) can be regarded as close to FP($KB$) |
| $m$ | FP($G_{2,1}$) can be regarded as close to 0 |
| $n$ | tumour positives covered by $G_{2,1}$ can be regarded as true tumor positives |
| $o$ | TP($G_{2,1}$) can be regarded as a constant positive integer |
| $p$ | $G_{2,2}$ should not be revised to remain $IC_{2,3}$ |
| $q$ | $G_{2,2}$ should not be revised to remain the fact that inconsistency between TN($G_{2,2}$) and true tumour negatives covered by $KB$ (TN($KB$)) is low |
| $r$ | TN($G_{2,2}$) can be regarded as close to TN($KB$) |
| $s$ | TN($G_{2,2}$) can be regarded as a constant positive integer |
| $t$ | the precision of $T_2$ for true tumour positives is close to 1 |
| $u$ | the recall of $T_2$ for true tumour negatives is close to 1 |
| $v$ | $T_2$ has a high precision of true tumour positives and a high recall of true tumour negatives |

Thirdly, referring to Table 8, we signify the propositional formalizations of the preconditions 1-18 for Reasoning 6 and Reasoning 6 via the propositional connectives listed in Table 1 as follows.

| | |
|---|---|
| 1) $a \rightarrow b$ | Precondition |
| 2) $c \rightarrow d$ | Precondition |
| 3) $(b \wedge d) \rightarrow (e \wedge f)$ | Precondition |
| 4) $(e \wedge f) \rightarrow g$ | Precondition |
| 5) $(g \wedge b \wedge d) \rightarrow (h \wedge i)$ | Precondition |
| 6) $a \rightarrow j$ | Precondition |
| 7) $j \rightarrow k$ | Precondition |
| 8) $k \rightarrow l$ | Precondition |
| 9) $l \rightarrow m$ | Precondition |
| 10) $m \rightarrow n$ | Precondition |
| 11) $n \rightarrow o$ | Precondition |
| 12) $b \rightarrow p$ | Precondition |
| 13) $p \rightarrow q$ | Precondition |
| 14) $q \rightarrow r$ | Precondition |
| 15) $r \rightarrow s$ | Precondition |
| 16) $(o \wedge m \wedge h) \rightarrow t$ | Precondition |
| 17) $(s \wedge m \wedge i) \rightarrow u$ | Precondition |
| 18) $(t \wedge u) \rightarrow v$ | Precondition |
| $(a \wedge c) \rightarrow (g \wedge v)$ | Reasoning 6 |

Fourthly, we show the validity of Reasoning 6 via the rules for proof of propositional logical reasoning listed in Table 2 as follows.

$\therefore (\boldsymbol{a} \wedge \boldsymbol{c}) \rightarrow (\boldsymbol{g} \wedge \boldsymbol{v})$

| | |
|---|---|
| 19) $a \wedge c$ | Hypothesis |
| 20) $a$ | 19); $\wedge -$ |
| 21) $c$ | 19); $\wedge -$ |
| 22) $b$ | 1),20); MP |
| 23) $d$ | 2),21); MP |
| 24) $b \wedge d$ | 22),23); $\wedge +$ |
| 25) $(b \wedge d) \rightarrow g$ | 3),24); HS |
| 26) $g$ | 25),24); MP |
| 27) $g \wedge b \wedge d$ | 26),24); $\wedge +$ |
| 28) $h \wedge i$ | 5),27); MP |
| 29) $h$ | 28); $\wedge -$ |
| 30) $i$ | 28); $\wedge -$ |
| 31) $a \rightarrow m$ | 6),7),8),9); HS |
| 32) $m$ | 31),20); MP |
| 33) $m \rightarrow o$ | 10),11); HS |
| 34) $o$ | 33),32); MP |
| 35) $b \rightarrow s$ | 12),13),14),15); HS |
| 36) $s$ | 35),22); MP |

37) $o \wedge m \wedge h$ 36),32),29); $\wedge +$
38) $s \wedge m \wedge i$ 34),32),30); $\wedge +$
39) $t$ 16),37); MP
40) $u$ 17),38); MP
41) $t \wedge u$ 39),40); $\wedge +$
42) $v$ 18),41); MP
43) $g \wedge v$ 26),42); $\wedge +$
48) $(a \wedge c) \rightarrow (g \wedge v)$ 19)-43); Conditional Proof

Since the hypothesis $a \wedge c$ of the 19) step can be fulfilled by the Abduction step, Reasoning 6 is proved to be valid.

## Proof of Reasoning 7

**Reasoning 7**. *If the target ($T_1$) abduced from the union of $RG_1$ and $RG_2$ is given, $RG_3$ is given and $RG_4$ is given, then $T_1$ has a low precision of true tumour positives and a low recall of true tumour negatives.*

**Proof-R7**. Firstly, with $T_1$, $RG_3$ and $RG_4$, we have following derived preconditions for Reasoning 7.

1. If the target ($T_1$) abduced from the union of $RG_1$ and $RG_2$ is given, then $T_1$ is abduced based on $G_{1,1}$: pixels of $IS_1$ outside the polygons of $NLS_1$ are tumour negatives and $G_{1,2}$: pixels of $IS_1$ inside the polygons of $NLS_1$ are tumour positives. (revised grounding produced by Abduction)
2. If the target ($T_1$) abduced from the union of $RG_1$ and $RG_2$ is given, then the fact that $T_1$ has a high recall of true tumour positives and a high precision of true tumour negatives exists. (Reasoning 5)
3. If $RG_3$ is given, then $RG_3$ is equivalent to the fact that pixels of $IS_1$ outside the polygons of $NLS_1$ are not exactly true tumour negatives. (revised grounding produced by Abduction)
4. If $RG_4$ is given, then $RG_4$ is equivalent to the fact that pixels of $IS_1$ inside the polygons of $NLS_1$ are not exactly true tumour positives. (revised grounding produced by Abduction)
5. If $T_1$ is abduced based on $G_{1,1}$: pixels of $IS_1$ outside the polygons of $NLS_1$ are tumour negatives and $G_{1,2}$: pixels of $IS_1$ inside the polygons of $NLS_1$ are tumour positives, the fact that $T_1$ has a high recall of true tumour positives and a high precision of true tumour negatives exists, $RG_3$ is equivalent to the fact that pixels of $IS_1$ outside the polygons of $NLS_1$ are not exactly true tumour negatives, and $RG_4$ is equivalent to the fact that pixels of $IS_1$ inside the polygons of $NLS_1$ are not exactly true tumour positives, then many true tumour negatives are taken as as tumour positives by $G_{1,2}$.
6. If many true tumour negatives are taken as tumour positives by $G_{1,2}$, then false tumour positives are covered by $G_{1,2}$ (FP($G_{1,2}$)) can be regarded as large.
7. If FP($G_{1,2}$) can be regarded as large, and then $T_1$ has a low precision of true tumour positives and a low recall of true tumour negatives.

Secondly, we give the propositional symbols for the above preconditions 1-7 for Reasoning 7, which are shown in Table 9.

Table 9. Propositional symbols of preconditions for Reasoning 7

| Symbol | Meaning |
|---|---|
| $a$ | the target ($T_1$) abduced from the union of $RG_1$ and $RG_2$ is given |
| $b$ | $RG_3$ is given |
| $c$ | $RG_4$ is given |
| $d$ | $T_1$ is abduced based on $G_{1,1}$: pixels of $IS_1$ outside the polygons of $NLS_1$ are tumour negatives and $G_{1,2}$: pixels of $IS_1$ inside the polygons of $NLS_1$ are tumour positives |
| $e$ | the fact that $T_1$ has a high recall of true tumour positives and a high precision of true tumour negatives exists |
| $f$ | $RG_3$ is equivalent to the fact that pixels of $IS_1$ outside the polygons of $NLS_1$ are not exactly true tumour negatives |
| $g$ | $RG_4$ is equivalent to the fact that pixels of $IS_1$ inside the polygons of $NLS_1$ are not exactly true tumour positives |
| $h$ | many true tumour negatives are taken as as tumour positives by $G_{1,2}$ |
| $i$ | false tumour positives are covered by $G_{1,2}$ (FP($G_{1,2}$)) can be regarded as large |
| $v$ | $T_1$ has a low precision of true tumour positives and a low recall of true tumour negatives |

Thirdly, referring to Table 9, we signify the propositional formalizations of the preconditions 1-7 for Reasoning 7 and Reasoning 7 via the propositional connectives listed in Table 1 as follows.

1) $a \rightarrow d$ Precondition
2) $a \rightarrow e$ Precondition
3) $b \rightarrow f$ Precondition
4) $c \rightarrow g$ Precondition
5) $(d \wedge e \wedge f \wedge g) \rightarrow h$ Precondition
6) $h \rightarrow i$ Precondition
7) $i \rightarrow v$ Precondition

$(a \wedge b \wedge c) \rightarrow v$ Reasoning 7

Fourthly, we show the validity of Reasoning 7 via the rules for proof of propositional logical reasoning listed in Table 2 as follows.

$\therefore (\boldsymbol{a} \wedge \boldsymbol{b} \wedge \boldsymbol{c}) \rightarrow \boldsymbol{v}$

8) $a \wedge b \wedge c$ Hypothesis
9) $a$ 8); $\wedge -$
10) $b$ 8); $\wedge -$
11) $c$ 8); $\wedge -$
12) $d$ 1),9); MP
13) $e$ 2),9); MP
14) $f$ 3),10); MP
15) $g$ 4),11); MP
16) $d \wedge e \wedge f \wedge g$ 12),13),14),15); $\wedge +$
17) $h$ 5),16); MP
18) $h \rightarrow v$ 6),7); HS

19) $v$ 18),17); MP

20) $(a \wedge b \wedge c) \to v$ 8)-19); Conditional Proof

Since the hypothesis $a \wedge b \wedge c$ of the 8) step can be fulfilled by Reasoning 5 and the Abduction step, Reasoning 7 is proved to be valid.

## Proof of Reasoning 8

**Reasoning 8**. *If the target ($T_2$) abduced from the union of $RG_5$ and $RG_6$ is given, $RG_7$ is given and $RG_8$ is given, then $T_2$ has a low recall of true tumour positives and a low precision of true tumour negatives.*

**Proof-R8**. Firstly, with $T_2$, $RG_7$ and $RG_8$, we have following derived preconditions for Reasoning 8.

1. If the target ($T_2$) abduced from the union of $RG_5$ and $RG_6$ is given, then $T_2$ is abduced based on $G_{2,1}$: pixels of $IS_2$ inside the polygons of $NLS_2$ are tumour positives and $G_{2,2}$: pixels of $IS_2$ outside the polygons of $NLS_2$ are tumour negatives. (revised grounding produced by Abduction)
2. If the target ($T_2$) abduced from the union of $RG_5$ and $RG_6$ is given, then the fact that $T_2$ has a high precision of true tumour positives and a high recall of true tumour negatives exists. (Reasoning 6)
3. If $RG_7$ is given, then $RG_7$ is equivalent to the fact that pixels of $IS_1$ inside the polygons of $NLS_1$ are not exactly true tumour positives. (revised grounding produced by Abduction)
4. If $RG_8$ is given, then $RG_8$ is equivalent to the fact that pixels of $IS_1$ outside the polygons of $NLS_1$ are not exactly true tumour negatives. (revised grounding produced by Abduction)
5. If $T_2$ is abduced based on $G_{2,1}$: pixels of $IS_2$ inside the polygons of $NLS_2$ are tumour positives and $G_{2,2}$: pixels of $IS_2$ outside the polygons of $NLS_2$ are tumour negatives, the fact that $T_2$ has a high precision of true tumour positives and a high recall of true tumour negatives exists, $RG_7$ is equivalent to the fact that pixels of $IS_1$ inside the polygons of $NLS_1$ are not exactly true tumour positives, and $RG_8$ is equivalent to the fact that pixels of $IS_1$ outside the polygons of $NLS_1$ are not exactly true tumour negatives, then many true tumour positives are taken as as tumour negatives by $G_{2,2}$.
6. If many true tumour positives are taken as as tumour negatives by $G_{2,2}$, then false tumour negatives are covered by $G_{2,2}$ (FN($G_{2,2}$)) can be regarded as large.
7. If FN($G_{2,2}$) can be regarded as large, and then $T_2$ has a low recall of true tumour positives and a low precision of true tumour negatives.

Secondly, we give the propositional symbols for the above preconditions 1-7 for Reasoning 8, which are shown in Table 10.

Table 10. Propositional symbols of preconditions for Reasoning 8

| Symbol | Meaning |
|---|---|
| $a$ | the target ($T_2$) abduced from the union of $RG_5$ and $RG_6$ is given |
| $b$ | $RG_7$ is given |
| $c$ | $RG_8$ is given |
| $d$ | $T_2$ is abduced based on $G_{2,1}$: pixels of $IS_2$ inside the polygons of $NLS_2$ are tumour positives and $G_{2,2}$: pixels of $IS_2$ outside the polygons of $NLS_2$ are tumour negatives |
| $e$ | the fact that $T_2$ has a high precision of true tumour positives and a high recall of true tumour negatives exists |
| $f$ | $RG_7$ is equivalent to the fact that pixels of $IS_1$ inside the polygons of $NLS_1$ are not exactly true tumour positives |
| $g$ | If $RG_8$ is given, then $RG_8$ is equivalent to the fact that pixels of $IS_1$ outside the polygons of $NLS_1$ are not exactly true tumour negatives |
| $h$ | many true tumour positives are taken as as tumour negatives by $G_{2,2}$ |
| $i$ | false tumour negatives are covered by $G_{2,2}$ (FN($G_{2,2}$)) can be regarded as large |
| $j$ | $T_2$ has a low recall of true tumour positives and a low precision of true tumour negatives |

Thirdly, referring to Table 10, we signify the propositional formalizations of the preconditions 1-7 for Reasoning 8 and Reasoning 8 via the propositional connectives listed in Table 1 as follows.

1) $a \to d$ Precondition
2) $a \to e$ Precondition
3) $b \to f$ Precondition
4) $c \to g$ Precondition
5) $(d \wedge e \wedge f \wedge g) \to h$ Precondition
6) $h \to i$ Precondition
7) $i \to j$ Precondition

$(a \wedge b \wedge c) \to j$ Reasoning 8

Fourthly, we show the validity of Reasoning 8 via the rules for proof of propositional logical reasoning listed in Table 2 as follows.

$\therefore (\boldsymbol{a} \wedge \boldsymbol{b} \wedge \boldsymbol{c}) \to \boldsymbol{j}$

8) $a \wedge b \wedge c$ Hypothesis
9) $a$ 8); $\wedge -$
10) $b$ 8); $\wedge -$
11) $c$ 8); $\wedge -$
12) $d$ 1),9); MP
13) $e$ 2),9); MP
14) $f$ 3),10); MP
15) $g$ 4),11); MP
16) $d \wedge e \wedge f \wedge g$ 12),13),14),15); $\wedge +$
17) $h$ 5),16); MP
18) $h \to j$ 6),7); HS
19) $j$ 18),17); MP
20) $(a \wedge b \wedge c) \to j$ 8)-19); Conditional Proof

Since the hypothesis $a \wedge b \wedge c$ of the 8) step can be fulfilled by Reasoning 5 and the Abduction step, Reasoning 8 is proved to be valid.

## Proof of Reasoning 9

**Reasoning 9**. *If the target ($T_1$) abduced from the union of $RG_1$ and $RG_2$ is given and the target ($T_2$) abduced from the union of $RG_5$ and $RG_6$ is given, then a target ($T_3$) can be abduced by improving $T_1$ with $T_2$, a target ($T_4$) can be abduced by improving $T_2$ with $T_1$, $T_3$ can have a relatively higher precision of true tumour positives than $T_1$ and a relatively higher recall of true tumour negatives than $T_1$, and $T_4$ can have a relatively higher recall of true tumour positives than $T_2$ and a relatively higher precision of true tumour negatives than $T_2$.*

**Proof-R9**. Firstly, with $T_1$ and $T_2$, we have following derived preconditions for Reasoning 9.

1. If the target ($T_1$) abduced from the union of $RG_1$ and $RG_2$ is given, then $T_1$ has a high recall of true tumour positives and a high precision of true tumour negatives and $T_1$ has a low precision of true tumour positives and a low recall of true tumour negatives. (Reasoning 5 and Reasoning 7)
2. If the target ($T_2$) abduced from the union of $RG_5$ and $RG_6$ is given, $T_2$ has a high precision of true tumour positives and a high recall of true tumour negatives and $T_2$ has a low recall of true tumour positives and a low precision of true tumour negatives. (Reasoning 6 and Reasoning 8)
3. If $T_1$ has a low precision of true tumour positives and a low recall of true tumour negatives and $T_2$ has a high precision of true tumour positives and a high recall of true tumour negatives, then $T_2$ is complementary to $T_1$ to represent the true target.
4. If $T_2$ is complementary to $T_1$ to represent the true target, then a target ($T_3$) can be abduced by improving $T_1$ with $T_2$.
5. If $T_3$ can be abduced by improving $T_1$ with $T_2$, $T_2$ has a high precision of true tumour positives and a high recall of true tumour negatives, and $T_1$ has a low precision of true tumour positives and a low recall of true tumour negatives, then $T_3$ can have a relatively higher precision of true tumour positives than $T_1$ and a relatively higher recall of true tumour negatives than $T_1$.
6. If $T_2$ has a low recall of true tumour positives and a low precision of true tumour negatives and $T_1$ has a high recall of true tumour positives and a high precision of true tumour negatives, then $T_1$ is complementary to $T_2$ to represent the true target.
7. If $T_1$ is complementary to $T_2$ to represent the true target, then a target ($T_4$) can be abduced by improving $T_2$ with $T_1$.
8. If $T_4$ can be abduced by improving $T_2$ with $T_1$, $T_1$ has a high recall of true tumour positives and a high precision of true tumour negatives, and $T_2$ has a low recall of true tumour positives and a low precision of true tumour negatives, then $T_4$ can have a relatively higher recall of true tumour positives than $T_2$ and a relatively higher precision of true tumour negatives than $T_2$.

Secondly, we give the propositional symbols for the above preconditions 1-8 for Reasoning 9, which are shown in Table 11.

Table 11. Propositional symbols of preconditions for Reasoning 9

| Symbol | Meaning |
|---|---|
| $a$ | the target ($T_1$) abduced from the union of $RG_1$ and $RG_2$ is given |
| $b$ | the target ($T_2$) abduced from the union of $RG_5$ and $RG_6$ is given |
| $c$ | $T_1$ has a high recall of true tumour positives and a high precision of true tumour negatives |
| $d$ | $T_1$ has a low precision of true tumour positives and a low recall of true tumour negatives |
| $e$ | $T_2$ has a high precision of true tumour positives and a high recall of true tumour negatives |
| $f$ | $T_2$ has a low recall of true tumour positives and a low precision of true tumour negatives |
| $g$ | $T_2$ is complementary to $T_1$ to represent the true target |
| $h$ | a target ($T_3$) can be abduced by improving $T_1$ with $T_2$ |
| $i$ | $T_3$ can have a relatively higher precision of true tumour positives than $T_1$ and a relatively higher recall of true tumour negatives than $T_1$ |
| $j$ | $T_1$ is complementary to $T_2$ to represent the true target |
| $k$ | a target ($T_4$) can be abduced by improving $T_2$ with $T_1$ |
| $l$ | $T_4$ can have a relatively higher recall of true tumour positives than $T_2$ and a relatively higher precision of true tumour negatives than $T_2$ |

Thirdly, referring to Table 11, we signify the propositional formalizations of the preconditions 1-8 for Reasoning 9 and Reasoning 9 via the propositional connectives listed in Table 1 as follows.

| | |
|---|---|
| 1) $a \rightarrow (c \wedge d)$ | Precondition |
| 2) $b \rightarrow (e \wedge f)$ | Precondition |
| 3) $(d \wedge e) \rightarrow g$ | Precondition |
| 4) $g \rightarrow h$ | Precondition |
| 5) $(g \wedge e \wedge d) \rightarrow i$ | Precondition |
| 6) $(f \wedge c) \rightarrow j$ | Precondition |
| 7) $j \rightarrow k$ | Precondition |
| 8) $(k \wedge c \wedge f) \rightarrow l$ | Precondition |
| $(a \wedge b) \rightarrow (h \wedge k \wedge i \wedge l)$ | Reasoning 9 |

Fourthly, we show the validity of Reasoning 9 via the rules for proof of propositional logical reasoning listed in Table 2 as follows.

$\therefore (\boldsymbol{a} \wedge \boldsymbol{b}) \rightarrow (\boldsymbol{h} \wedge \boldsymbol{k} \wedge \boldsymbol{i} \wedge \boldsymbol{l})$

| | |
|---|---|
| 9) $a \wedge b$ | Hypothesis |
| 10) $a$ | 9); $\wedge -$ |
| 11) $b$ | 9); $\wedge -$ |
| 12) $c \wedge d$ | 1),10); MP |
| 13) $e \wedge f$ | 2),11); MP |
| 14) $c$ | 12); $\wedge -$ |
| 15) $d$ | 12); $\wedge -$ |
| 16) $e$ | 13); $\wedge -$ |
| 17) $f$ | 13); $\wedge -$ |
| 18) $d \wedge e$ | 15),16); $\wedge +$ |
| 19) $g$ | 3),18); MP |
| 20) $h$ | 4),19); MP |
| 21) $g \wedge e \wedge d$ | 19),16),15); $\wedge +$ |
| 22) $i$ | 21),5); MP |
| 23) $f \wedge c$ | 17),14); $\wedge +$ |
| 24) $j$ | 6),23); MP |
| 25) $k$ | 7),24); MP |
| 26) $k \wedge c \wedge f$ | 25),14),17); $\wedge +$ |
| 27) $l$ | 8),26); MP |
| 28) $h \wedge k \wedge i \wedge l$ | 20),25),22),27); $\wedge +$ |
| 29) $(a \wedge b) \rightarrow (h \wedge k \wedge i \wedge l)$ | 9)-28); Conditional Proof |

Since the hypothesis $a \wedge b$ of the 9) step can be fulfilled by Reasoning 5 and Reasoning 6, Reasoning 9 is proved to be valid.

## Proof of Reasoning 10

**Reasoning 10**. *If $T_1$ is given and $T_3$ is given, then $T_1$ and $T_3$ can be combined to approximate the true target for $IS_1$.*

**Proof-R10**. Firstly, with $T_1$ and $T_3$, we have following derived preconditions for Reasoning 10.

1. If $T_1$ is given and $T_3$ is given, then $T_1$ and $T_3$ are both corresponding to $IS_1$.
2. If $T_1$ is given, then $T_1$ has a high recall of true tumour positives and a high precision of true tumour negatives and $T_1$ has a low precision of true tumour positives and a low recall of true tumour negatives. (Reasoning 5 and Reasoning 7)
3. If $T_3$ is given, then $T_3$ can have a relatively higher precision of true tumour positives than $T_1$ and a relatively higher recall of true tumour negatives than $T_1$. (Reasoning 9)
4. If $T_1$ has a high recall of true tumour positives and a high precision of true tumour negatives and $T_3$ can have a relatively higher precision of true tumour positives than $T_1$ and a relatively higher recall of true tumour negatives than $T_1$, then $T_1$ and $T_3$ can be combined to possess a high recall of true tumour positives and a high precision of true tumour negatives while having a relatively higher precision of true tumour positives and a relatively higher recall of true tumour negatives.
5. If $T_1$ and $T_3$ can be combined to possess a high recall of true tumour positives and a high precision of true tumour negatives while having a relatively higher precision of true tumour positives and a relatively higher recall of true tumour negatives, and $T_1$ and $T_3$ are both corresponding to $IS_1$, then $T_1$ and $T_3$ can be combined to approximate the true target for $IS_1$.

Secondly, we give the propositional symbols for the above preconditions 1-5 for Reasoning 10, which are shown in Table 12.

Table 12. Propositional symbols of preconditions for Reasoning 10

| Symbol | Meaning |
|---|---|
| $a$ | $T_1$is given |
| $b$ | $T_3$is given |
| $c$ | $T_1$ and $T_3$ are both corresponding to $IS_1$ |
| $d$ | $T_1$ has a high recall of true tumour positives and a high precision of true tumour negatives and $T_1$ has a low precision of true tumour positives and a low recall of true tumour negatives |
| $e$ | $T_3$ can have a relatively higher precision of true tumour positives than $T_1$ and a relatively higher recall of true tumour negatives than $T_1$ |
| $f$ | $T_1$ and $T_3$ can be combined to possess a high recall of true tumour positives and a high precision of true tumour negatives while having a relatively higher precision of true tumour positives and a relatively higher recall of true tumour negatives |
| $g$ | $T_1$ and $T_3$ can be combined to approximate the true target for $IS_1$ |

Thirdly, referring to Table 12, we signify the propositional formalizations of the preconditions 1-5 for Reasoning 10 and Reasoning 10 via the propositional connectives listed in Table 1 as follows.

| | |
|---|---|
| 1) $(a \land b) \rightarrow c$ | Precondition |
| 2) $a \rightarrow d$ | Precondition |
| 3) $b \rightarrow e$ | Precondition |
| 4) $(d \land e) \rightarrow f$ | Precondition |
| 5) $(f \land c) \rightarrow g$ | Precondition |
| $(a \land b) \rightarrow g$ | Reasoning 10 |

Fourthly, we show the validity of Reasoning 10 via the rules for proof of propositional logical reasoning listed in Table 2 as follows.

$\therefore (\boldsymbol{a} \land \boldsymbol{b}) \rightarrow \boldsymbol{g}$

| | |
|---|---|
| 6) $a \land b$ | Hypothesis |
| 7) $a$ | 6); $\land -$ |
| 8) $b$ | 6); $\land -$ |
| 9) $c$ | 1),6); MP |
| 10) $d$ | 2),7); MP |
| 11) $e$ | 3),8); MP |
| 12) $d \land e$ | 10),11); $\land +$ |
| 13) $f$ | 4),12); MP |
| 14) $f \land c$ | 13),9); $\land +$ |
| 15) $g$ | 5),14); MP |
| 16) $(a \land b) \rightarrow g$ | 6)-15); Conditional Proof |

Since the hypothesis $a \land b$ of the 6) step can be fulfilled by the Target Abduce step, Reasoning 10 is proved to be valid.

## Proof of Reasoning 11

**Reasoning 11**. *If $T_2$ is given and $T_4$ is given, then $T_2$ and $T_4$ can be combined to approximate the true target for $NS_2$.*

**Proof-R11**. Firstly, with $T_2$ and $T_4$, we have following derived preconditions for Reasoning 11.

1. If $T_2$ is given and $T_4$ is given, then $T_2$ and $T_4$ are both corresponding to $IS_2$.

2. If $T_2$ is given, then $T_2$ has a high precision of true tumour positives and a high recall of true tumour negatives and $T_2$ has a low recall of true tumour positives and a low precision of true tumour negatives. (Reasoning 6 and Reasoning 8)
3. If $T_4$ is given, then $T_4$ can have a relatively higher recall of true tumour positives than $T_2$ and a relatively higher precision of true tumour negatives than $T_2$. (Reasoning 9)
4. If $T_2$ has a high precision of true tumour positives and a high recall of true tumour negatives and $T_2$ has a low recall of true tumour positives and a low precision of true tumour negatives, and $T_4$ can have a relatively higher recall of true tumour positives than $T_2$ and a relatively higher precision of true tumour negatives than $T_2$, then $T_2$ and $T_4$ can be combined to possess a high precision of true tumour positives and a high recall of true tumour negatives while having a relatively higher recall of true tumour positives and a relatively higher precision of true tumour negatives.
5. If $T_2$ and $T_4$ can be combined to possess a high precision of true tumour positives and a high recall of true tumour negatives while having a relatively higher recall of true tumour positives and a relatively higher precision of true tumour negatives, and $T_2$ and $T_4$ are both corresponding to $IS_2$, then $T_2$ and $T_4$ can be combined to approximate the true target for $IS_2$.

Secondly, we give the propositional symbols for the above preconditions 1-5 for Reasoning 11, which are shown in Table 13.

Table 13. Propositional symbols of preconditions for Reasoning 11

| Symbol | Meaning |
|---|---|
| $a$ | $T_2$is given |
| $b$ | $T_4$is given |
| $c$ | $T_2$ and $T_4$ are both corresponding to $IS_2$ |
| $d$ | $T_2$ has a high precision of true tumour positives and a high recall of true tumour negatives and $T_2$ has a low recall of true tumour positives and a low precision of true tumour negatives |
| $e$ | $T_4$ can have a relatively higher recall of true tumour positives than $T_2$ and a relatively higher precision of true tumour negatives than $T_2$ |
| $f$ | $T_2$ and $T_4$ can be combined to possess a high precision of true tumour positives and a high recall of true tumour negatives while having a relatively higher recall of true tumour positives and a relatively higher precision of true tumour negatives |
| $g$ | $T_2$ and $T_4$ can be combined to approximate the true target for $IS_2$ |

Thirdly, referring to Table 13, we signify the propositional formalizations of the preconditions 1-5 for Reasoning 11 and Reasoning 11 via the propositional connectives listed in Table 1 as follows.

1) $(a \wedge b) \rightarrow c$ Precondition
2) $a \rightarrow d$ Precondition
3) $b \rightarrow e$ Precondition
4) $(d \wedge e) \rightarrow f$ Precondition
5) $(f \wedge c) \rightarrow g$ Precondition

$(a \wedge b) \rightarrow g$ Reasoning 10

Fourthly, we show the validity of Reasoning 11 via the rules for proof of propositional logical reasoning listed in Table 2 as follows.

$\therefore (\boldsymbol{a} \wedge \boldsymbol{b}) \rightarrow \boldsymbol{g}$

| | |
|---|---|
| 6) $a \wedge b$ | Hypothesis |
| 7) $a$ | 6); $\wedge -$ |
| 8) $b$ | 6); $\wedge -$ |
| 9) $c$ | 1),6); MP |
| 10) $d$ | 2),7); MP |
| 11) $e$ | 3),8); MP |
| 12) $d \wedge e$ | 10),11); $\wedge +$ |
| 13) $f$ | 4),12); MP |
| 14) $f \wedge c$ | 13),9); $\wedge +$ |
| 15) $g$ | 5),14); MP |
| 16) $(a \wedge b) \rightarrow g$ | 6)-15); Conditional Proof |

Since the hypothesis $a \wedge b$ of the 6) step can be fulfilled by the Target Abduce step, Reasoning 11 is proved to be valid.

# Supplementary 2

## Multiple Targets Abduced from DiNS

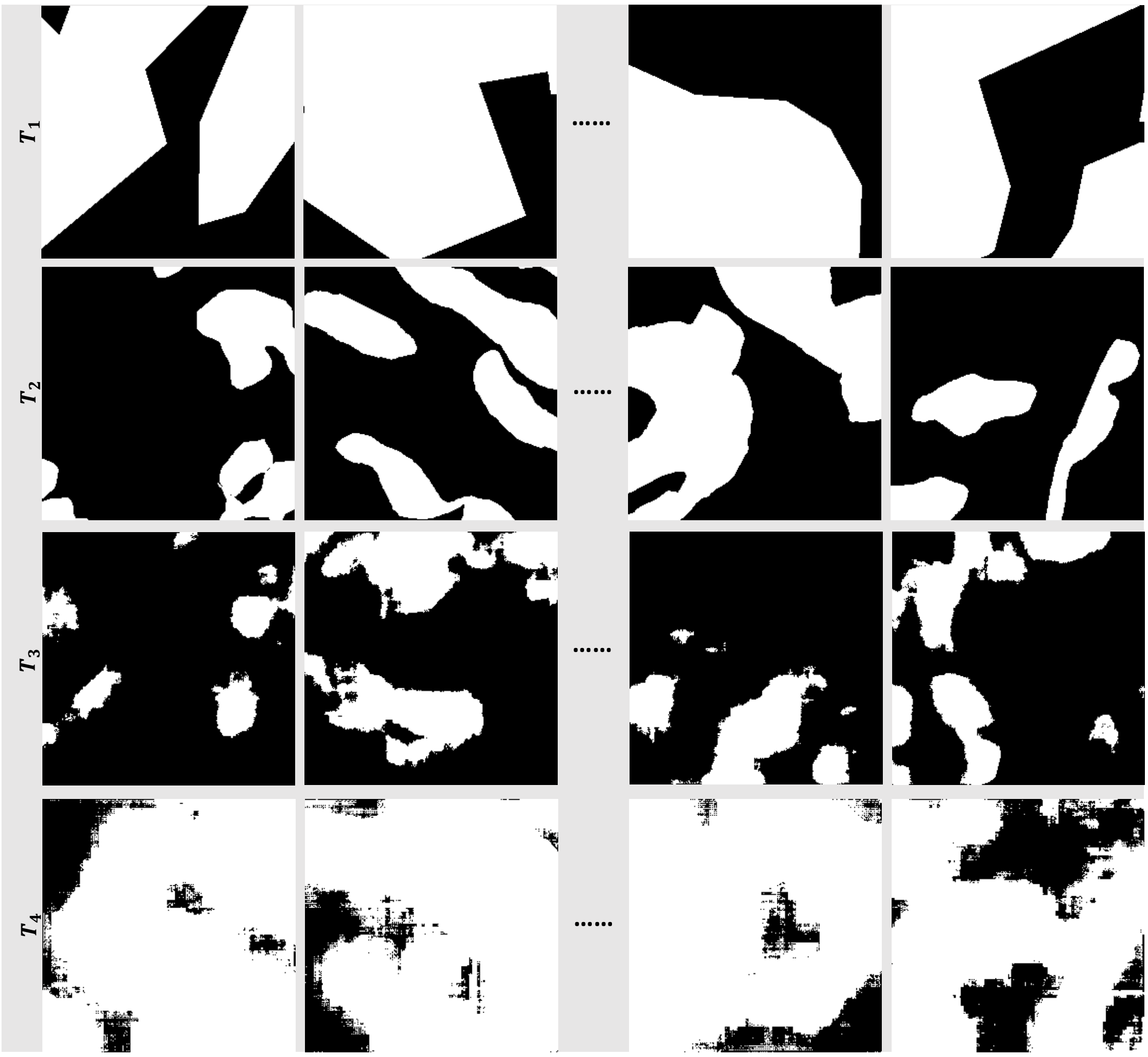

Fig. 1. Examples of multiple targets abduced from the diverse noisy samples provided for tumour segmentation in HE-stained pre-treatment biopsy images.

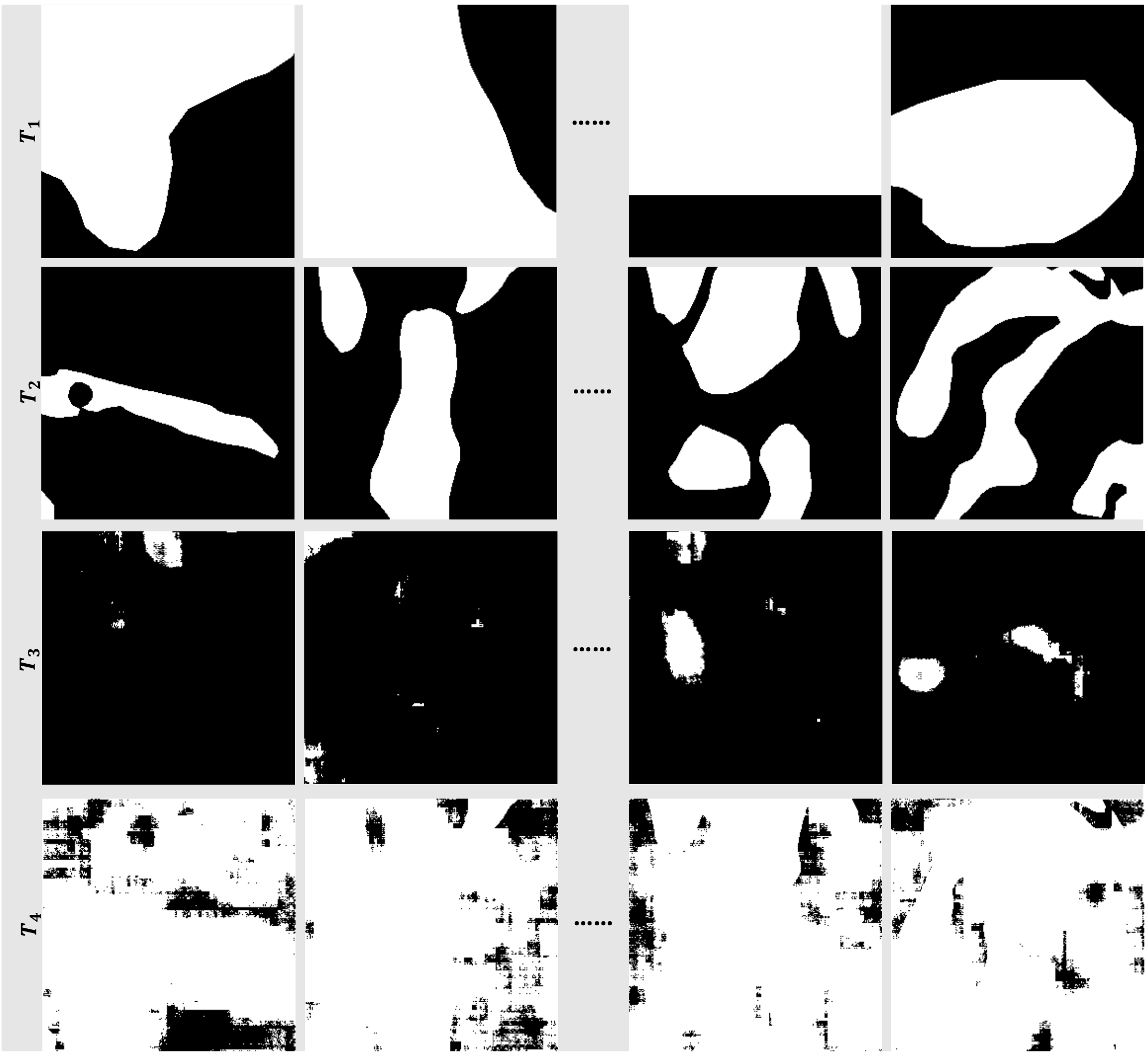

Fig. 2. Examples of multiple targets abduced from the diverse noisy samples provided for tumour segmentation in HE-stained post-treatment surgical resection images.

## Rearranged Multiple Targets

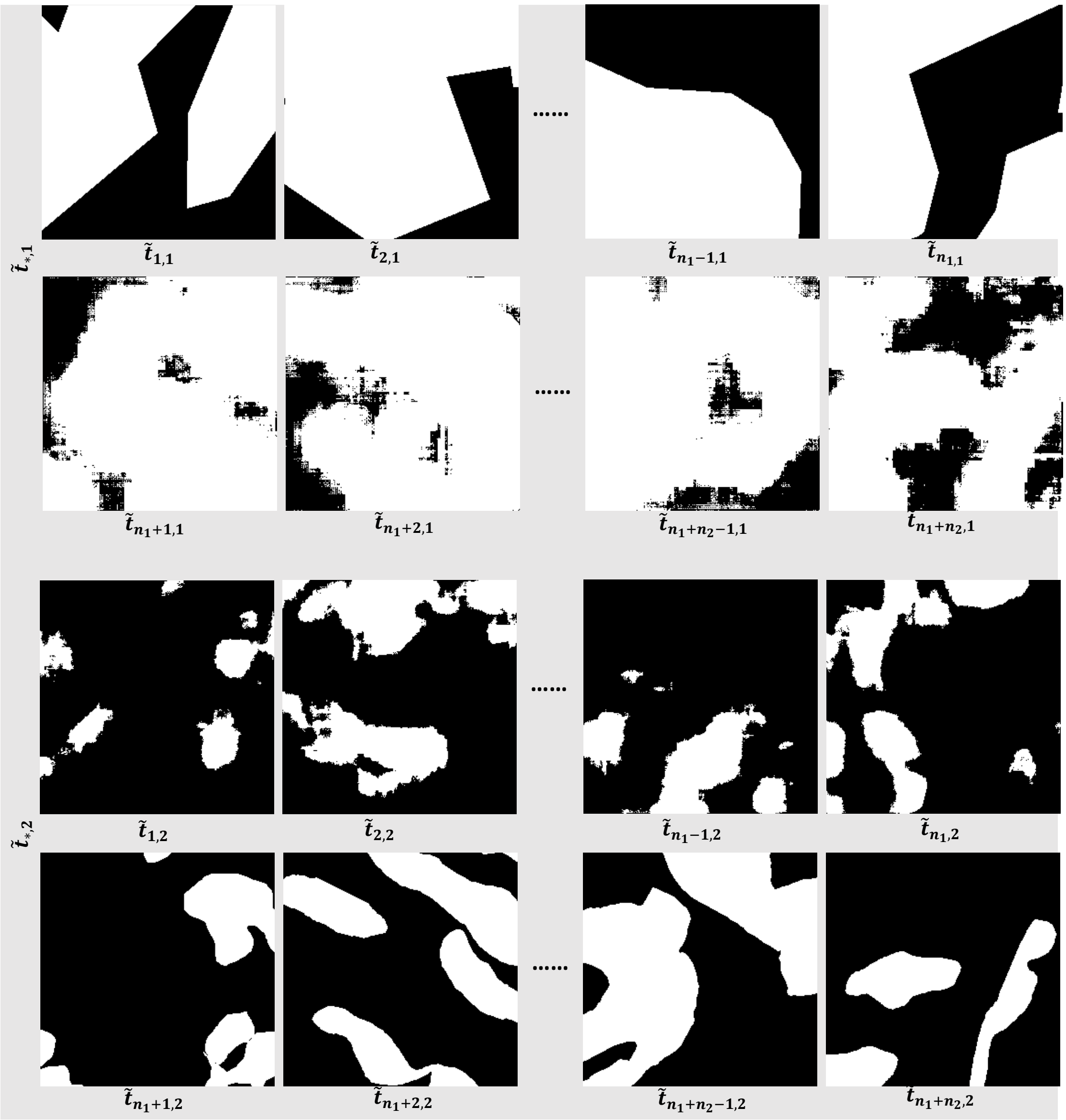

Fig. 3. Examples of rearranged multiple targets corresponding to Fig. 1.

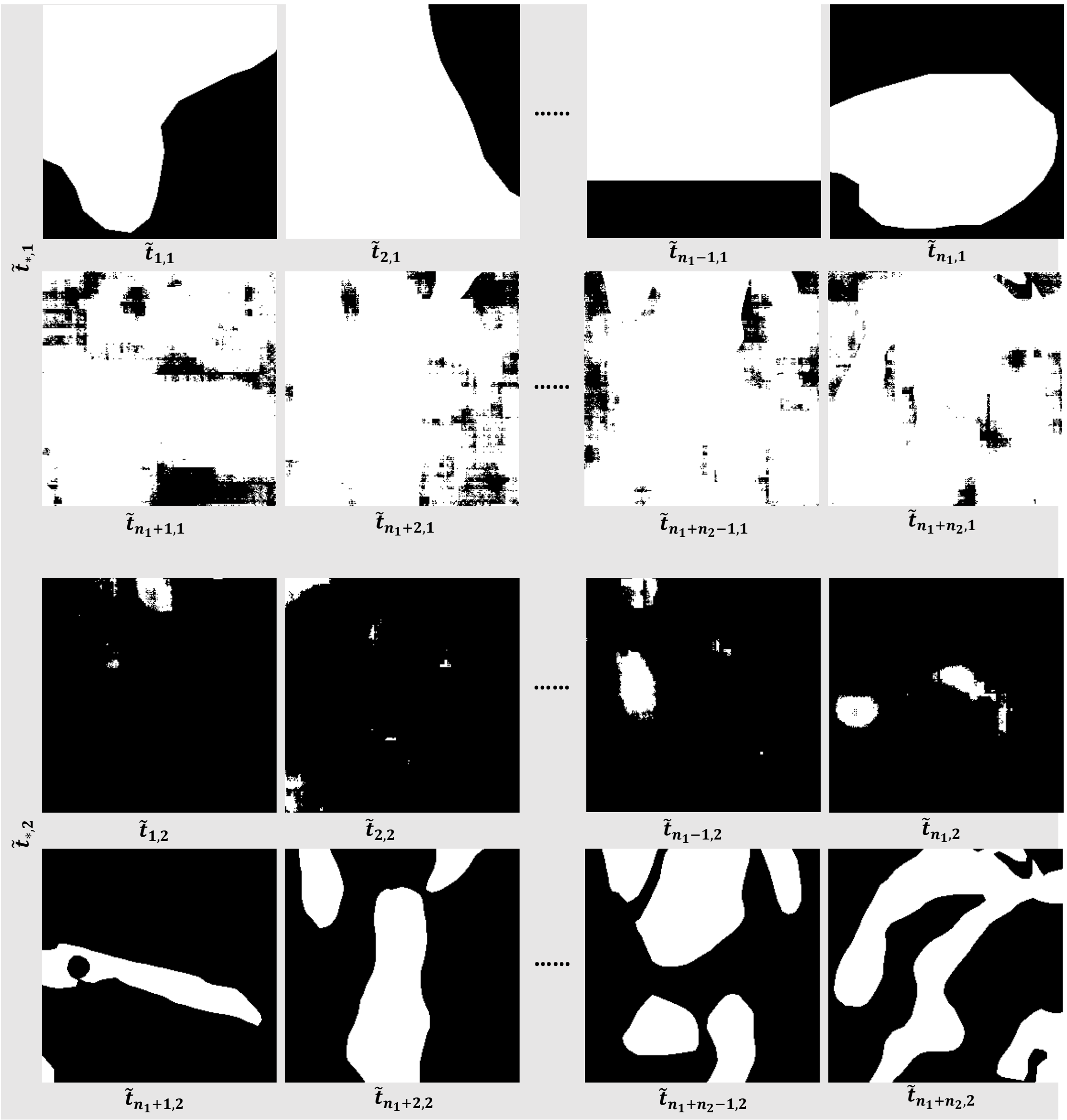

Fig. 4. Examples of rearranged multiple targets corresponding to Fig. 2.

# Supplementary 3

More typical testing results of various state-of-the-art approaches (respectively without OSAMTL-DiNS introduced and with OSAMTL-DiNS introduced) for handling complex noisy labels on Task1:

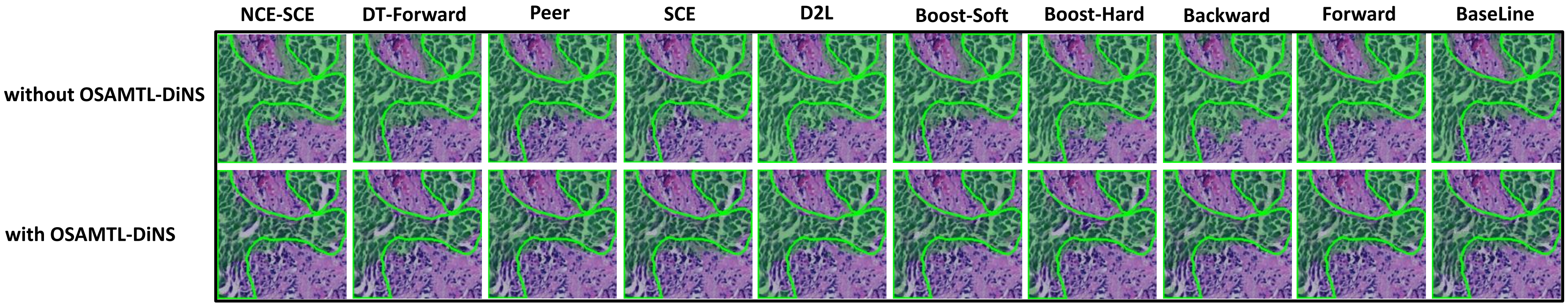

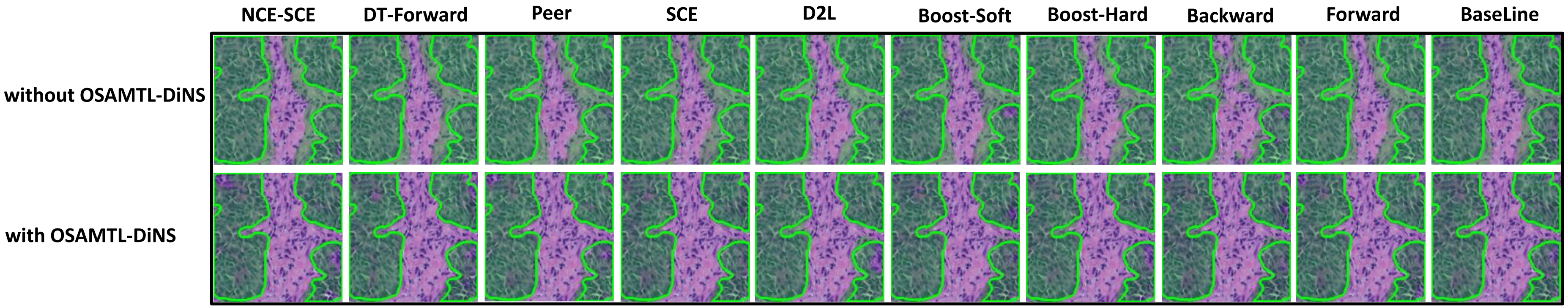

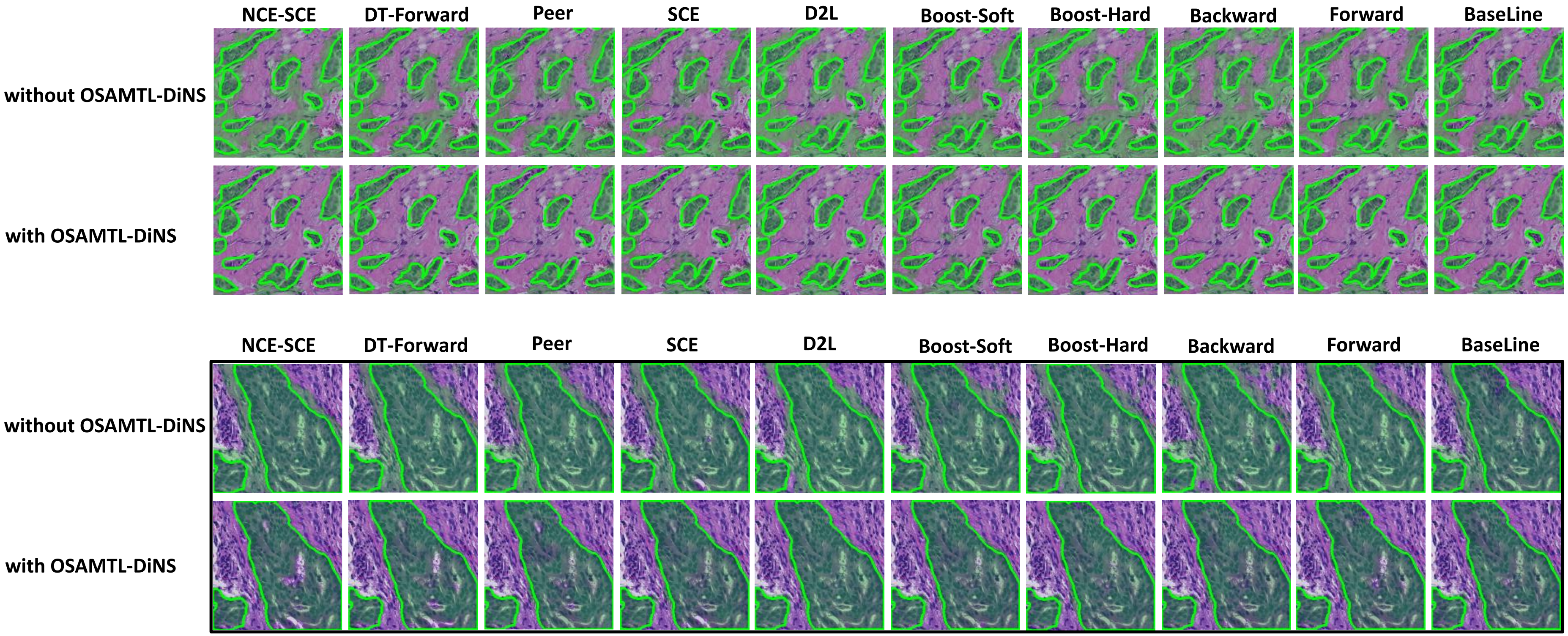
NCE-SCE
DT-Forward
Peer
SCE
D2L
Boost-Soft
Boost-Hard
Backward
Forward
BaseLine
without OSAMTL-DiNS
with OSAMTL-DiNS
NCE-SCE
DT-Forward
Peer
SCE
D2L
Boost-Soft
Boost-Hard
Backward
Forward
BaseLine
without OSAMTL-DiNS
with OSAMTL-DiNS

More typical testing results of various state-of-the-art approaches (respectively without OSAMTL-DiNS introduced and with OSAMTL-DiNS introduced) for handling complex noisy labels on Task2:

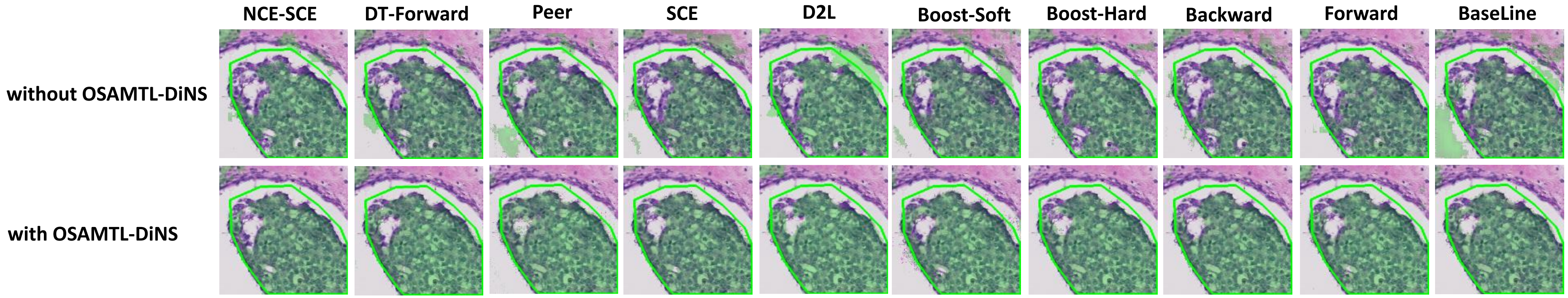

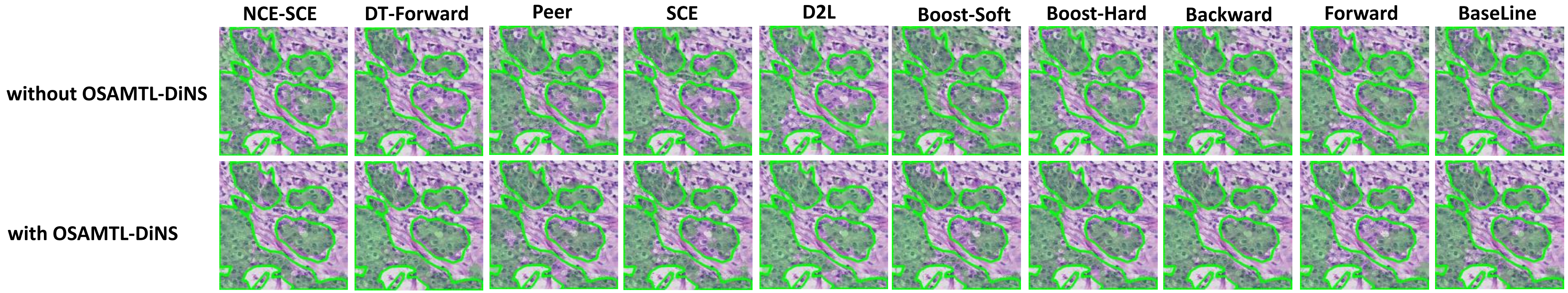

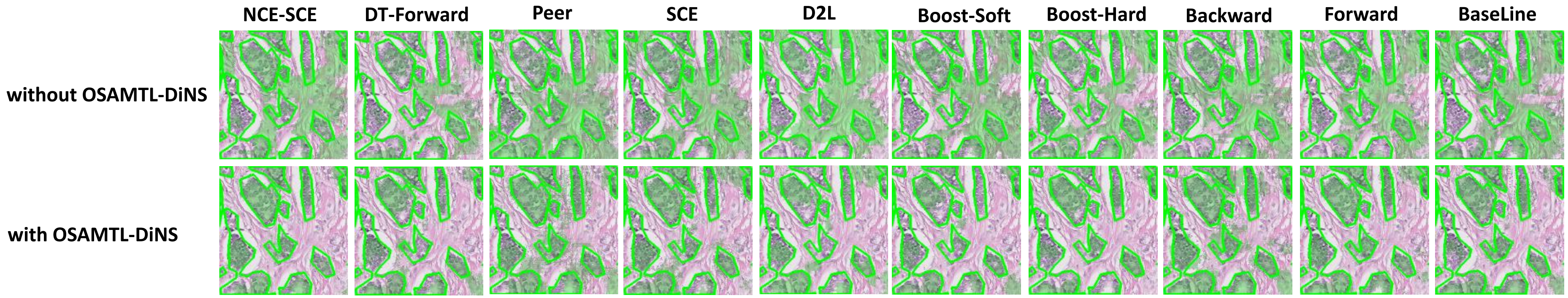
NCE-SCE
DT-Forward
Peer
SCE
D2L
Boost-Soft
Boost-Hard
Backward
Forward
BaseLine
without OSAMTL-DiNS
with OSAMTL-DiNS

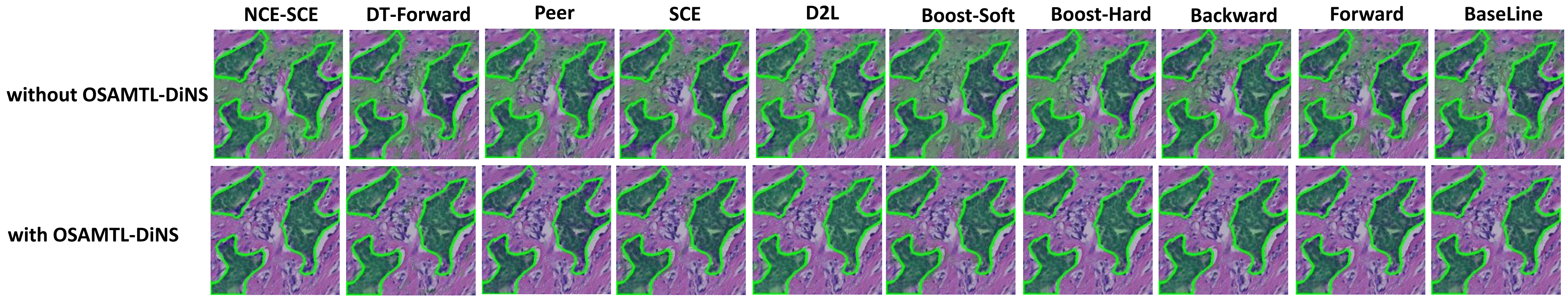
NCE-SCE
DT-Forward
Peer
SCE
D2L
Boost-Soft
Boost-Hard
Backward
Forward
BaseLine
without OSAMTL-DiNS
with OSAMTL-DiNS